\documentclass[10pt,twocolumn,letterpaper]{article}

\usepackage[pagenumbers]{wacv} % To force page numbers, e.g. for an arXiv version

\usepackage{colortbl}
\usepackage{amsthm}
\usepackage{algorithm}
\usepackage{algorithmic}
\usepackage{listings}
\usepackage{multirow}
\newcommand{\tworow}[2]{\begin{tabular}{@{}l@{}}#1 \\ #2\end{tabular}}
\theoremstyle{plain}
\newtheorem{theorem}{Theorem}[section]

\theoremstyle{definition}
\newtheorem{definition}[theorem]{Definition}

\theoremstyle{remark}

\definecolor{wacvblue}{rgb}{0.21,0.49,0.74}
\usepackage[pagebackref,breaklinks,colorlinks,allcolors=wacvblue]{hyperref}

\def\wacvPaperID{232} % *** Enter the WACV Paper ID here
\def\confName{WACV}
\def\confYear{2026}

\title{Quantifying the Limits of Segmentation Foundation Models: \\ Modeling Challenges in Segmenting Tree-Like and Low-Contrast Objects}

\author{
  Yixin Zhang$^{1}$\thanks{Equal contribution.} \quad Nicholas Konz$^{1*}$ \quad Kevin Kramer$^{5}$ \quad Maciej A. Mazurowski$^{1,2,3,4}$ \\
  $^{1}$ Department of Electrical and Computer Engineering, $^{2}$ Department of Computer Science, \\
  $^{3}$ Department of Radiology, $^{4}$ Department of Biostatistics \& Bioinformatics,
  Duke University\\
  $^{5}$ Minnesota Health Solutions \\
  {\tt\small \{yixin.zhang7, nicholas.konz, maciej.mazurowski\}@duke.edu, kevin@minnhealth.com}
}

\begin{document}
\maketitle
\begin{abstract}
Image segmentation foundation models (SFMs) like Segment Anything Model (SAM) have achieved impressive zero-shot and interactive segmentation across diverse domains.
However, they struggle to segment objects with certain structures, particularly those with dense, tree-like morphology and low textural contrast from their surroundings. 
These failure modes are crucial for understanding the limitations of SFMs in real-world applications. 
To systematically study this issue, we introduce interpretable metrics quantifying object tree-likeness and textural separability. On carefully controlled synthetic experiments and real-world datasets, we show that SFM performance (\eg, SAM, SAM 2, HQ-SAM) noticeably correlates with these factors. We attribute these failures to SFMs misinterpreting local structure as global texture, resulting in over-segmentation or difficulty distinguishing objects from similar backgrounds. Notably, targeted fine-tuning fails to resolve this issue, indicating a fundamental limitation. Our study provides the first quantitative framework for modeling the behavior of SFMs on challenging structures, offering interpretable insights into their segmentation capabilities.\footnote{Code: \url{https://github.com/mazurowski-lab/SAMFailureMetrics}}
\end{abstract}
\section{Introduction}
\label{sec:intro}
Segment Anything Model (SAM) \cite{sam}, the most widely-used segmentation foundation model (SFM), has demonstrated promising zero-shot and fine-tuned segmentation ability across diverse domains, including biomedical imaging \cite{MAZUROWSKI2023102918,huang2024segment,osti_10447851}, remote sensing \cite{chen2024rsprompter,ren2024segment}, and others. However, certain \textit{failure modes} for SAM have been found empirically, where SAM produces underwhelming performance on datasets with certain uncommon objects or atypical contexts, such as dense, branching structures (\eg, retinal blood vessels) \cite{shi2023generalist,dong_efficient_2024}, concealed/low-contrast objects \cite{chen2023sam,tang2023can,ji2024segment,wang2025orderaware}, small or irregular objects, and others \cite{ji2024segment}.
% \footnote{We focus on ``unintentional'' failure modes, as opposed to intentional adversarial attacks/examples \cite{zhang2023attack,zheng2023black}.}. 

We will show that newer SFMs, including SAM 2 \cite{sam2} and HQ-SAM \cite{hqsam}, are also susceptible to these failure modes, despite some stated improvements over these limitations for SAM---HQ-SAM on thin, branching structures and SAM 2 on low-contrast objects.
Fine-tuning or adapting these models for such challenging examples using various strategies \cite{ma2024segment,peng2024sam,gu2024build,hqsam} (or utilizing specialized techniques, \eg, matting algorithms \cite{lin2021real}) could mitigate these issues, yet we will show that such strategies do not fully resolve the problems, nor explain the root causes. This underscores the need for a theoretical understanding of why SFMs struggle with such cases, paving the way for better model design and adaptation for new applications.

\begin{figure}[t!]
    \centering
    \includegraphics[width=0.99\linewidth]{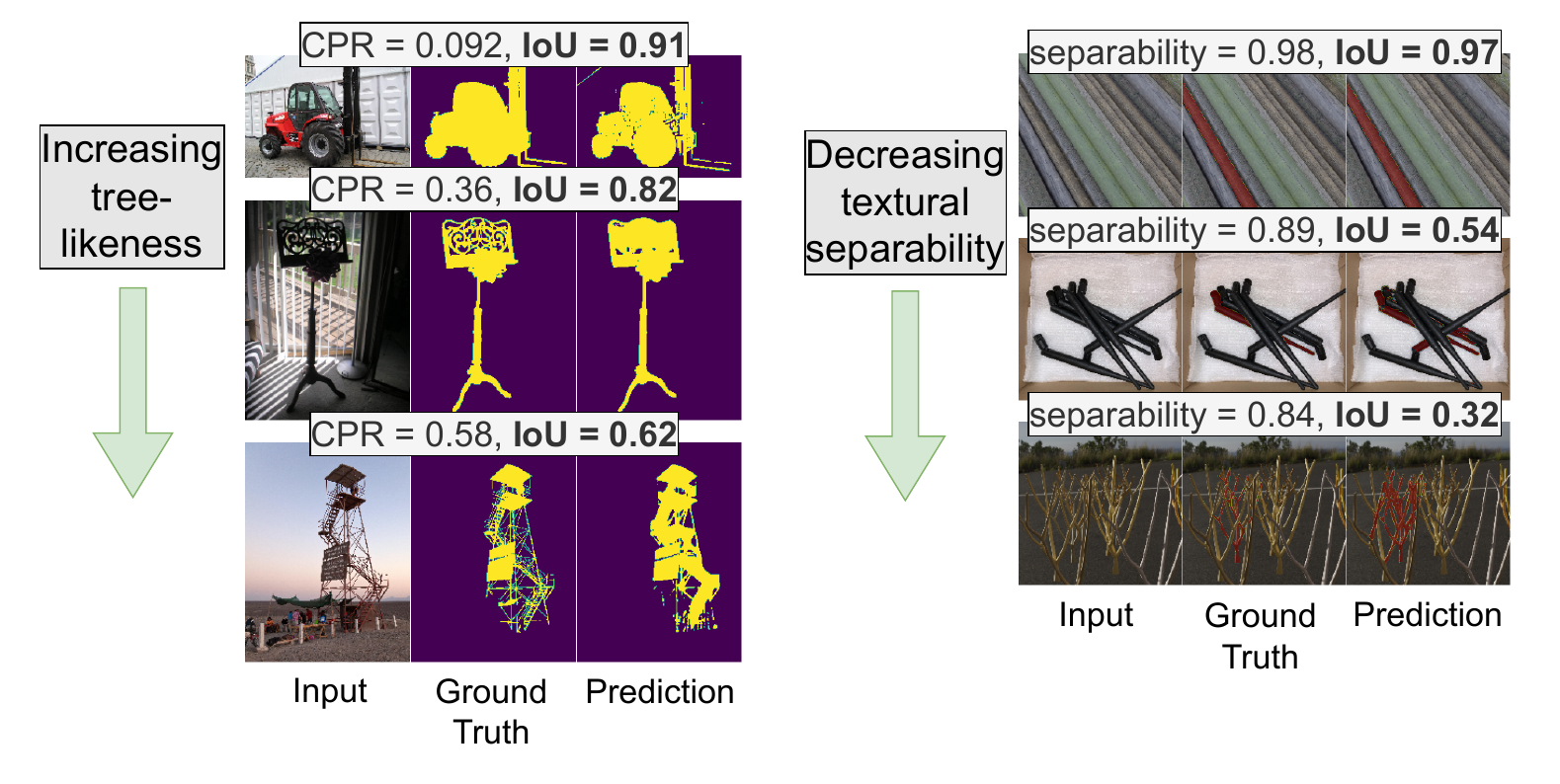}
    \caption{SAM's segmentation performance tends to drop noticeably when the object has high tree-likeness (left, on DIS) or low textural separability (right, on iShape)--even with heavy prompting--which we investigate in this work.}
    \label{fig:teaser}
\end{figure}

Here, we propose that several of these failure modes in SFMs stem from misinterpreting textural cues when distinguishing objects from their surroundings. Specifically, SFMs often confuse dense, tree-like structures as large-scale textures, leading to over-segmentation, and struggle with objects that have similar textures to their surroundings. 
To quantify these challenges, we introduce interpretable metrics for object \textbf{tree-likeness} and \textbf{textural separability}, designed to quantify specific aspects of objects that may make them challenging to segment by typical SFM architectures. Tree-likeness (Sec. \ref{sec:treelikeness_methods}) captures structural complexity with two new metrics, involving spatial contour frequency and the difference in object variability between global and local scales, inspired by how vision transformer encoders and attention patterns in SFMs can be ``confused'' by tree-like objects. Textural separability (Sec. \ref{sec:sep_methods}) measures how distinguishable an object’s textural features are from those of its background. These metrics are designed for broad applicability across datasets and objects and are implemented in a PyTorch-compatible form for efficiency.

We then investigate how tree-likeness and textural separability affect SFM performance through controlled synthetic experiments (Secs. \ref{sec:treelikeness_exp_synth}, \ref{sec:exp_nst}) and real-world datasets (Secs. \ref{sec:treelikeness_exp_real}, \ref{sec:sep_exp_real}). These experiments systematically isolate the influence of object shape and texture while minimizing confounding factors.
Across all datasets, we find that SFM performance (segmentation IoU) noticeably correlates with both tree-likeness and textural separability across diverse object and image types. Notably, HQ-SAM, designed for intricate structures, still exhibits these limitations, highlighting the persistence of these challenges. Furthermore, our fine-tuning experiments (Sec. \ref{sec:finetune}) confirm that these failure modes remain robust to fine-tuning, even when explicitly trained on the most difficult cases. \textbf{Our overall contributions are as follows:}
\begin{enumerate}
    \item We introduce quantitative novel metrics for object tree-likeness---Contour Pixel Rate (CPR) and Difference of Gini Impurity Deviation (DoGD)---and an object textural separability metric based on neural feature contrast.
    \item Through carefully controlled synthetic experiments and real datasets, we demonstrate that the segmentation performance of SAM, SAM 2, and HQ-SAM negatively correlates with tree-likeness and positively correlates with textural separability.
    \item We show that these segmentation challenges persist even after fine-tuning, suggesting that SFMs struggle with these object types regardless of adaptation.
\end{enumerate}

Our findings provide the first quantitative modeling that links SFM performance to measurable object characteristics. Despite the complexity of SFMs and their evaluation datasets, we demonstrate that their underperformance on challenging cases can be effectively modeled using our straightforward and interpretable object descriptors.
%\section*{Related Work: Failure Modes and Biases of Vision Models}
\section{Related Work}

% \paragraph{Failure Modes and Biases of Vision Foundation Models}

% We first note that we focus on ``unintentional'' failure modes of SAM, as opposed to intentional adversarial attacks/examples \cite{zhang2023attack,zheng2023black}.
% A variety of works have found that SAM has trouble segmenting objects with dense tree/branch-like structures (such as retinal blood vessels) \cite{dong_efficient_2024,shi2023generalist,ma2024segment,wan2024trisam} as well as generally irregular objects \cite{ji2024segment}, which was the initial motivation for this work. Additionally, SAM has been found to have poor performance for objects with textures similar to their surroundings (\ie, low contrast), such as camouflaged objects or shadows \cite{chen2023sam,ji2024segment}. In this work, we will introduce metrics that quantify both of these characteristics of objects: tree-likeness, and textural separability from their surroundings.

Beyond the original Segment Anything Model (SAM) \cite{sam} and the subsequent SFMs of HQ-SAM \cite{hqsam} and SAM 2 \cite{sam2}, other works involving universal, generalist, and/or foundation models or architectures for segmentation include OneFormer \cite{jain2023oneformer}, a framework for multi-task universal image segmentation; Segment and Caption Anything (SCA), a method equipping SAM with caption generation \cite{huang2024sca}; Semantic-SAM \cite{li2024semanticsam}, for segmenting with controllable granularity between objects; Open-Vocabulary SAM \cite{yuan2024open}, a framework unifying SAM and CLIP \cite{clip}; and SEEM \cite{zou2023seem}, a less commonly-used multimodal foundation model developed concurrently with SAM.

Previous work on biases and failure modes in computer vision models for tasks such as segmentation has primarily focused on understanding whether networks rely more on object shape or texture for making visual predictions. Zhang et al. \cite{zhang2024convolutional} proposed a framework describing objects by shape, texture, and texture composition within shapes, highlighting how these characteristics impact segmentation tasks. Extensive literature shows that convolutional neural networks (CNNs) tend to be biased toward texture over shape, affecting performance in both classification \cite{geirhosimagenet2019,subramanian2024spatial,hermann2020origins,baker2018deep} and segmentation \cite{zhang2024convolutional} tasks. Notably, CNN performance significantly drops when tested on shape-only images (silhouettes), underscoring texture bias \cite{kubilius2016deep}.

Moreover, neural networks’ shape-learning behaviors vary based on the training dataset and objectives \cite{hermann2020origins,hermann2020shapes}. This variation motivates our detailed investigation into the specific behaviors of SFMs such as the Segment Anything Model (SAM), given their unique training characteristics. While vision transformers (such as SAM’s image encoder architecture) have demonstrated a preference for shape in classification tasks \cite{tuli2021convolutional}, it remains uncertain if this extends to segmentation. Our results suggest SAM predominantly leverages texture cues, sometimes even interpreting complex shapes as textures, corroborating prior studies \cite{zhang2023understanding}.
\section{Methods: SFM Usage and Prompting}
\label{sec:prompting}

We experiment with the ViT-H (default, if unstated) and ViT-B standard pre-trained SAM models \cite{sam} (labeled ``SAM-H'' and ``SAM-B''), the ViT-L and ViT-B+ variants of SAM 2 \cite{sam2} (``SAM 2-L'' and ``SAM 2-B+''), and the ViT-H and ViT-B variants of HQ-SAM \cite{hqsam} (``HQ-SAM-H'' and ``HQ-SAM-B``), a SAM adaptation designed particularly for thin structures. As is the default for SAM, all input images are resized to a resolution of $1024\times 1024$, and normalized to $[0, 255]$. All images that possess instance segmentations for multiple objects
%(\eg, from the iShape \cite{yang2021ishape} and Plittersdorf \cite{haucke2022socrates} datasets) 
will be evaluated by the given SFM segmenting each of the images' objects one at a time (guided by clear prompting described in the next paragraph), with all other mask pixels set as background. In all experiments, we use the oracle-chosen prediction from the given SFM.

For all experiments on the synthetic images of Sec. \ref{sec:treelikeness_exp_synth}, we provide a tight bounding box about the object of interest as the prompt. For all experiments on real images as well as the style-transferred images of Sec. \ref{sec:exp_nst}, we provide the same bounding box as well as a certain number of positive and/or negative point prompts randomly sampled from the object foreground and background \textit{within the bounding box}, respectively, depending on the dataset (full details in Appendix \ref{app:prompting}). We use this relatively heavy prompting strategy to minimize the ambiguity of instructions provided to the given SFM, which will also help to minimize any differences in the oracle prediction from the other two.

\section{The Challenge of Tree-Like Structures}
\subsection{Motivation}
\label{sec:motivation_treelike}

\begin{figure}[htbp!]
    \centering
    \includegraphics[width=0.8\linewidth]{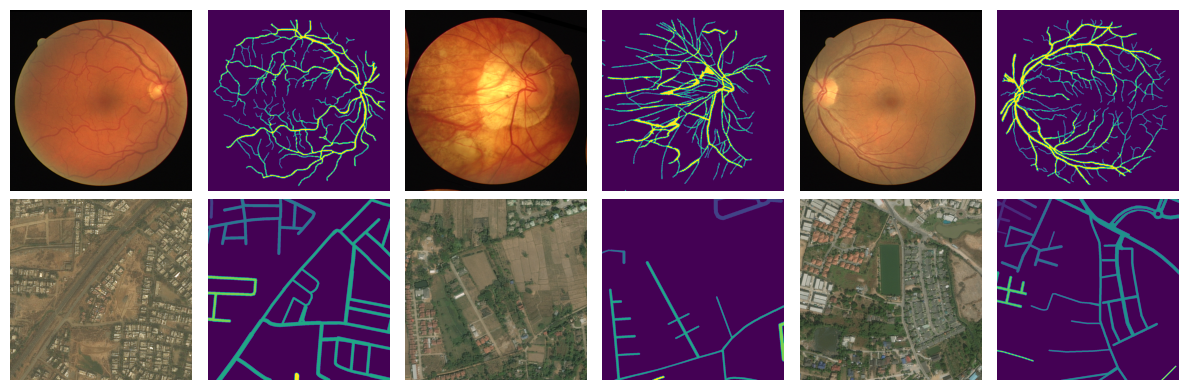}
    \caption{Example retinal blood vessel (top) and satellite road (bottom) images and accompanying object segmentation masks.}
    \label{fig:eg_vesselsroads}
\end{figure}

We first wished to understand the challenges for SFMs in segmenting certain objects with dense, tree-like structures, and then quantify the features driving these challenges, following our observation of a trend of SAM having difficulty with such objects, including retinal blood vessels \cite{qiu2023learnable,ma2024segment} or satellite road images \cite{feng2024road,xu2023leveraging} (see \eg, Fig. \ref{fig:eg_vesselsroads}). Interestingly, while both of these object types have branching features, a characteristic which naively could relate to the SFM's ``failure'', SAM's performance on retinal vessel images is noticeably worse (avg. $\mathrm{IoU}\simeq0.05$ \cite{qiu2023learnable}) compared to on satellite road structures (avg. $\mathrm{IoU}\simeq0.2$ \cite{feng2024road}), which was reproduced in our own experiments (Fig. \ref{figtab:treelikeness_synthetic})\footnote{This could potentially simply be due to retinal vessels being thinner than road structures, but our experiments do not support this (App. \ref{app:thickness}).}.

We hypothesize that objects with dense and irregularly-spaced branching structures--which we refer to as "tree-like"--pose fundamental challenges to SFMs because their structural properties create a mismatch with patch-based attention mechanisms. When a ViT \citep{dosovitskiy2020image} divides an object into fixed-size patches, tree-like structures present a critical issue: their boundary pixels are distributed throughout the object rather than concentrated at the perimeter. Consequently, most patches contain predominantly boundary information, leaving insufficient object-internal features to build coherent representations. In the following section, we will formalize this via our proposed Contour Pixel Rate (CPR) metric. SFM self-attention maps for tree-like objects for SAM then have more fragmented, locally-focused attention patterns that respond to local edges independently rather than forming coherent global object representations, which we show empirically (via the spatial autocorrelation of attention maps) in App. \ref{app:attentioninterp} with examples shown in App. \ref{app:attentionmaps}. Moreover, we empirically show that typical object CPR increases with a model's effective patch size (App. \ref{app:patchsizeinterp}).

Beyond boundary density, tree-like structures also exhibit dramatic appearance variation across different scales: thick trunks, medium branches, and fine tendrils create inconsistent local patterns. We capture this through our Distance of Gini impurity Deviation (DoGD) metric, proposed in the following section, which measures how object pixel distribution varies across scales. High DoGD values indicate scale-dependent spatial inconsistency, preventing stable object representations.

The combination of high CPR and high DoGD explains why SFMs fail on tree-like structures: boundary-dominated patches (high CPR) produce fragmented attention lacking semantic coherence, while scale-dependent variations (high DoGD) prevent consistent object-level representations. This causes SFMs to interpret tree-like objects as textures--collections of local edges without global structure--leading to over-segmentation where unified objects are split into multiple regions. Our metrics thus provide a quantitative framework for predicting SFM failures on complex branching structures.

\subsection{Quantifying Tree-Like Structures}
\label{sec:treelikeness_methods}

\subsubsection{Contour Pixel Rate}
Consider some image $x\in \mathbb{R}^{C\times H\times W}$ with a ``ground truth'' binary segmentation mask for some object within it, $m\in \{0,1\}^{H\times W}$ (in general, this could also be just one class of some multi-class segmentation mask). We first propose to measure the degree of tree-like structure of the object $m$ according to the percentage of the object's pixels which lie on it's \textit{contour}, which are defined as follows. 
\begin{definition}[Contour Pixels]
Given some mask $m$, a pixel $m_{ij}$ is a contour pixel if $m_{ij}=1$ and there exists a different pixel $m_{kl}$ such that $m_{kl} \neq m_{ij}$ and $||(k,l) - (i,j)||_1 < R$, given some smallthreshold $R>0$.
\end{definition}
In other words, contour pixels have at least one pixel of a different class within a small neighborhood. Taking $F$ to be the set of foreground pixels $F:=\{(i,j):m_{ij}=1\}$, the set of contour pixels of $m$ is then $C = \{ (i,j) \in F : \exists (k, l) \in F^c$ \text{ such that } $||(k, l) - (i, j)||_1 < R\}$, which we use to define the \textit{Contour Pixel Rate} (CPR) of the object as
\begin{equation}
    \label{eq:cpr}
    \mathrm{CPR}(m) := \frac{|C|}{|F|}.
\end{equation}
The more tree-like an object is, the higher the percentage of its pixels that are contour pixels, resulting in a higher CPR. We demonstrate fast computation of CPR in full detail in Algorithm \ref{alg:cpr}, via vectorized PyTorch-like pseudocode.

\begin{algorithm}[htbp!]
\caption{Contour Pixel Rate (CPR) of an object (PyTorch-like pseudocode).}
\label{alg:cpr}
\definecolor{codeblue}{rgb}{0.25,0.5,0.5}
\lstset{
  basicstyle=\linespread{1.2}\fontsize{8pt}{8pt}\ttfamily\bfseries,
  commentstyle=\fontsize{8pt}{8pt}\color{codeblue},
  keywordstyle=\fontsize{8pt}{8pt}\color{codeblue},
}
% \small
\textbf{Input}: Object \texttt{\textbf{mask}} ($H\times W$ \texttt{tensor}), contour width threshold \texttt{\textbf{R}} (\texttt{int}).
% \begin{algorithmic}[1] %[1] enables line numbers
%     \STATE \texttt{from skimage.morphology import diamond}
%     \STATE \texttt{neighb\_kernel = diamond(scale)}
%     \STATE \texttt{neighb\_counts = conv2d(mask, neighb\_kernel, padding=r)}
%     \STATE \texttt{contour\_pix = logical\_and(mask>0, neighb\_counts <= (neighb\_kernel.sum()-1))}
%     \STATE \texttt{cpr = contour\_pix.sum()/mask.sum()}
%     \RETURN \texttt{cpr.item()}
% \end{algorithmic}
\begin{lstlisting}[language=python]
from skimage.morphology import diamond

def CPR(mask, R):
    neighb_kernel = diamond(R)
    neighb_counts = conv2d(mask, 
        neighb_kernel, padding=R)
    contour_pix = logical_and(mask>0, 
        neighb_counts <= (neighb_kernel.sum()-1))
    cpr = contour_pix.sum()/mask.sum()
    return cpr.item()
\end{lstlisting}
\end{algorithm}

\subsubsection{Difference of Gini-impurity Deviation (DoGD)} 
We alternatively propose to measure the tree-likeness of some object according to how the variability of object presence across different locations in the image differs between global and local scales. Intuitively, irregularly-spaced tree-like structures have high variability at small scales due to alternating frequently between areas of mixed and uniform pixel classes, but low variability at large scales due to the repetitive nature of the structure becoming more homogeneous, which we propose to quantify as follows.

First, we quantify the object presence within some $k\times k$ square window of the mask anchored at some coordinates $h_0 < H$, $w_0 < W$ via the \textit{Gini impurity}: 
\begin{equation}
    \label{eq:fullgini}
    \mathrm{Gini}(m; k, h_0, w_0) := 1 - {\sum}_j[p_j(\mathcal{W}^k_{h_0,w_0}(m))]^2
\end{equation}
where we denote the $k\times k$ square window of the mask anchored at $h_0,w_0$ as $\mathcal{W}^k_{h_0,w_0}(m):= m[h_0: h_0+k, w_0: w_0+k]$. Here, $p_j(\mathcal{W}^k_{h_0,w_0}(m))$ denotes the probability of the object of class $j$ being in some pixel within the window, which is computed simply as $n_j/k^2$, where $n_j$ is the number of pixels in the window of class $j$. In our binary segmentation case, the Gini impurity simplifies to
\begin{align}
    &\mathrm{Gini}(m; k, h_0, w_0) := \\ \nonumber
    & 1 - [p(\mathcal{W}^k_{h_0,w_0}(m))]^2 - [1- p(\mathcal{W}^k_{h_0,w_0}(m))]^2
\end{align}

where we write $p:= p_1$. The Gini impurity measures the degree of uncertainty (ranging from $0$ to $0.5$) of whether an object is present in the given window. For example, mask windows containing pixels of mostly one class will have $\mathrm{Gini}\simeq0$, while having similar pixel amounts of both classes will result in $\mathrm{Gini}\simeq0.5$\footnote{Following common decision tree practices, we use Gini impurity over entropy because it is a symmetric and faster alternative \cite{breiman1984classification}.}.

Next, we compute the variability of object presence at a given scale/window size across the entire mask by sampling all possible $k\times k$ windows with anchors $(h_0,w_0)$, and computing the standard deviation of the Gini impurity across these windows, as
\begin{equation}
    \sigma^\mathrm{Gini}_k(m) := \sqrt{\mathrm{Var}_{h_0,w_0}[\mathrm{Gini}(m; k, h_0, w_0)]}.
\end{equation}
Finally, we define the Difference of Gini Impurity Deviation (\textbf{DoGD}) between global and local scales as
\begin{equation}
    \label{eq:dogd}
    \mathrm{DoGD}(m) := \sigma^\mathrm{Gini}_a(m) - \sigma^\mathrm{Gini}_b(m),
\end{equation}
where the global and local window sizes $k=a$ and $k=b$ are chosen such that $a\gg b$. We present DoGD in optimized PyTorch-like form in Algorithm \ref{alg:dogd}.

\begin{algorithm}[htbp!]
\caption{Difference of Gini Impurity Deviation (DoGD) of an object (PyTorch-like pseudocode).}
\label{alg:dogd}
\definecolor{codeblue}{rgb}{0.25,0.5,0.5}
\lstset{
  basicstyle=\linespread{1.2}\fontsize{8pt}{8pt}\ttfamily\bfseries,
  commentstyle=\fontsize{8pt}{8pt}\color{codeblue},
  keywordstyle=\fontsize{8pt}{8pt}\color{codeblue},
}
\textbf{Input}: Object \texttt{\textbf{mask}} ($H\times W$ \texttt{tensor}), global and local window sizes \texttt{\textbf{a}, \textbf{b}}  (\texttt{int}s).
%\begin{algorithmic}[1] %[1] enables line numbers
%    \STATE \texttt{gini\_std = \{\}}
%    \FOR{\texttt{k} in \texttt{a,b}}
%        \STATE \texttt{avg\_kernel = ones(k,k)}
%        \STATE \texttt{p = conv2d(mask, avg\_kernel)}
%        \STATE \texttt{gini = 1 - p**2 - (1-p)**2}
%        \STATE \texttt{gini\_std[k] = gini.std().item()}
%    \ENDFOR
%    \RETURN \texttt{gini\_std[a] - gini\_std[b]}
%\end{algorithmic}
\begin{lstlisting}[language=python]
def DoGD(mask, a, b):
    gini_std = {}
    for k in [a,b]:
        avg_kernel = ones(k,k)
        p = conv2d(mask, avg_kernel)
        gini = 1 - p**2. - (1-p)**2.
        gini_std[k] = gini.std().item()
    return gini_std[a] - gini_std[b]
\end{lstlisting}
\end{algorithm}

Objects with significant tree-like or fractal-like structure will exhibit relatively large values of $\sigma^\mathrm{Gini}_b(m)$ due to high variability in pixel composition at small scales (frequently alternating between areas of mixed classes and areas of a single class), yet small $\sigma^\mathrm{Gini}_a(m)$ due to structures with high uniformity and/or repetitions at large scales, altogether increasing the DoGD.

We perform all experiments with the hyperparameters for CPR and DoGD set to $R=5$, $a=127$, and $b=3$, which we found via grid search for the values that resulted in the Kendall's $\tau$ between IoU and DoGD with the lowest $p$-value on the held-out DIS training set using SAM-H. We show results using a wide range of other values for these hyperparameters in App. \ref{app:treelikeness_exp}, where our findings were consistent for most other settings. Moreover, we note that CPR and DoGD are noticeably correlated (Pearson $|r|=0.83$ on average; App. \ref{app:treelikeness_corr}), showing their consistency in quantifying different aspects of tree-likeness.

\subsection{The Relationship between Object Tree-likeness and Segmentation Performance}
\label{sec:treelikeness_exp}
% In this section, we will explore the effect of tree-likeness of objects as measured by CPR or DoGD on SAM's segmentation performance. We begin with experiments on carefully controlled synthetic data, followed by real data.

\subsubsection{Experiments on Synthetic Data}
\label{sec:treelikeness_exp_synth}
\begin{figure}[htbp!]
    \centering
    \includegraphics[width=0.99\linewidth]{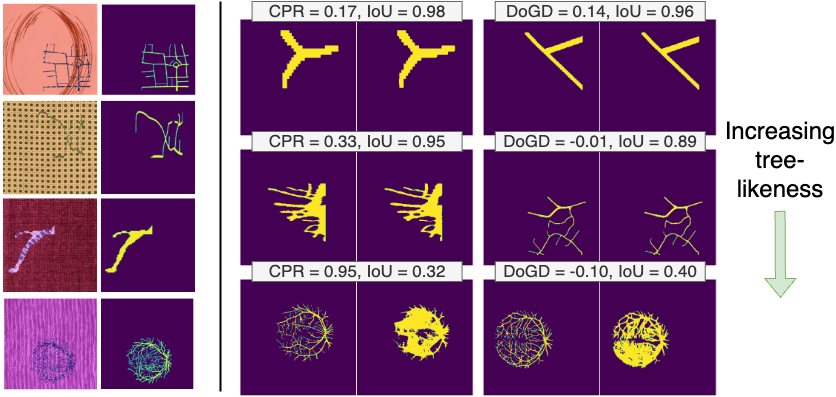}
    \caption{\textbf{Left:} Example synthetic tree-like images and object masks. \textbf{Right:} Trend of increasing tree-likeness (increasing CPR/decreasing DoGD) of these objects (left in each pair) resulting in worse SAM segmentation predictions (right in each pair)}
    \label{fig:eg_synthetic_treelike}
\end{figure}

\begin{figure}[htbp!]
    \begin{minipage}[l]{0.18\textwidth}
        \centering
        \includegraphics[width=0.99\linewidth]{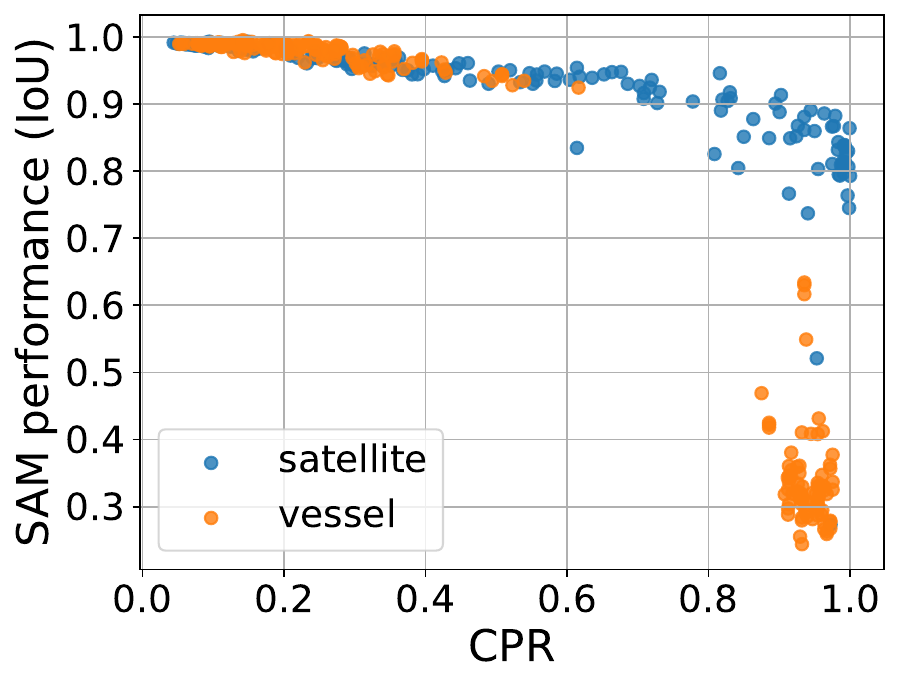}
        \includegraphics[width=0.99\linewidth]{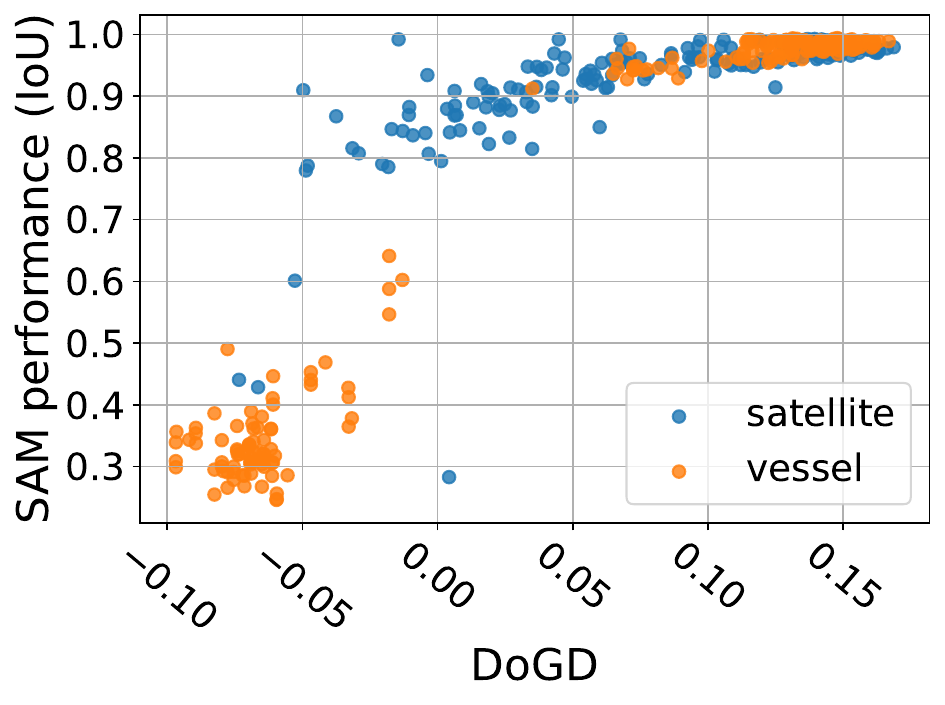}
    \end{minipage}
    \hfill
    \begin{minipage}[l]{0.29\textwidth}
        \centering
        \setlength{\tabcolsep}{2pt}
        \fontsize{7pt}{7pt}\selectfont
        % \small
        % \scriptsize
        \begin{tabular}{l|cc|cc}
        \multicolumn{1}{c}{} & \multicolumn{2}{c|}{\textbf{CPR}} &  \multicolumn{2}{c}{\textbf{DoGD}}  \\
        \toprule
        % breast brain lumbar abdom
        \textbf{Model} & $\tau$ & $\rho$ & $\tau$ & $\rho$ \\
        \midrule
        SAM-H & -0.77 & -0.93 & 0.61 & 0.80 \\ 
        SAM-B & -0.75 & -0.93 & 0.66 & 0.84 \\
        SAM 2-L & -0.73 & -0.91 & 0.67 & 0.85 \\
        SAM 2-B+ & -0.71 & -0.90 & 0.69 & 0.86 \\
        HQ-SAM-H & -0.74 & -0.92 & 0.70 & 0.87 \\
        HQ-SAM-B & -0.81 & -0.95 & 0.66 & 0.84 \\ \midrule 
        \textbf{Average} & \textbf{-0.75} & \textbf{-0.92} & \textbf{0.67} & \textbf{0.84} \\
        \bottomrule
        \end{tabular}
    \end{minipage}
\caption{\textbf{Left:} SFM prediction IoU vs. object tree-likeness (CPR and DoGD), on the synthetic dataset, shown for SAM-H. \textbf{Right:} Rank correlations between IoU and tree-likeness for all SFMs.}
\label{figtab:treelikeness_synthetic}
\end{figure}

In this section, we will first probe the effect of object tree-likeness on the performance of SFMs by testing them on synthetic images that solely possess objects of varying tree-likeness, with different independently chosen, uniform foreground and background textures. These objects are contiguous components sampled from retinal blood vessel and satellite road masks; the full procedure of generating these images is detailed in App. \ref{app:treelikesynth_creation}. Example generated images, masks, and SAM predictions for them are shown in Fig. \ref{fig:eg_synthetic_treelike}. As shown (also in Fig. \ref{figtab:treelikeness_synthetic}, left), the objects cover a wide range of tree-likeness as measured by these quantities.
%We do so via an algorithm that randomly places contiguousretinal blood vessel or satellite road object masks of fixed size within  These objects have a range of tree-likeness, and are designed to have high contrast with their background, in order to disentangle any effects from the phenomena studied in the later Section \ref{sec:sep} (the separability of an object's texture from its surroundings).

In order to mitigate any confounding effects on a given SFM's performance due to an object's textural contrast from its surroundings (which we study in Sec. \ref{sec:sep}), for each mask $m_c$ generated by our procedure, we apply the SFM to $K=7$ images created by applying $K$ different randomly-sampled pairs of textures to the object's foreground ($m_c=1$) and background ($m_c=0$). We then obtain the SFM's final prediction for this object via pixel-wise majority voting over its predictions on these $K$ images.

In Fig. \ref{figtab:treelikeness_synthetic} left, we show the relationship between an object's tree-likeness (via CPR or DoGD) with SAM's performance (IoU) in segmenting the object, for all generated synthetic objects. We quantify the strength of this relation via non-linear correlation, measured by Kendall's tau ($\tau$) \cite{kendall1938new} and Spearman's rho ($\rho$) \cite{spearman_proof_1904}, for all three SFMs, shown in the table to the right of the plots.
%\footnote{We evaluate $\tau$ in addition to $\rho$ due to it being more robust to outliers.}. 
We see that object tree-likeness is quite predictive of an SFM's performance, with average absolute correlations of $|\tau| = 0.75$ and $|\rho| = 0.92$ for CPR and $|\tau| = 0.67$ and $|\rho| = 0.84$ for DoGD; \ie, more prevalent dense tree-like structures corresponding to worse performance. In the following section, we will show that this finding is also present for real data, which is subject to noise from a variety of uncontrollable factors.

% \begin{table}[htbp]
% \setlength{\tabcolsep}{4pt}
% \centering
% \fontsize{7pt}{7pt}\selectfont
% % \small
% % \scriptsize
% \begin{tabular}{l|cc|cc}
% \multicolumn{1}{c}{} & \multicolumn{2}{c|}{\textbf{CPR}} &  \multicolumn{2}{c}{\textbf{DoGD}}  \\
% \toprule
% % breast brain lumbar abdom
% \textbf{Model} & $\tau$ & $\rho$ & $\tau$ & $\rho$ \\
% \midrule
% SAM-H & -0.77 & -0.93 & 0.61 & 0.80 \\ 
% SAM-B & -0.75 & -0.93 & 0.66 & 0.84 \\
% SAM 2-L & -0.73 & -0.91 & 0.67 & 0.85 \\
% SAM 2-B+ & -0.71 & -0.90 & 0.69 & 0.86 \\
% HQ-SAM-H & -0.74 & -0.92 & 0.70 & 0.87 \\
% HQ-SAM-B & -0.81 & -0.95 & 0.66 & 0.84 \\ \midrule 
% \textbf{Average} & \textbf{-0.75} & \textbf{-0.92} & \textbf{0.67} & \textbf{0.84} \\
% \bottomrule
% \end{tabular}
% \end{table}

\subsubsection{Experiments on Real Data}
\label{sec:treelikeness_exp_real}
We will now perform the same analysis on three real datasets which contain objects which cover a spectrum of tree-likeness, DIS5k, MOSE, and iShape.
DIS5k \cite{DIS} (or ``DIS'') is a dataset containing extremely detailed annotations of objects with varying degrees of hollowness, and both regular and irregular tree-like structures. All DIS experiments will be reported on its validation set.  We show example images with objects with varying degrees of tree-likeness (by CPR) in Fig. \ref{fig:teaser} left. We additionally performed experiments on the MOSE dataset \citep{MOSE} of videos of objects in complex, cluttered, and/or camouflaged scenes, randomly sampled 4 frames with non-empty labels from each of the 1,507 videos in the training set, resulting in $\sim$6,000 frames with $\sim$13,000 total objects.

iShape \cite{yang2021ishape} consists of six sub-datasets of real and realistic-appearing synthetic images for instance segmentation of different objects: antenna, branches, fences, logs, hangers, and wires (see \eg Fig. \ref{fig:eg_plitters}). We analyze these classes individually to mitigate potential confounding/noise factors due to inter-class variations. %\footnote{We do not do this for DIS due to the larger number of classes which are much more fine-grained in their differences, additionally because there are few images per class for DIS.

% \begin{figure}[htbp!]
%     \centering
%     \includegraphics[width=0.99\linewidth]{eg_realimages.png}
%     \caption{Example images and object masks from DIS (left group) and iShape (right group).}
%     \label{fig:eg_DISiShape}
% \end{figure}

% While the synthetic experiments had objects with perfect textural contrast to the background (allowing us to disentangle shape-based features from textural features), this dataset has objects which may not satisfy this. 

\begin{figure*}
    \begin{minipage}[l]{0.35\textwidth}
    \centering
    % \begin{tabular}{cc}
    %     \includegraphics[width=0.49\linewidth]{figs/r=5_avg_vit_h.pdf} & \vspace{-1cm}\includegraphics[width=0.49\linewidth]{figs/contour_width=5_VIT-H.pdf}  \\
    %      & 
    % \end{tabular}
    \includegraphics[width=0.49\linewidth, trim={0 0 0 0}]{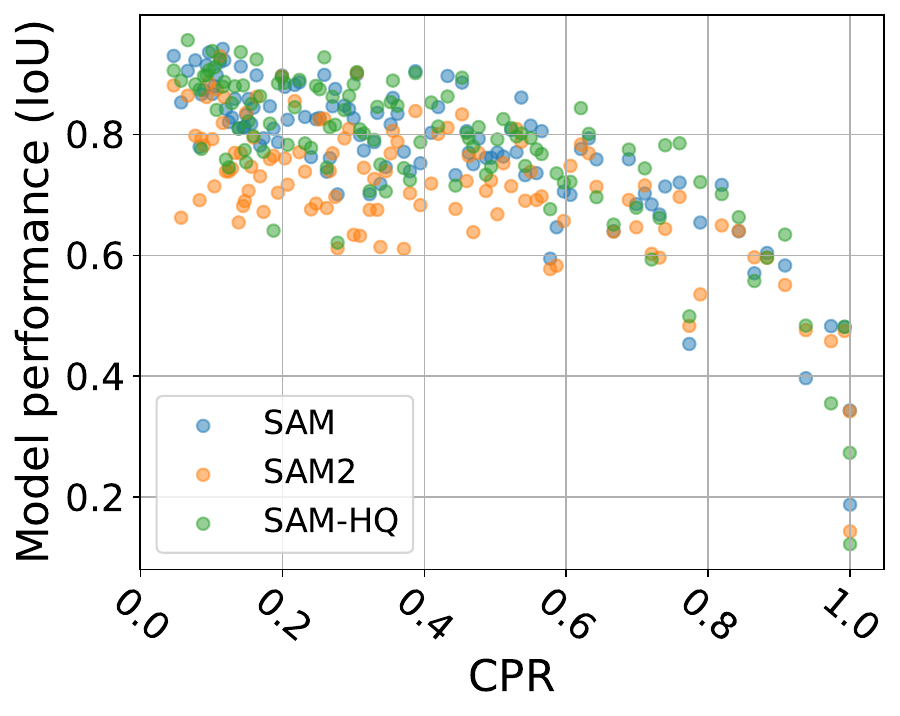}
    \includegraphics[width=0.49\linewidth, trim={0.5cm -0.5cm -0.5cm 0}]{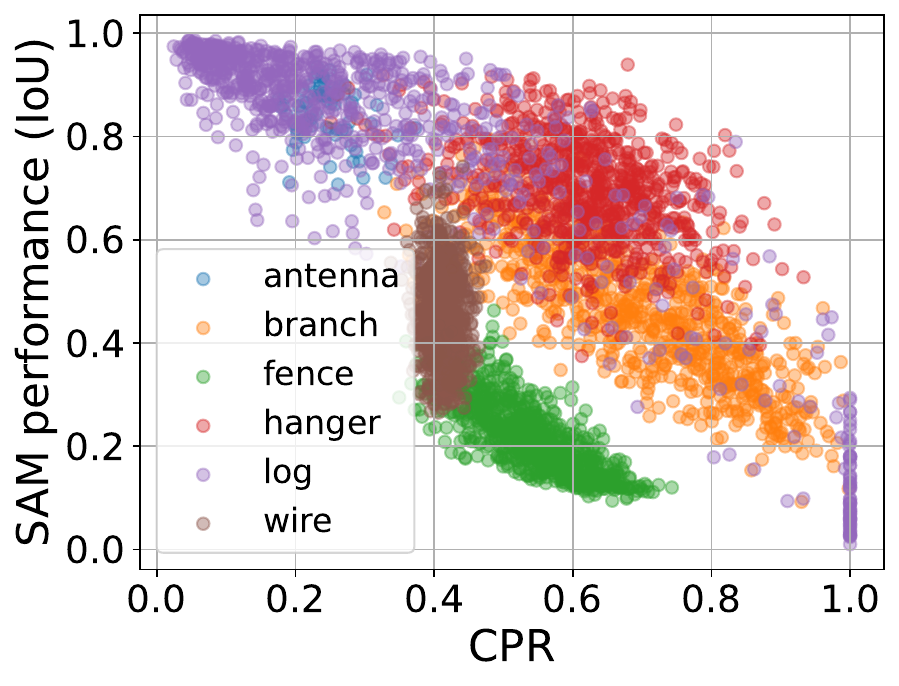}
    \includegraphics[width=0.49\linewidth]{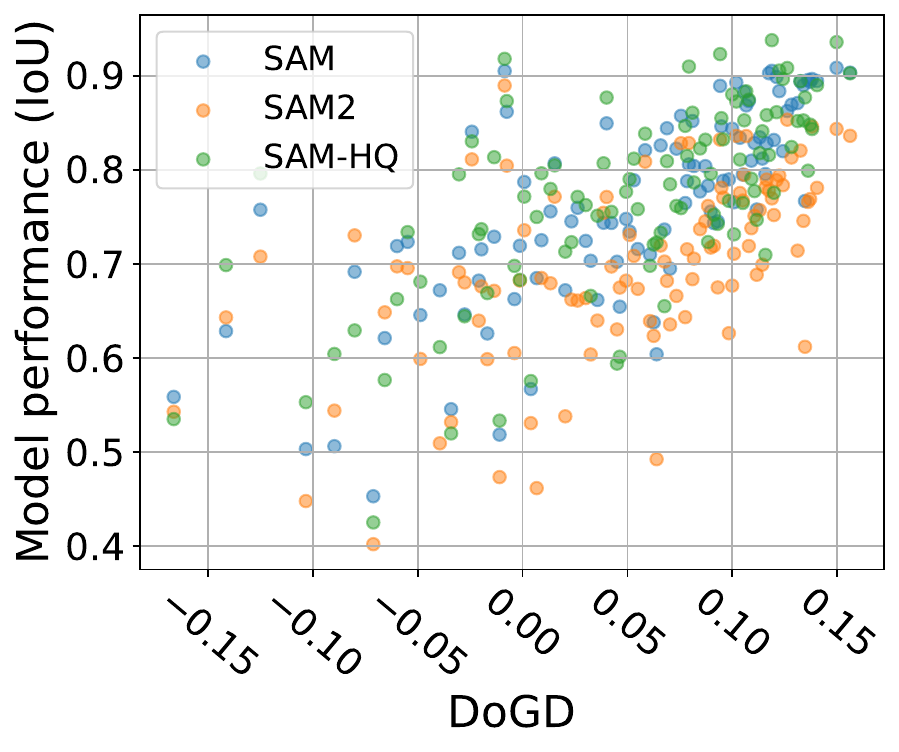}
    \includegraphics[width=0.49\linewidth, trim={0.5cm 0 0 0}]{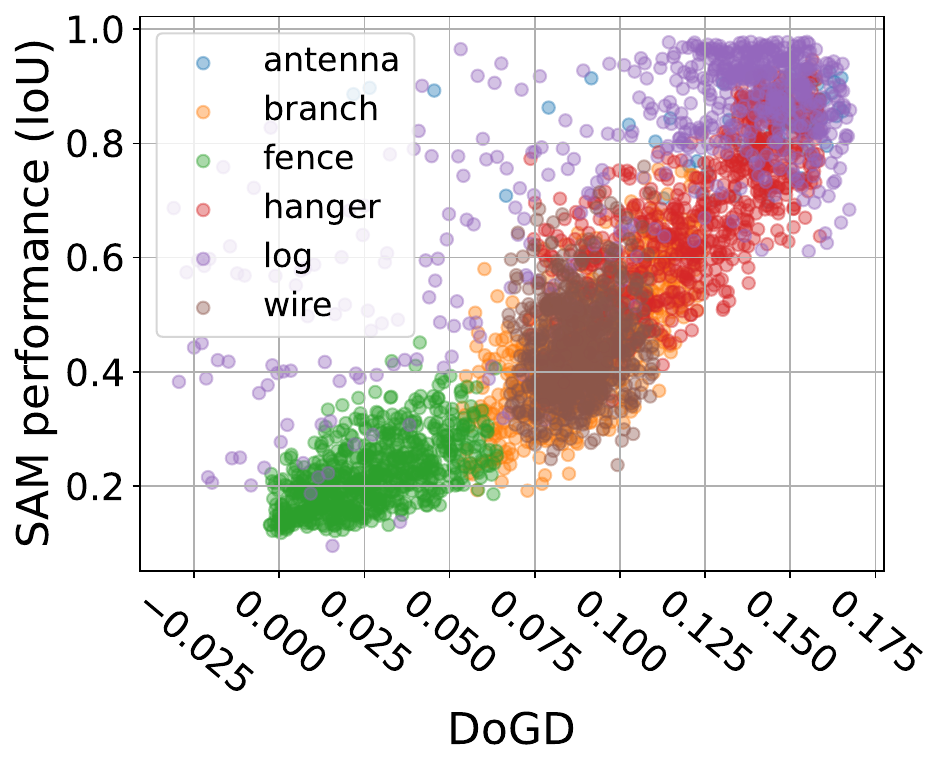}
    % \caption{Segmentation IoU vs. object tree-likeness (CPR, \textbf{top}; DoGD, \textbf{bottom}) on DIS (\textbf{left}) for all three SFMs and iShape (\textbf{right}) shown for SAM.}
    % \label{figtab:treelikeness_results_real}
    \end{minipage}
    \hfill
    \begin{minipage}[t]{0.65\textwidth}
    \setlength{\tabcolsep}{1.5pt}
    \centering
    \fontsize{6.5pt}{6.5pt}\selectfont
    % \small
    % \scriptsize
    \begin{tabular}{l|l|cc||cc||cc|cc|cc|cc|cc|cc}
    \multicolumn{2}{c}{} & \multicolumn{2}{c}{} & \multicolumn{2}{c}{} & \multicolumn{12}{c}{\textbf{iShape}}\\
    \multicolumn{2}{c}{} &  \multicolumn{2}{c||}{\textbf{DIS}} & \multicolumn{2}{c||}{\textbf{MOSE}} & \multicolumn{2}{c|}{antenna} & \multicolumn{2}{c|}{branch} & \multicolumn{2}{c|}{fence} & \multicolumn{2}{c|}{hanger} & \multicolumn{2}{c|}{log} & \multicolumn{2}{c}{wire}\\
    \toprule
    
     & \textbf{Model} & $\tau$ & $\rho$ & $\tau$ & $\rho$ & $\tau$ & $\rho$ & $\tau$ & $\rho$ & $\tau$ & $\rho$ & $\tau$ & $\rho$ & $\tau$ & $\rho$ & $\tau$ & $\rho$ \\
    \midrule
    \multirow{6}{*}{\rotatebox{90}{\textbf{CPR:}}} & SAM-H & -0.59 & -0.76 & -0.80 & -0.95 & -0.13 & -0.21 & -0.61 & -0.81 & -0.61 & -0.81 & -0.25 & -0.37 & -0.60 & -0.77 & -0.06 & -0.10 \\
     & SAM-B & -0.63 & -0.81 & -0.76 & -0.93 & -0.18 & -0.27 & -0.63 & -0.82 & -0.63 & -0.83 & -0.44 & -0.62 & -0.62 & -0.80 & -0.12 & -0.17 \\
    & SAM 2-L & -0.41 & -0.57 & -0.84 & -0.96 & -0.30   & -0.43 & -0.64   & -0.84 & -0.61   & -0.81 & -0.48   & -0.66 & -0.67   & -0.85 & -0.14   & -0.21 \\
    & SAM 2-B+ & -0.38 & -0.52 & -0.81 & -0.95 & -0.16   & -0.23 & -0.52   & -0.71 & -0.57   & -0.78 & -0.42   & -0.59 & -0.60   & -0.77 & -0.12   & -0.18 \\
    & HQ-SAM-H & -0.52 & -0.69 & -0.79 & -0.94 & -0.23   & -0.33 & -0.77   & -0.93 & -0.79   & -0.95 & -0.51   & -0.70 & -0.69   & -0.85 & -0.22   & -0.33 \\
    & HQ-SAM-B & -0.62 & -0.81  & -0.76 & -0.92 & -0.25   & -0.34 & -0.75   & -0.92 & -0.83   & -0.96 & -0.50   & -0.69 & -0.71   & -0.87 & -0.24   & -0.35 \\\midrule
    & \textbf{Average} & \textbf{-0.53} & \textbf{-0.69} & \textbf{-0.79} & \textbf{-0.94} & \textbf{-0.21} & \textbf{-0.30} & \textbf{-0.65} & \textbf{-0.84} & \textbf{-0.67} & \textbf{-0.86} & \textbf{-0.43} & \textbf{-0.61} & \textbf{-0.65} & \textbf{-0.82} & \textbf{-0.15} & \textbf{-0.22} \\
    
    \midrule
    \multirow{6}{*}{\rotatebox{90}{\textbf{DoGD:}}} & SAM-H & 0.58 & 0.77 & 0.43 & 0.60 & 0.23 & 0.30 & 0.52 & 0.70 & 0.46 & 0.65 & 0.52 & 0.72 & 0.30 & 0.46 & 0.08 & 0.11 \\
     & SAM-B & 0.45 & 0.64 & 0.27 & 0.39 & 0.12 & 0.14 & 0.51 & 0.69 & 0.45 & 0.64 & 0.50 & 0.70 & 0.13 & 0.22 & 0.06 & 0.08 \\
    & SAM 2-L & 0.44 & 0.60 & 0.52 & 0.70 & 0.42    & 0.59  & 0.48    & 0.65  & 0.49    & 0.68  & -0.05   & -0.08 & 0.56    & 0.75  & -0.03   & -0.05 \\
    & SAM 2-B+ & 0.26 & 0.39 & 0.47 & 0.66 & 0.40    & 0.56  & 0.29    & 0.40  & 0.30    & 0.44  & -0.06   & -0.10 & 0.42    & 0.58  & -0.08   & -0.12 \\
    & HQ-SAM-H & 0.48 & 0.66 & 0.41 & 0.58 & 0.53    & 0.72  & 0.58    & 0.72  & 0.48    & 0.67  & -0.31   & -0.45 & 0.57    & 0.75  & -0.13   & -0.19 \\
    & HQ-SAM-B & 0.50 & 0.69 & 0.33 & 0.48 & 0.67    & 0.85  & 0.59    & 0.73  & 0.57    & 0.77  & -0.25   & -0.34 & 0.53    & 0.71  & -0.04   & -0.06 \\\midrule
    & \textbf{Average} & \textbf{0.45} & \textbf{0.63}  & \textbf{0.41} & \textbf{0.57} & \textbf{0.40} & \textbf{0.53} & \textbf{0.50} & \textbf{0.65} & \textbf{0.46} & \textbf{0.64} & \textbf{0.06} & \textbf{0.08} & \textbf{0.42} & \textbf{0.58} & \textbf{-0.02} & \textbf{-0.04} \\
    \bottomrule
\end{tabular}
    \end{minipage}
    \caption{\textbf{Left:} Segmentation IoU vs. object tree-likeness (CPR, \textbf{top}; DoGD, \textbf{bottom}) on DIS (\textbf{left}) for all three SFMs and iShape (\textbf{right}) shown for SAM-H. \textbf{Right:} rank correlations between IoU and tree-likeness for all SFMs.}
    \label{figtab:treelikeness_results_real}
\end{figure*}

In Fig. \ref{figtab:treelikeness_results_real} left, we show how SAM's performance (IoU) on these images relates to the tree-likeness of the objects which it is segmenting, alongside the other SFMs (SAM 2 and HQ-SAM) for comparison for DIS, with accompanying quantitative correlation results for all three SFMs shown in the table to the right of the plots. The plots for MOSE results are provided in App. \ref{app:MOSEallresults} due to space constraints.
In order to reduce the noise incurred by the large variety of segmented objects in the dataset and nuisance confounding factors in the images, we analyze results with \textit{aggregated objects}: we cluster groups of $5$ objects/images ($20$ for MOSE due to its size) with similar IoU and tree-likeness (CPR or DoGD) into single datapoints of the average value of these metrics, and similar for the IoU vs. textural separability experiments of Sec. \ref{sec:sep}.

We see that despite the many potential confounding factors in real data, there is still a clear correlation between object tree-likeness as measured by the proposed metrics and SFM segmentation performance. In particular, we see average absolute correlations of $|\tau| = 0.62$ and $|\rho| = 0.79$ for CPR (particularly high for MOSE) and $|\tau| = 0.38$ and $|\rho| = 0.53$ for DoGD---excluding the antenna and wire objects of iShape which had outlying correlations likely because those two object classes cover only small ranges of tree-likeness as opposed to the other iShape classes (Fig. \ref{figtab:treelikeness_results_real}, left), such that noise from other confounding factors obscures any dependence of performance on tree-likeness.

% In general, we see that while the correlation between tree-likeness and segmentation performance is noticeable, the relationship is noisier than what was seen in the synthetic data experiments (Table \ref{tab:treelikeness_synthetic}). This is because while the synthetic images were designed to have objects with perfect textural contrast to their background, allowing us to disentangle shape-based features from textural features, real datasets have objects which may not satisfy this, and so noise due to confounding factors is unavoidable.

\section{The Challenge of Textural Separability}
\label{sec:sep}

In the previous section, we demonstrated that SFMs struggle with segmenting non-conventional shapes, in particular, dense tree-like structures, which we hypothesize is due to the model confusing the dense structure as the \textit{texture} of a non-treelike, more regular shape, rather than a shape itself (see \eg Fig. \ref{fig:teaser} left). Similar to this behavior is that even for objects with simpler shapes, these SFMs can still be confused if the object's texture is even somewhat similar to its surroundings, which we will study in this section.

% In particular, we find that SAM's performance is affected by the contrast between an object's texture(s) and the surrounding background texture(s) supporting, prior empirical results of poor performance on low-contrast examples \cite{chen2023sam,ji2024segment}. In this section, we will formalize this study, first by defining textural separability and how to measure it, and then showing how it affects performance on both real data and synthetic data with carefully controlled textural characteristics.

\subsection{Measuring Textural Separability}
\label{sec:sep_methods}

We will define the textural contrast or \textit{separability} between some object mask and its surroundings 
% not in terms of conventional measures of hue, saturation, brightness, etc., but by the more abstract definition of 
by \textit{how easily their textures can be distinguished from one another}. Motivated by findings that early layers of classification-pretrained CNNs primarily capture low-level features involving edges and textures \cite{zeiler2014visualizing}, we characterize an image's textures using the first convolutional layer of a ResNet-18 \cite{he2016resnet} pretrained on ImageNet \cite{deng2009imagenet}. Denoted by $f_1 : \mathbb{R}^{C\times H\times W} \rightarrow \mathbb{R}^{C'\times H'\times W'}$, this outputs a textural feature map corresponding to all $7 \times 7$ windows in the image.
%, with each element (activation) corresponding to a center pixel. 

We then measure the textural separability of an object according to if a simple classifier $g$ can be trained to discriminate between (a) the activations for the pixels of the object foreground and (b) the activations right outside of the object boundary.
% \footnote{This can be considered as probing for the \textit{concept} of foreground vs. background texture, but at the activation level rather than the image level as in \eg, \citet{kim2018interpretability}.}.
% \footnote{This procedure is somewhat similar to the probing of hidden activation concepts \cite{alain2017understanding,kim2018interpretability}, although these works detected concepts (such as textures) at the image level, not at the single activation level as we do here.}. 
We define this in Algorithm \ref{alg:textural_sep}, where $\mathrm{Dilate}$ and $\mathrm{Disk}$ refer to \texttt{scikit-image.morphology} functions \cite{scikit-image}. We use simple logistic regression for $g$ with regularization parameter $C=2$ \cite{scikit-learn}. We also evaluate other hyperparameter settings, as well as a random forest classifier for $g$, in App. \ref{app:sep_exp}, where we found similar results.

\begin{algorithm}[htbp!]
    \caption{Textural separability of an object.}
    \label{alg:textural_sep}
    \small
\begin{algorithmic}[1] %[1] enables line numbers
    \REQUIRE{Image $x$ with binary mask $m$, first convolutional layer of pre-trained CNN $f_1$, simple classifier $g$.}
    % \REQUIRE{Image $x\in \mathbb{R}^{C\times H \times W}$ with mask $m\in \{0,1\}^{H \times W}$, first convolutional layer of pretrained CNN $f_1$, simple classifier $g$.}
    \STATE $h = f_1(x) \in \mathbb{R}^{C'\times H'\times W'}$ \textit{(Get textural features.)}
    \STATE $m = \mathrm{Resize}(m)$ to $H'\times W'$ \textit{via NN interpolation.}
    \STATE $m' := \mathrm{Dilate}(m, \mathrm{Disk}(5))$ \textit{(Expand boundary.)}
    \STATE $m' = m' - m$ \textit{(Extract boundary only.)}
    \STATE $h_\mathrm{obj} := h[:, m\neq 0]$, $h_\mathrm{bdry} := h[:, m'\neq 0]$
    \STATE \textit{Train $g$ to classify between activations in $h_\mathrm{obj}$ vs. $h_\mathrm{bdry}$.}
    \STATE \textbf{Return} \textit{Test classification accuracy of $g$.}
\end{algorithmic}
\end{algorithm}

% With this procedure, an object with low textural contrast with its surroundings will result in textural activations that are similar on both the object and its boundary, giving a low classification accuracy, and vice-versa for an object with high contrast, resulting in the classifier being able to more easily differentiate it from its background. 

\subsection{The Effect of Textural Separability on Segmentation Performance}
\label{sec:sep_exp}

\subsubsection{Style Transfer for Controlled Textural Contrast}
\label{sec:exp_nst}

To isolate the effects of textural separability from object shape, we first conduct controlled experiments using neural style transfer (NST) \cite{gatys2016image}, before evaluating on real data in the next section. This allows us to systematically adjust an object’s textural contrast with its background while keeping other factors controlled.

We begin with all objects sampled from VOC2012 \cite{pascal-voc-2012} that occupy 5–25\% of their image area, to ensure sufficient background context (484 objects total). Using a Stable Diffusion-based inpainting model (App. \ref{app:NSTmodels}), we remove each object and reconstruct its background, creating pairs of separate object-only and background-only images. This setup allows precise modifications before reinserting objects into their original scenes.
We then create three types of composite images with varying degrees of changes to object shape for each object (without modifying the background):
\begin{enumerate}
    \item \textbf{Controlled:} the object is not modified. %; it is inserted into the inpainted background at its original location.
    \item \textbf{Altered:} the object undergoes significant non-affine geometric transformations, modifying its shape and distorting its texture (App. \ref{app:NSTmodels}).
    \item \textbf{Mixed:} a hybrid of the original and altered object, using altered pixels where the masks overlap and original pixels elsewhere.
\end{enumerate}

\begin{figure}[htbp]
     \centering
     \includegraphics[width=.69\linewidth]{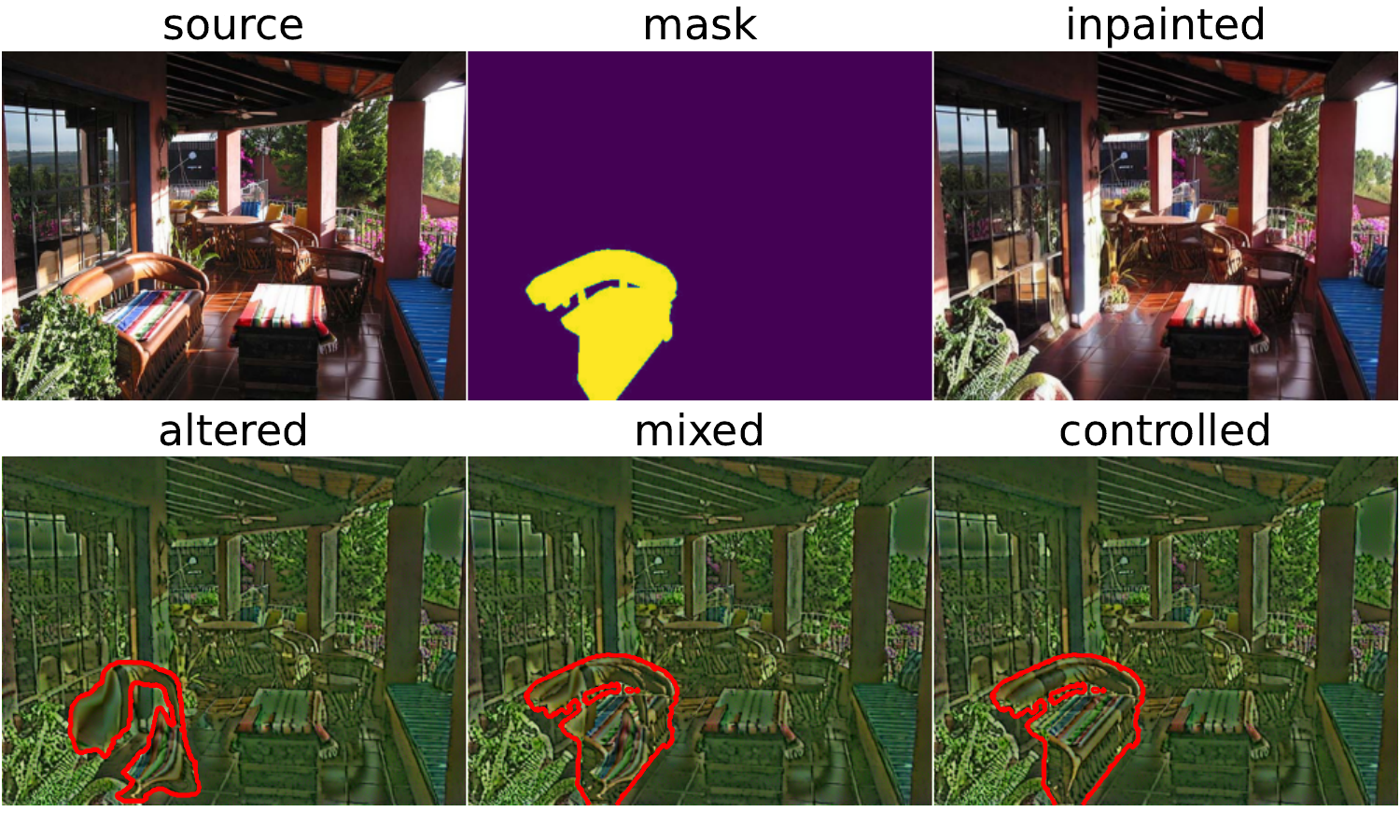}
     \caption{Example from the NST image creation pipeline.}
    \label{fig:NST_example}
\end{figure}

After creating these image variants, we apply NST (similar to \cite{geirhosimagenet2019}) using textures randomly sampled from Colored Brodatz \cite{abdelmounaime2013newMBT} with the Gatys et al. NST model \cite{gatys2016image} (details in App. \ref{app:NSTmodels}). An example of this entire procedure is in Fig. \ref{fig:NST_example}.

\begin{figure}[htbp]
     \centering
     \includegraphics[width=.79\linewidth]{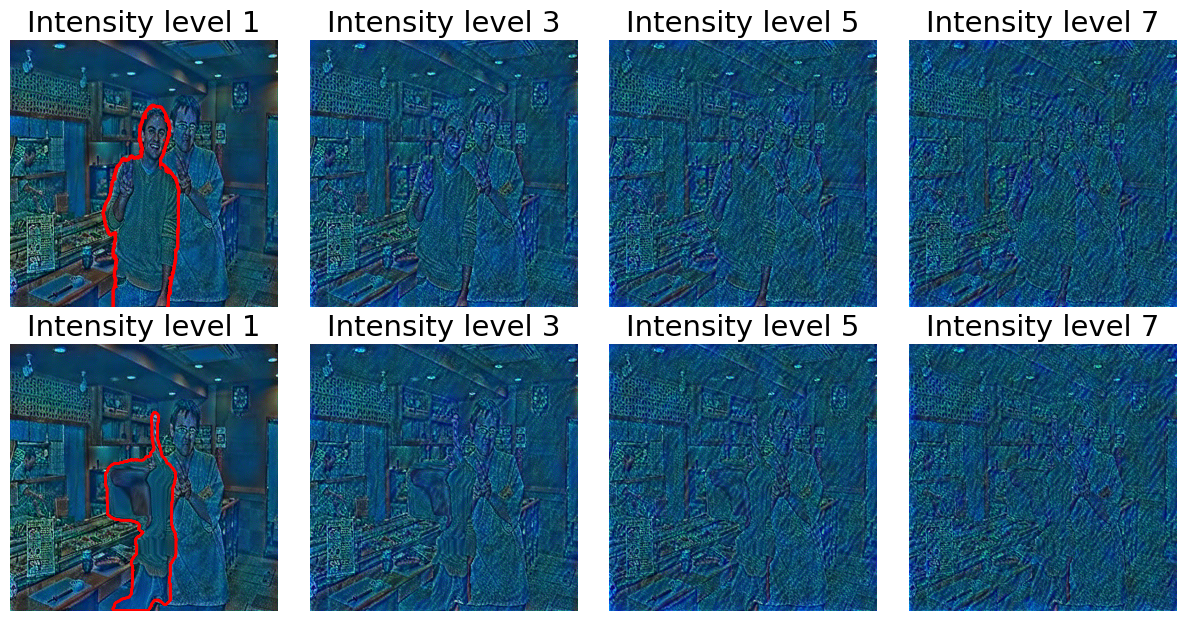}
     \caption{Using NST with different intensities to adjust the textural separability of an object (highlighted in red), on \textit{controlled} (upper row) and \textit{altered} (lower row) versions of the object.}
    \label{fig:NST_intensity}
\end{figure}

\begin{figure}[htbp]
     \centering
     \includegraphics[width=.45\linewidth]{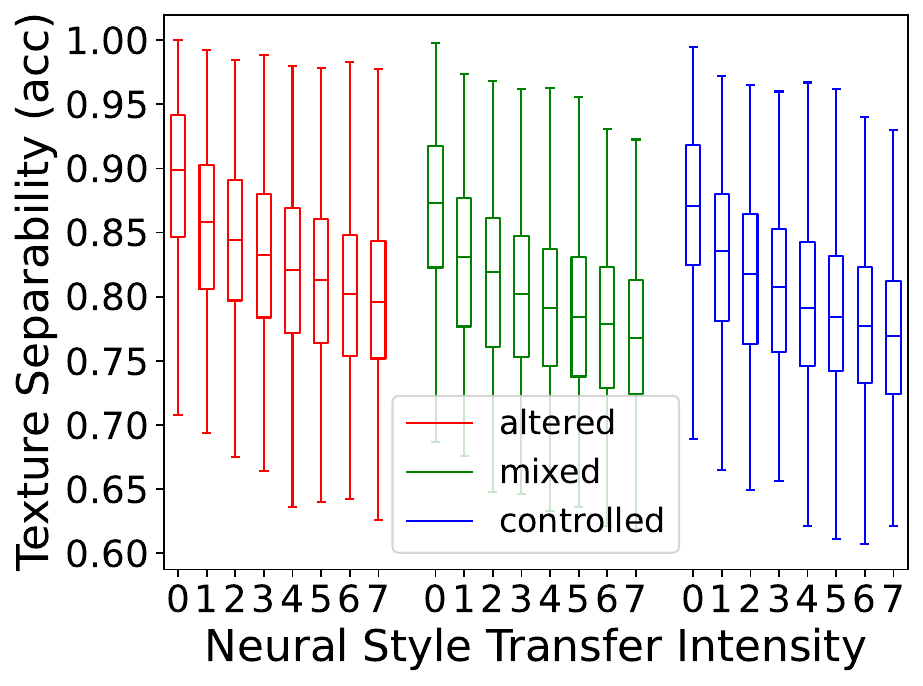}
     \includegraphics[width=.45\linewidth]{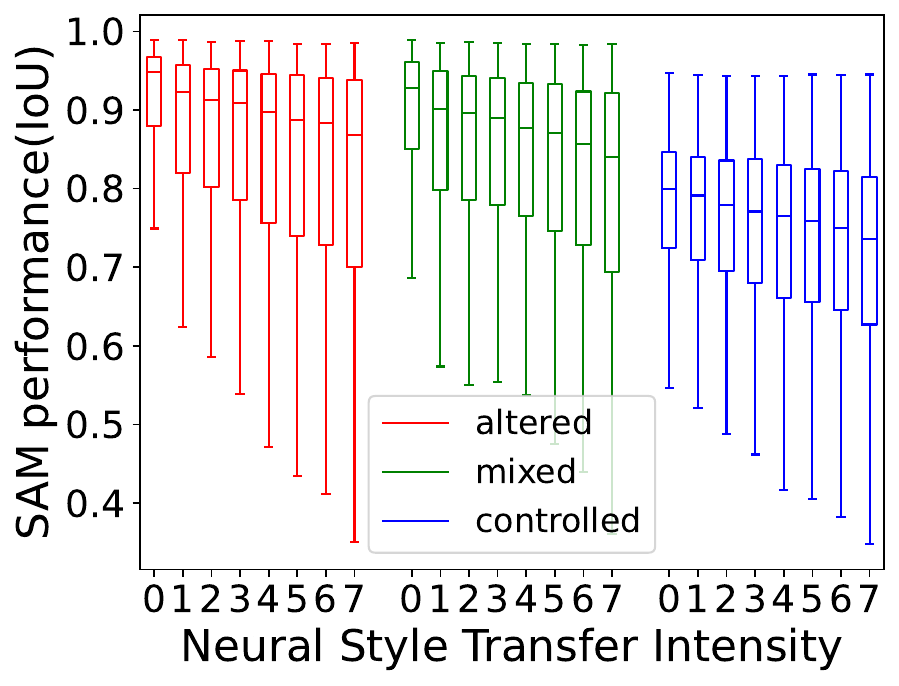}
     \caption{\textbf{Left:} Increasing NST intensity results in decreased textural separability. \textbf{Right:} SAM segmentation IoU vs. NST intensity, on the three composite image types.}
    \label{fig:NST_intensity_effect_sep}
\end{figure}

To systematically vary textural separability, we apply NST at eight different intensity levels, adjusting the balance between content preservation and style transfer (details in App. \ref{app:NSTmodels}). Fig. \ref{fig:NST_intensity} provides examples of this, and Fig. \ref{fig:NST_intensity_effect_sep} (left) confirms that increasing NST intensity steadily reduces textural separability as intended.

In Fig. \ref{fig:NST_intensity_effect_sep} right, we analyze how SAM’s performance varies with NST intensity across the three object transformation types (we found similar results for the other SFMs, shown in App. \ref{app:NSTperformance}). First, as textural separability decreases (higher NST intensity), SAM’s performance declines, confirming the impact of textural contrast on segmentation accuracy. Second, we observe clear performance gaps between transformation types: SAM performs best on altered objects, which undergo both shape and texture distortion, followed closely by mixed objects, and then performs worst on controlled objects. This suggests that additional textural contrast--introduced by shape, boundary, and texture transformations--enhances segmentation performance, further reinforcing the role of textural cues in SFM behavior.

\subsubsection{Experiments on Real Data}
\label{sec:sep_exp_real}

We will now evaluate datasets of real images: iShape, MOSE (Sec. \ref{sec:treelikeness_exp_real}), and Plittersdorf \cite{haucke2022socrates}---of which iShape and Plittersdorf were used in the original SAM paper \cite{sam}---which all possess objects with a wide range of textural separability from their surroundings. Plittersdorf consists of camera trap video frames of wild deer recorded in a wildlife park. These frames sometimes have low-contrast objects due to frequent low-light conditions, making it a useful dataset to analyze. Example images and objects from both datasets are shown in App. \ref{app:egplittersishape}.

\begin{figure*}
    \begin{minipage}[l]{0.35\textwidth}
    \centering
    \includegraphics[width=.49\linewidth]{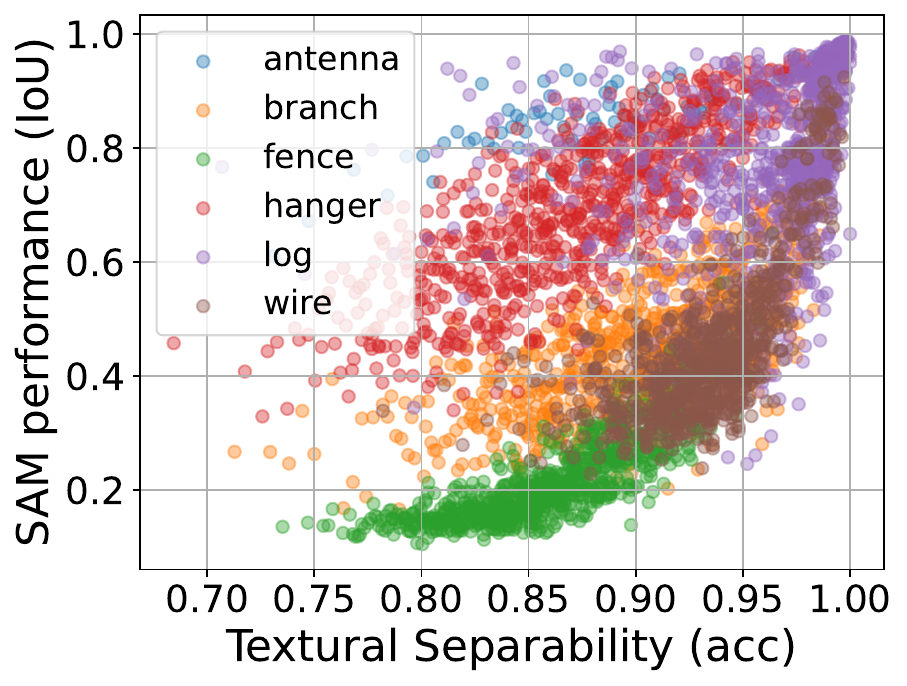}
    \includegraphics[width=.49\linewidth]{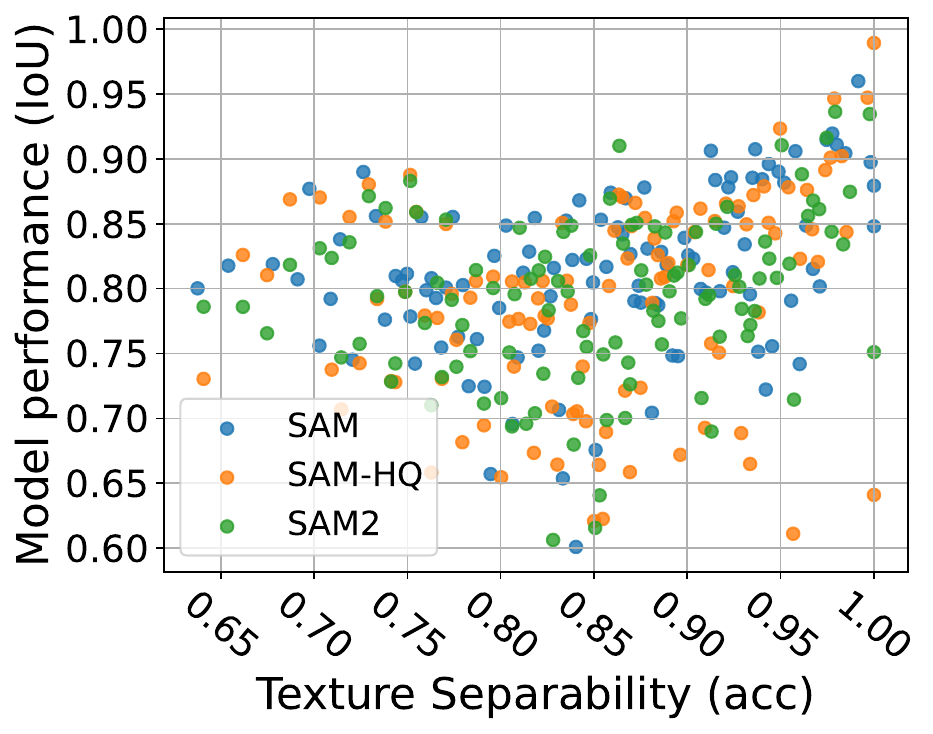}
    \end{minipage}
    \hfill
    \begin{minipage}[t]{0.65\textwidth}
    \setlength{\tabcolsep}{2pt}
    \centering
    % \small
    \fontsize{7pt}{7pt}\selectfont
    % \scriptsize
    \begin{tabular}{l|cc|cc|cc|cc|cc|cc||cc||cc}
    \multicolumn{1}{c}{} & \multicolumn{12}{c}{\textbf{iShape}} & \multicolumn{4}{c}{} \\
    \multicolumn{1}{c}{} & \multicolumn{2}{c|}{antenna} & \multicolumn{2}{c|}{branch} & \multicolumn{2}{c|}{fence} & \multicolumn{2}{c|}{hanger} & \multicolumn{2}{c|}{log} & \multicolumn{2}{c||}{wire} & \multicolumn{2}{c||}{\textbf{Plitters.}} & \multicolumn{2}{c}{\textbf{MOSE}} \\
    \toprule
    
    \textbf{Model} & $\tau$ & $\rho$ & $\tau$ & $\rho$ & $\tau$ & $\rho$ & $\tau$ & $\rho$ & $\tau$ & $\rho$ & $\tau$ & $\rho$ & $\tau$ & $\rho$ & $\tau$ & $\rho$ \\
    \midrule
    SAM-H & 0.44 & 0.59 & 0.50 & 0.68 & 0.65 & 0.85 & 0.63 & 0.83 & 0.33 & 0.48 & 0.49 & 0.67 & 0.26 & 0.38 & 0.65	&	0.83 \\
    SAM-B & 0.46 & 0.61 & 0.56 & 0.75 & 0.67 & 0.86 & 0.65 & 0.84 & 0.36 & 0.52 & 0.55 & 0.73 & 0.36 & 0.50 & 0.74	&	0.90 \\
    SAM 2-L & 0.46 & 0.64 & 0.56 & 0.76 & 0.64 & 0.83 & 0.65 & 0.84 & 0.25 & 0.36 & 0.54 & 0.71 & 0.20 & 0.29 & 0.58	&	0.76 \\
    SAM 2-B+ & 0.44 & 0.61 & 0.48 & 0.67 & 0.69 & 0.88 & 0.64 & 0.83 & 0.20 & 0.28 & 0.50 & 0.67 & 0.12 & 0.17 & 0.58	&	0.75 \\
    HQ-SAM-H & 0.39 & 0.54 & 0.50 & 0.68 & 0.69 & 0.88 & 0.62 & 0.80 & 0.20 & 0.28 & 0.47 & 0.63 & 0.20 & 0.28 & 0.64	&	0.82 \\
    HQ-SAM-B & 0.43 & 0.60 & 0.50 & 0.68 & 0.68 & 0.86 & 0.62 & 0.81 & 0.23 & 0.33 & 0.49 & 0.66 & 0.21 & 0.31 & 0.72	&	0.88 \\
    \midrule
    \textbf{Average} & \textbf{0.44} & \textbf{0.60} & \textbf{0.52} & \textbf{0.70} & \textbf{0.67} & \textbf{0.86} & \textbf{0.64} & \textbf{0.83} & \textbf{0.26} & \textbf{0.38} & \textbf{0.51} & \textbf{0.68} & \textbf{0.23} & \textbf{0.32} & \textbf{0.65} & \textbf{0.82} \\
    \bottomrule
\end{tabular}
    % \begin{tabular}{l|cc|cc|cc|cc|cc|cc||cc}
    % \multicolumn{1}{c}{} & \multicolumn{12}{c}{\textbf{iShape}} & \multicolumn{2}{c}{\textbf{}} \\
    % \multicolumn{1}{c}{} & \multicolumn{2}{c|}{antenna} & \multicolumn{2}{c|}{branch} & \multicolumn{2}{c|}{fence} & \multicolumn{2}{c|}{hanger} & \multicolumn{2}{c|}{log} & \multicolumn{2}{c||}{wire} & \multicolumn{2}{c}{\textbf{Plittersdorf}} \\
    % \toprule
    
    % \textbf{Model} & $\tau$ & $\rho$ & $\tau$ & $\rho$ & $\tau$ & $\rho$ & $\tau$ & $\rho$ & $\tau$ & $\rho$ & $\tau$ & $\rho$ & $\tau$ & $\rho$ \\
    % \midrule
    % SAM-H & 0.44 & 0.59 & 0.50 & 0.68 & 0.65 & 0.85 & 0.63 & 0.83 & 0.33 & 0.48 & 0.49 & 0.67 & 0.26 & 0.38 \\
    % SAM-B & 0.46 & 0.61 & 0.56 & 0.75 & 0.67 & 0.86 & 0.65 & 0.84 & 0.36 & 0.52 & 0.55 & 0.73 & 0.36 & 0.50 \\
    % SAM 2-L & 0.46 & 0.64 & 0.56 & 0.76 & 0.64 & 0.83 & 0.65 & 0.84 & 0.25 & 0.36 & 0.54 & 0.71 & 0.20 & 0.29  \\
    % SAM 2-B+ & 0.44 & 0.61 & 0.48 & 0.67 & 0.69 & 0.88 & 0.64 & 0.83 & 0.20 & 0.28 & 0.50 & 0.67 & 0.12 & 0.17  \\
    % HQ-SAM-H & 0.39 & 0.54 & 0.50 & 0.68 & 0.69 & 0.88 & 0.62 & 0.80 & 0.20 & 0.28 & 0.47 & 0.63 & 0.20 & 0.28  \\
    % HQ-SAM-B & 0.43 & 0.60 & 0.50 & 0.68 & 0.68 & 0.86 & 0.62 & 0.81 & 0.23 & 0.33 & 0.49 & 0.66 & 0.21 & 0.31  \\\midrule
    % \textbf{Average} & \textbf{0.44} & \textbf{0.60} & \textbf{0.52} & \textbf{0.70} & \textbf{0.67} & \textbf{0.86} & \textbf{0.64} & \textbf{0.83} & \textbf{0.26} & \textbf{0.38} & \textbf{0.51} & \textbf{0.68} & \textbf{0.23} & \textbf{0.32} \\
    % \bottomrule
    % \end{tabular}
    \end{minipage}
    \caption{\textbf{Left:} Segmentation IoU vs. object textural separability, for iShape (left) shown for SAM-H, and Plittersdorf (right) for all SFMs. \textbf{Right:} Rank correlations between SFM IoU and object textural separability, on iShape, Plittersdorf and MOSE for all SFMs.}
    \label{figtab:sep_real}
\end{figure*}

In Fig. \ref{figtab:sep_real} left, we show how the textural separability (Algorithm \ref{alg:textural_sep}) of these objects relates to SAM's performance on them, with correlation results shown for all three SFMs to the right of the plots (MOSE plots shown in App. \ref{app:MOSEallresults}).
Overall, across all datasets, we see a fairly strong correlation between textural separability and segmentation IoU (average corr. of $\tau = 0.49$ and $\rho =0.65$), especially considering the variety of objects and backgrounds, with objects with low separability typically being noticeably harder to segment. This relationship was especially strong for the MOSE dataset, with average $\tau=0.65$ and $\rho=0.82$. In these cases, the SFMs were confused by objects close to the object of interest which \textit{also} have similar textures.

\section{Robustness to Fine-tuning}
\label{sec:finetune}
Despite being designed for thin, branching structures, HQ-SAM exhibits performance drops on high tree-likeness objects, similar to SAM and SAM 2 (Fig. \ref{figtab:treelikeness_results_real}). This suggests that the difficulty SFMs face with such objects persists even after adaptation, a hypothesis we explore here.

To assess whether fine-tuning could improve SAM’s segmentation of high tree-likeness objects, we selected some $k$ examples from DIS-train using different ranking strategies: \textbf{Oracle} (direct IoU-based selection), \textbf{Rand} (random selection), \textbf{DoGD} / \textbf{CPR} (selecting examples based on high tree-likeness), and \textbf{Sum} (a combination of DoGD and CPR). We fine-tuned only SAM’s mask decoder on these, keeping the image encoder frozen to preserve its generalist capabilities (as in \cite{hqsam}); full details are in App. \ref{app:finetune}.

\begin{figure}[htbp!]
     \centering
     \includegraphics[width=.63\linewidth]{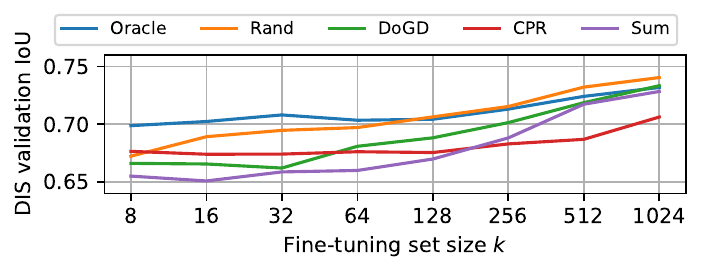}
     \includegraphics[width=.4\linewidth, trim={0 0 0 0cm}]{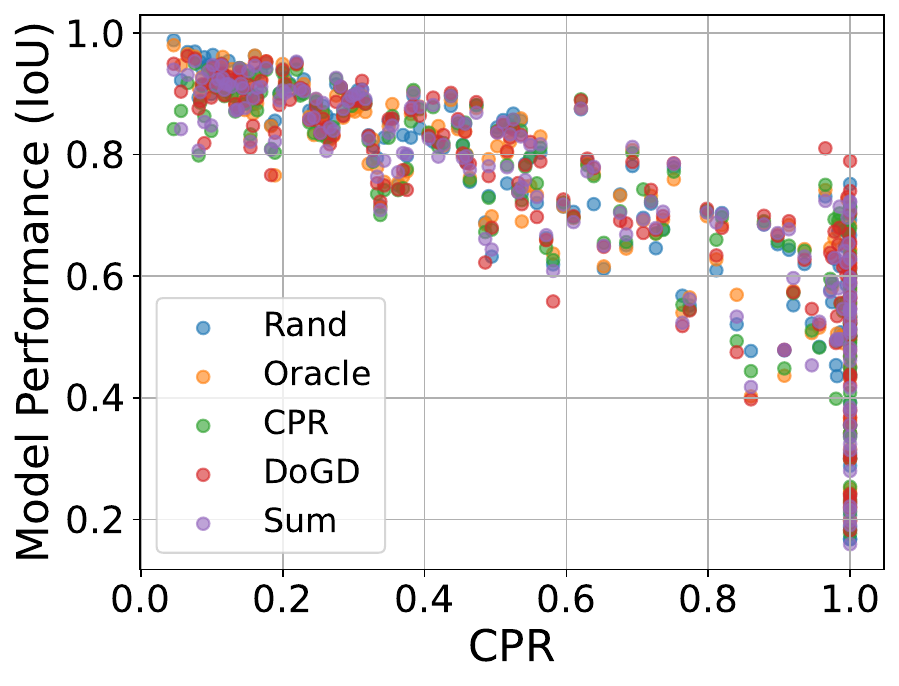}
     \includegraphics[width=.4\linewidth, trim={0 1.1cm 0 0cm}]{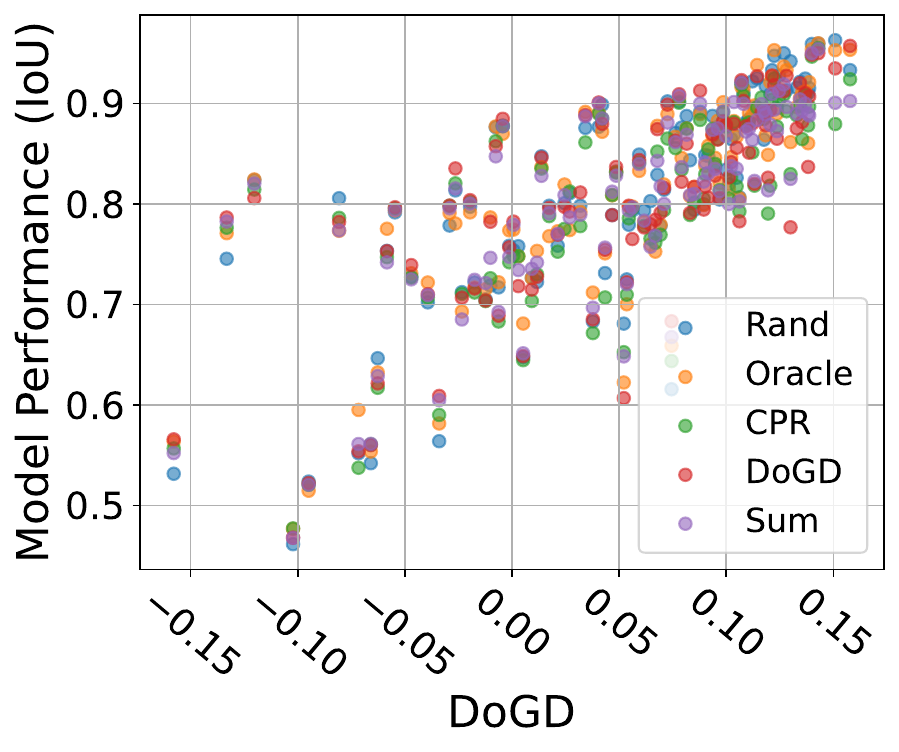}
     \includegraphics[width=.4\linewidth, trim={0 0cm 0 0}]{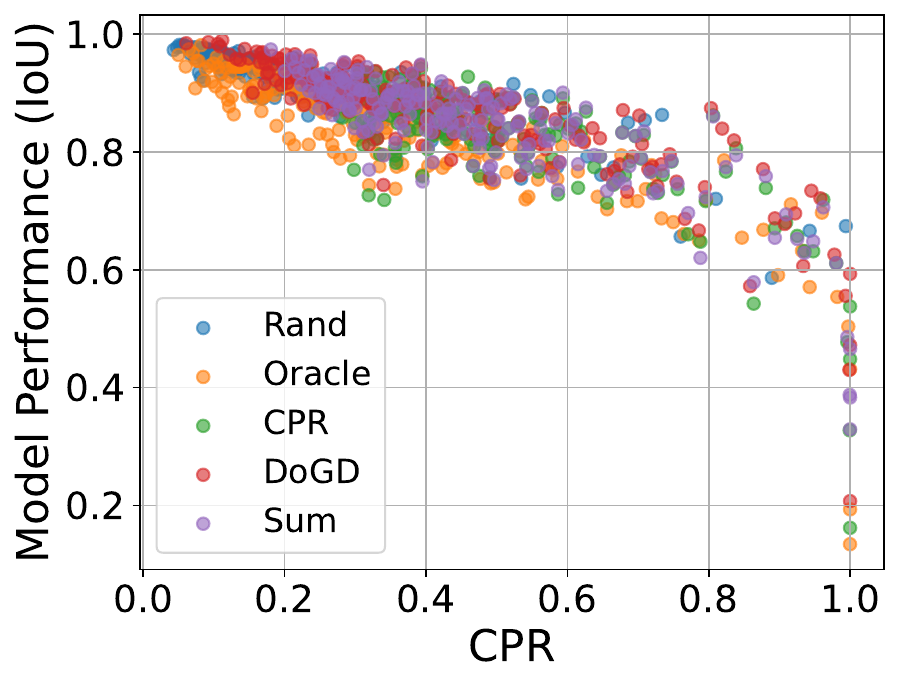}
     \includegraphics[width=.4\linewidth, trim={0 1cm 0 -1cm}]{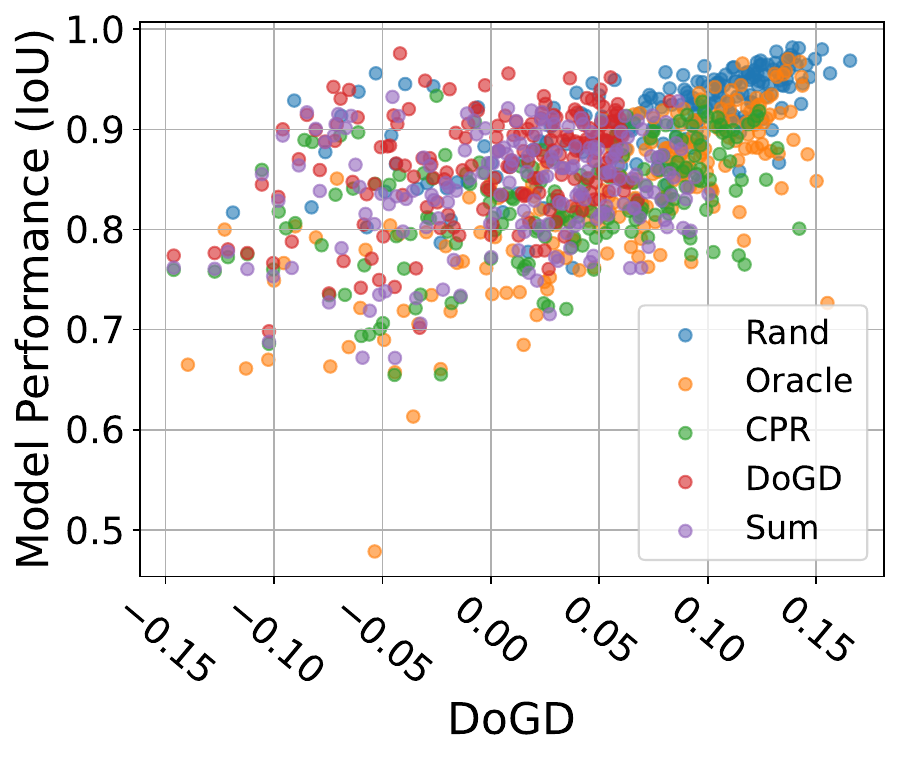}
    \caption{\textbf{Top:} Performance (IoU) of fine-tuned SAM on DIS val vs. number of fine-tuning examples $k$ sampled from DIS train, using different selection strategies. \textbf{Lower rows:} IoU vs. tree-likeness for each $k=1024$ fine-tuned model on DIS val (middle row) and on the fine-tuning set (bottom row).}
    \label{fig:finetune}
\end{figure}

Results (Fig. \ref{fig:finetune}) show that fine-tuning on high tree-likeness examples (DoGD, CPR, or Sum) led to slightly worse validation performance than random selection, likely due to overfitting on these atypical examples, harming generalization. Even though fine-tuning improved overall average IoU, the correlation between segmentation performance and tree-likeness remained---both on unseen validation data (middle plots, Fig. \ref{fig:finetune}) \textit{and} even on the fine-tuned examples themselves (lower plots). These findings indicate that fine-tuning alone is not a viable solution to overcoming SFMs’ performance drop on highly tree-like objects.

\section{Discussion}
\label{sec:discussion}
\textbf{Understanding SFM Failure Modes.}
We find that SFMs struggle primarily with over-segmentation rather than under-segmentation in high tree-likeness cases (Sec. \ref{sec:treelikeness_exp}), generating large false-positive regions within densely packed, thin structures (\eg, Figs. \ref{fig:eg_synthetic_treelike}, right; \ref{fig:teaser}, left). This cannot be explained by resolution limitations alone, as sparse thin structures exhibit reduced over-segmentation (Fig. \ref{fig:eg_synthetic_treelike}, center rightmost). Instead, we hypothesize that SFMs misinterpret dense, irregular patterns as texture, confusing these structures with the appearance of a more uniformly-shaped object.

For both tree-like and low-contrast objects, SFMs struggle to delineate object boundaries when faced with misleading textural cues, be it either a dense, irregular shape that appears to instead be a texture (the former case), or an object texture similar to its surroundings (the latter). This suggests that SAM’s training set (SA-1B) contained few such objects, as tree-like structures and low-contrast objects are uncommon in typical photographic datasets \cite{sam} (for the former, see \eg, the typically low object concavity of SA-1B in Fig. 6, right in \cite{sam}). Since HQ-SAM and SAM 2 were derived from SAM, their similar behavior is unsurprising.

\textbf{Comparison to Non-SFM Models.}
Additional experiments with models trained from scratch (IS-Net, iShapeInst, Mask R-CNN) and evaluated on the DIS, iShape and Plittersdorf datasets, respectively (App. \ref{app:trainedfromscratch}), revealed that their performance did not sharply degrade on objects with high tree-likeness or low textural separability. In contrast to SFMs, such objects were not consistent ``failure modes'' for these models, suggesting that these difficulties are specific to SFMs rather than being inherently hard cases.

\textbf{On HQ-SAM and Fine-tuning.}
Despite being designed and adapted for thin, intricate structures \cite{hqsam}, HQ-SAM still exhibits performance drops correlated with tree-likeness (Figs. \ref{figtab:treelikeness_synthetic}, \ref{figtab:treelikeness_results_real}), albeit with slight improvements over SAM and SAM 2 (Fig. \ref{figtab:treelikeness_results_real}, left). This may stem from HQ-SAM’s adapter-based approach, which leaves SAM’s image encoder and mask decoder frozen, preserving SAM’s limitations. Our fine-tuning experiments (Sec. \ref{sec:finetune}) further suggest that this issue is not easily mitigated, indicating that these structures pose a fundamental segmentation challenge.

\textbf{Limitations and Future Work.}
To ensure our findings were robust, we minimized confounding factors by first establishing trends in controlled synthetic experiments before validating them on real datasets spanning diverse object types. However, real-world segmentation is inherently complex, making it impossible to eliminate all confounding variables---which may explain why observed correlations were stronger in synthetic data than in real datasets.

A key future direction is to identify the specific model components responsible for these failures. SAM’s image encoder certainly plays a central role, as it serves as the primary feature extractor for the lightweight mask decoder, but a more precise mechanistic analysis could guide architectural improvements to better handle complex structures.

% \textbf{Conclusion.}
% In this paper, we quantitatively modeled how SFM segmentation performance relates to tree-likeness and textural separability. Our findings reveal a consistent correlation between these object characteristics and segmentation accuracy, highlighting the need for further research to better understand and mitigate these failure modes—especially since even targeted fine-tuning fails to fully resolve them.

\section*{Acknowledgements}
Research reported in this publication was supported by the National Heart, Lung, and Blood Institute of the National Institutes of Health under Award Number R44HL152825. The content is solely the responsibility of the authors and does not necessarily represent the official views of the National Institutes of Health.

{
    \small
    \bibliographystyle{ieeenat_fullname}
    \bibliography{main}

@inproceedings{sam,
  title={Segment anything},
  author={Kirillov, Alexander and Mintun, Eric and Ravi, Nikhila and Mao, Hanzi and Rolland, Chloe and Gustafson, Laura and Xiao, Tete and Whitehead, Spencer and Berg, Alexander C and Lo, Wan-Yen and others},
  booktitle={Proceedings of the IEEE/CVF International Conference on Computer Vision},
  pages={4015--4026},
  year={2023}
}

@article{shi2023generalist,
  title={Generalist vision foundation models for medical imaging: A case study of segment anything model on zero-shot medical segmentation},
  author={Shi, Peilun and Qiu, Jianing and Abaxi, Sai Mu Dalike and Wei, Hao and Lo, Frank P-W and Yuan, Wu},
  journal={Diagnostics},
  volume={13},
  number={11},
  pages={1947},
  year={2023},
  publisher={MDPI}
}

@article{dong_efficient_2024,
	title = {An efficient segment anything model for the segmentation of medical images},
	volume = {14},
	issn = {2045-2322},
	url = {https://doi.org/10.1038/s41598-024-70288-8},
	doi = {10.1038/s41598-024-70288-8},
	number = {1},
	journal = {Scientific Reports},
	author = {Dong, Guanliang and Wang, Zhangquan and Chen, Yourong and Sun, Yuliang and Song, Hongbo and Liu, Liyuan and Cui, Haidong},
	month = aug,
	year = {2024},
	pages = {19425},
}

@inproceedings{
loshchilov2018decoupled,
title={Decoupled Weight Decay Regularization},
author={Ilya Loshchilov and Frank Hutter},
booktitle={International Conference on Learning Representations},
year={2019},
url={https://openreview.net/forum?id=Bkg6RiCqY7},
}

@article{cardoso2022monai,
  title={Monai: An open-source framework for deep learning in healthcare},
  author={Cardoso, M Jorge and Li, Wenqi and Brown, Richard and Ma, Nic and Kerfoot, Eric and Wang, Yiheng and Murrey, Benjamin and Myronenko, Andriy and Zhao, Can and Yang, Dong and others},
  journal={arXiv preprint arXiv:2211.02701},
  year={2022}
}

@article{zhang2024convolutional,
  title={Convolutional neural networks rarely learn shape for semantic segmentation},
  author={Zhang, Yixin and Mazurowski, Maciej A},
  journal={Pattern Recognition},
  volume={146},
  pages={110018},
  year={2024},
  publisher={Elsevier}
}

@inproceedings{geirhosimagenet2019,
  title={ImageNet-trained CNNs are biased towards texture; increasing shape bias improves accuracy and robustness},
  author={Geirhos, Robert and Rubisch, Patricia and Michaelis, Claudio and Bethge, Matthias and Wichmann, Felix A and Brendel, Wieland},
  booktitle={International Conference on Learning Representations},
  year={2019}
}

@article{subramanian2024spatial,
  title={Spatial-frequency channels, shape bias, and adversarial robustness},
  author={Subramanian, Ajay and Sizikova, Elena and Majaj, Najib and Pelli, Denis},
  journal={Advances in Neural Information Processing Systems},
  volume={36},
  year={2024}
}

@article{hermann2020origins,
  title={The origins and prevalence of texture bias in convolutional neural networks},
  author={Hermann, Katherine and Chen, Ting and Kornblith, Simon},
  journal={Advances in Neural Information Processing Systems},
  volume={33},
  pages={19000--19015},
  year={2020}
}

@article{tuli2021convolutional,
  title={Are convolutional neural networks or transformers more like human vision?},
  author={Tuli, Shikhar and Dasgupta, Ishita and Grant, Erin and Griffiths, Thomas L},
  journal={arXiv preprint arXiv:2105.07197},
  year={2021}
}

@article{baker2018deep,
  title={Deep convolutional networks do not classify based on global object shape},
  author={Baker, Nicholas and Lu, Hongjing and Erlikhman, Gennady and Kellman, Philip J},
  journal={PLoS computational biology},
  volume={14},
  number={12},
  pages={e1006613},
  year={2018},
  publisher={Public Library of Science San Francisco, CA USA}
}

@inproceedings{li2024semanticsam,
  title={Segment and Recognize Anything at Any Granularity},
  author={Li, Feng and Zhang, Hao and Sun, Peize and Zou, Xueyan and Liu, Shilong and Li, Chunyuan and Yang, Jianwei and Zhang, Lei and Gao, Jianfeng},
  booktitle={European Conference on Computer Vision},
  pages={467--484},
  year={2024},
  organization={Springer}
}

@inproceedings{yuan2024open,
  title={Open-vocabulary SAM: Segment and recognize twenty-thousand classes interactively},
  author={Yuan, Haobo and Li, Xiangtai and Zhou, Chong and Li, Yining and Chen, Kai and Loy, Chen Change},
  booktitle={European Conference on Computer Vision},
  pages={419--437},
  year={2024},
  organization={Springer}
}

@article{ma2024segment,
  title={Segment anything in medical images},
  author={Ma, Jun and He, Yuting and Li, Feifei and Han, Lin and You, Chenyu and Wang, Bo},
  journal={Nature Communications},
  volume={15},
  number={1},
  pages={654},
  year={2024},
  publisher={Nature Publishing Group UK London}
}

@inproceedings{chen2023sam,
  title={Sam-adapter: Adapting segment anything in underperformed scenes},
  author={Chen, Tianrun and Zhu, Lanyun and Deng, Chaotao and Cao, Runlong and Wang, Yan and Zhang, Shangzhan and Li, Zejian and Sun, Lingyun and Zang, Ying and Mao, Papa},
  booktitle={Proceedings of the IEEE/CVF International Conference on Computer Vision Workshops},
  pages={3367--3375},
  year={2023}
}

@article{tang2023can,
  title={Can sam segment anything? when sam meets camouflaged object detection},
  author={Tang, Lv and Xiao, Haoke and Li, Bo},
  journal={arXiv preprint arXiv:2304.04709},
  year={2023}
}

@misc{wu2019detectron2,
  author =       {Yuxin Wu and Alexander Kirillov and Francisco Massa and
                  Wan-Yen Lo and Ross Girshick},
  title =        {Detectron2},
  howpublished = {\url{https://github.com/facebookresearch/detectron2}},
  year =         {2019}
}

@inproceedings{MOSE,
  title={{MOSE}: A New Dataset for Video Object Segmentation in Complex Scenes},
  author={Ding, Henghui and Liu, Chang and He, Shuting and Jiang, Xudong and Torr, Philip HS and Bai, Song},
  booktitle={ICCV},
  year={2023}
}

@article{moran1950notes,
  title={Notes on continuous stochastic phenomena},
  author={Moran, Patrick AP},
  journal={Biometrika},
  volume={37},
  number={1/2},
  pages={17--23},
  year={1950},
  publisher={JSTOR}
}

@article{chen2024rsprompter,
  title={RSPrompter: Learning to prompt for remote sensing instance segmentation based on visual foundation model},
  author={Chen, Keyan and Liu, Chenyang and Chen, Hao and Zhang, Haotian and Li, Wenyuan and Zou, Zhengxia and Shi, Zhenwei},
  journal={IEEE Transactions on Geoscience and Remote Sensing},
  year={2024},
  publisher={IEEE}
}

@inproceedings{ren2024segment,
  title={Segment anything, from space?},
  author={Ren, Simiao and Luzi, Francesco and Lahrichi, Saad and Kassaw, Kaleb and Collins, Leslie M and Bradbury, Kyle and Malof, Jordan M},
  booktitle={Proceedings of the IEEE/CVF Winter Conference on Applications of Computer Vision},
  pages={8355--8365},
  year={2024}
}

@article{ji2024segment,
	title = {Segment {Anything} {Is} {Not} {Always} {Perfect}: {An} {Investigation} of {SAM} on {Different} {Real}-world {Applications}},
	volume = {21},
	issn = {2731-5398},
	url = {https://doi.org/10.1007/s11633-023-1385-0},
	doi = {10.1007/s11633-023-1385-0},
	number = {4},
	journal = {Machine Intelligence Research},
	author = {Ji, Wei and Li, Jingjing and Bi, Qi and Liu, Tingwei and Li, Wenbo and Cheng, Li},
	month = aug,
	year = {2024},
	pages = {617--630},
}

@inproceedings{osti_10447851,
  title={Segment Anything Model (SAM) for Digital Pathology: Assess Zero-shot Segmentation on Whole Slide Imaging},
  author={Deng, Ruining and Cui, Can and Liu, Quan and Yao, Tianyuan and Remedios, Lucas Walker and Bao, Shunxing and Landman, Bennett A and Tang, Yucheng and Wheless, Lee E and Coburn, Lori A and others},
  booktitle={Medical Imaging with Deep Learning, short paper track},
year={2023}
}

@article{MAZUROWSKI2023102918,
title = {Segment anything model for medical image analysis: An experimental study},
journal = {Medical Image Analysis},
volume = {89},
pages = {102918},
year = {2023},
issn = {1361-8415},
doi = {https://doi.org/10.1016/j.media.2023.102918},
url = {https://www.sciencedirect.com/science/article/pii/S1361841523001780},
author = {Maciej A. Mazurowski and Haoyu Dong and Hanxue Gu and Jichen Yang and Nicholas Konz and Yixin Zhang}
}

@article{huang2024segment,
  title={Segment anything model for medical images?},
  author={Huang, Yuhao and Yang, Xin and Liu, Lian and Zhou, Han and Chang, Ao and Zhou, Xinrui and Chen, Rusi and Yu, Junxuan and Chen, Jiongquan and Chen, Chaoyu and others},
  journal={Medical Image Analysis},
  volume={92},
  pages={103061},
  year={2024},
  publisher={Elsevier}
}

@inproceedings{zeiler2014visualizing,
  title={Visualizing and understanding convolutional networks},
  author={Zeiler, Matthew D and Fergus, Rob},
  booktitle={Computer Vision--ECCV 2014: 13th European Conference, Zurich, Switzerland, September 6-12, 2014, Proceedings, Part I 13},
  pages={818--833},
  year={2014},
  organization={Springer}
}

@article{haucke2022socrates,
author = {Haucke, Timm and Kühl, Hjalmar S. and Steinhage, Volker},
title = {SOCRATES: Introducing Depth in Visual Wildlife Monitoring Using Stereo Vision},
journal = {Sensors},
volume = {22},
year = {2022},
number = {23},
article-number = {9082},
url = {https://www.mdpi.com/1424-8220/22/23/9082},
issn = {1424-8220},
doi = {10.3390/s22239082}
}

@article{yang2021ishape,
  title={iShape: A first step towards irregular shape instance segmentation},
  author={Yang, Lei and Wei, Yan Zi and He, Yisheng and Sun, Wei and Huang, Zhenhang and Huang, Haibin and Fan, Haoqiang},
  journal={arXiv preprint arXiv:2109.15068},
  year={2021}
}

@InProceedings{DeepGlobe18,
 author = {Demir, Ilke and Koperski, Krzysztof and Lindenbaum, David and Pang, Guan and Huang, Jing and Basu, Saikat and Hughes, Forest and Tuia, Devis and Raskar, Ramesh},
 title = {DeepGlobe 2018: A Challenge to Parse the Earth Through Satellite Images},
 booktitle = {The IEEE Conference on Computer Vision and Pattern Recognition (CVPR) Workshops},
 month = {June},
 year = {2018}
}

@misc{noauthor_retina_nodate,
	title = {Retina {Blood} {Vessel}},
	url = {https://www.kaggle.com/datasets/abdallahwagih/retina-blood-vessel},
	language = {en},
	urldate = {2024-10-24},
	author = {Ibrahim, Abdallah Wagih},
    year={2023}
}

@article{qiu2023learnable,
  title={Learnable ophthalmology sam},
  author={Qiu, Zhongxi and Hu, Yan and Li, Heng and Liu, Jiang},
  journal={arXiv preprint arXiv:2304.13425},
  year={2023}
}

@inproceedings{dosovitskiy2020image,
  title={An Image is Worth 16x16 Words: Transformers for Image Recognition at Scale},
  author={Dosovitskiy, Alexey and Beyer, Lucas and Kolesnikov, Alexander and Weissenborn, Dirk and Zhai, Xiaohua and Unterthiner, Thomas and Dehghani, Mostafa and Minderer, Matthias and Heigold, Georg and Gelly, Sylvain and others},
  booktitle={International Conference on Learning Representations},
  year={2020}
}

@article{feng2024road,
  title={Road-SAM: Adapting the Segment Anything Model to Road Extraction From Large Very-High-Resolution Optical Remote Sensing Images},
  author={Feng, Wenqing and Guan, Fangli and Sun, Chenhao and Xu, Wei},
  journal={IEEE Geoscience and Remote Sensing Letters},
  volume={21},
  pages={1--5},
  year={2024},
  publisher={IEEE}
}

@inproceedings{xu2023leveraging,
  title={Leveraging Segment-Anything model for automated zero-shot road width extraction from aerial imagery},
  author={Xu, Nan and Nice, Kerry and Seneviratne, Sachith and Stevenson, Mark},
  booktitle={2023 International Conference on Digital Image Computing: Techniques and Applications (DICTA)},
  pages={176--183},
  year={2023},
  organization={IEEE}
}

@article{kubilius2016deep,
  title={Deep neural networks as a computational model for human shape sensitivity},
  author={Kubilius, Jonas and Bracci, Stefania and Op de Beeck, Hans P},
  journal={PLoS computational biology},
  volume={12},
  number={4},
  pages={e1004896},
  year={2016},
  publisher={Public Library of Science San Francisco, CA USA}
}

@article{hermann2020shapes,
  title={What shapes feature representations? exploring datasets, architectures, and training},
  author={Hermann, Katherine and Lampinen, Andrew},
  journal={Advances in Neural Information Processing Systems},
  volume={33},
  pages={9995--10006},
  year={2020}
}

@inproceedings{he2016resnet,
  title={Deep residual learning for image recognition},
  author={He, Kaiming and Zhang, Xiangyu and Ren, Shaoqing and Sun, Jian},
  booktitle={Proceedings of the IEEE conference on computer vision and pattern recognition},
  pages={770--778},
  year={2016}
}

@inproceedings{deng2009imagenet,
  title={Imagenet: A large-scale hierarchical image database},
  author={Deng, Jia and Dong, Wei and Socher, Richard and Li, Li-Jia and Li, Kai and Fei-Fei, Li},
  booktitle={2009 IEEE conference on computer vision and pattern recognition},
  pages={248--255},
  year={2009},
  organization={Ieee}
}

@inproceedings{gatys2016image,
  title={Image style transfer using convolutional neural networks},
  author={Gatys, Leon A and Ecker, Alexander S and Bethge, Matthias},
  booktitle={Proceedings of the IEEE conference on computer vision and pattern recognition},
  pages={2414--2423},
  year={2016}
}

@article{scikit-image,
 title = {scikit-image: image processing in {P}ython},
 author = {van der Walt, {S}t\'efan and {S}ch\"onberger, {J}ohannes {L}. and
           {Nunez-Iglesias}, {J}uan and {B}oulogne, {F}ran\c{c}ois and {W}arner,
           {J}oshua {D}. and {Y}ager, {N}eil and {G}ouillart, {E}mmanuelle and
           {Y}u, {T}ony and the scikit-image contributors},
 year = {2014},
 month = {6},
 volume = {2},
 pages = {e453},
 journal = {PeerJ},
 issn = {2167-8359},
 url = {https://doi.org/10.7717/peerj.453},
 doi = {10.7717/peerj.453}
}

@inproceedings{abnar-zuidema-2020-quantifying,
    title = "Quantifying Attention Flow in Transformers",
    author = "Abnar, Samira  and
      Zuidema, Willem",
    editor = "Jurafsky, Dan  and
      Chai, Joyce  and
      Schluter, Natalie  and
      Tetreault, Joel",
    booktitle = "Proceedings of the 58th Annual Meeting of the Association for Computational Linguistics",
    month = jul,
    year = "2020",
    address = "Online",
    publisher = "Association for Computational Linguistics",
    url = "https://aclanthology.org/2020.acl-main.385/",
    doi = "10.18653/v1/2020.acl-main.385",
    pages = "4190--4197"
}

@article{scikit-learn,
  title={Scikit-learn: Machine Learning in {P}ython},
  author={Pedregosa, F. and Varoquaux, G. and Gramfort, A. and Michel, V.
          and Thirion, B. and Grisel, O. and Blondel, M. and Prettenhofer, P.
          and Weiss, R. and Dubourg, V. and Vanderplas, J. and Passos, A. and
          Cournapeau, D. and Brucher, M. and Perrot, M. and Duchesnay, E.},
  journal={Journal of Machine Learning Research},
  volume={12},
  pages={2825--2830},
  year={2011}
}

@article{abdelmounaime2013newMBT,
  title={New Brodatz-Based Image Databases for Grayscale Color and Multiband Texture Analysis},
  author={Abdelmounaime, Safia and Dong-Chen, He},
  journal={International Scholarly Research Notices},
  volume={2013},
  number={1},
  pages={876386},
  year={2013},
  publisher={Wiley Online Library}
}

@article{kendall1938new,
  title={A new measure of rank correlation},
  author={Kendall, Maurice G},
  journal={Biometrika},
  volume={30},
  number={1-2},
  pages={81--93},
  year={1938},
  publisher={Oxford University Press}
}

@article{spearman_proof_1904,
	title = {The proof and measurement of association between two things.},
	volume = {15},
	issn = {1939-8298(Electronic),0002-9556(Print)},
	doi = {10.2307/1412159},
	number = {1},
	journal = {The American Journal of Psychology},
	author = {Spearman, C.},
	year = {1904},
	note = {Place: US
Publisher: Univ of Illinois Press},
	pages = {72--101},
}

@book{breiman1984classification,
  title={Classification and regression trees},
  author={Breiman, Leo},
  year={1984},
  publisher={Routledge}
}

@misc{pascal-voc-2012,
	author = "Everingham, M. and Van~Gool, L. and Williams, C. K. I. and Winn, J. and Zisserman, A.",
	title = "The {PASCAL} {V}isual {O}bject {C}lasses {C}hallenge 2012 {(VOC2012)} {R}esults",
	howpublished = "http://host.robots.ox.ac.uk/pascal/VOC/voc2012/",
    year={2012}
}

@Article{info11020125,
    AUTHOR = {Buslaev, Alexander and Iglovikov, Vladimir I. and Khvedchenya, Eugene and Parinov, Alex and Druzhinin, Mikhail and Kalinin, Alexandr A.},
    TITLE = {Albumentations: Fast and Flexible Image Augmentations},
    JOURNAL = {Information},
    VOLUME = {11},
    YEAR = {2020},
    NUMBER = {2},
    ARTICLE-NUMBER = {125},
    URL = {https://www.mdpi.com/2078-2489/11/2/125},
    ISSN = {2078-2489},
    DOI = {10.3390/info11020125}
}

@inproceedings{peng2024sam,
  title={Sam-parser: Fine-tuning sam efficiently by parameter space reconstruction},
  author={Peng, Zelin and Xu, Zhengqin and Zeng, Zhilin and Yang, Xiaokang and Shen, Wei},
  booktitle={Proceedings of the AAAI Conference on Artificial Intelligence},
  volume={38},
  number={5},
  pages={4515--4523},
  year={2024}
}

@article{gu2024build,
  title={How to build the best medical image segmentation algorithm using foundation models: a comprehensive empirical study with Segment Anything Model},
  author={Gu, Hanxue and Dong, Haoyu and Yang, Jichen and Mazurowski, Maciej A},
  journal={arXiv preprint arXiv:2404.09957},
  year={2024}
}

@article{zhang2023understanding,
  title={Understanding segment anything model: SAM is biased towards texture rather than shape},
  author={Zhang, Chaoning and Qiao, Yu and Tariq, Shehbaz and Zheng, Sheng and Zhang, Chenshuang and Li, Chenghao and Shin, Hyundong and Hong, Choong Seon},
  journal={arXiv preprint arXiv:2311.11465},
  year={2023}
}

@inproceedings{
wang2025orderaware,
title={Order-aware Interactive Segmentation},
author={Bin Wang and Anwesa Choudhuri and Meng Zheng and Zhongpai Gao and Benjamin Planche and Andong Deng and Qin Liu and Terrence Chen and Ulas Bagci and Ziyan Wu},
booktitle={The Thirteenth International Conference on Learning Representations},
year={2025},
url={https://openreview.net/forum?id=8ZLzw5pIrc}
}

@inproceedings{lin2021real,
  title={Real-time high-resolution background matting},
  author={Lin, Shanchuan and Ryabtsev, Andrey and Sengupta, Soumyadip and Curless, Brian L and Seitz, Steven M and Kemelmacher-Shlizerman, Ira},
  booktitle={Proceedings of the IEEE/CVF Conference on Computer Vision and Pattern Recognition},
  pages={8762--8771},
  year={2021}
}

@inproceedings{DIS,
  title={Highly accurate dichotomous image segmentation},
  author={Qin, Xuebin and Dai, Hang and Hu, Xiaobin and Fan, Deng-Ping and Shao, Ling and Van Gool, Luc},
  booktitle={European Conference on Computer Vision},
  pages={38--56},
  year={2022},
  organization={Springer}
}

@article{hqsam,
  title={Segment anything in high quality},
  author={Ke, Lei and Ye, Mingqiao and Danelljan, Martin and Tai, Yu-Wing and Tang, Chi-Keung and Yu, Fisher and others},
  journal={Advances in Neural Information Processing Systems},
  volume={36},
  pages={29914--29934},
  year={2023}
}

@article{sam2,
  title={Sam 2: Segment anything in images and videos},
  author={Ravi, Nikhila and Gabeur, Valentin and Hu, Yuan-Ting and Hu, Ronghang and Ryali, Chaitanya and Ma, Tengyu and Khedr, Haitham and R{\"a}dle, Roman and Rolland, Chloe and Gustafson, Laura and others},
  journal={arXiv preprint arXiv:2408.00714},
  year={2024}
}

@article{zou2023seem,
  title={Segment everything everywhere all at once},
  author={Zou, Xueyan and Yang, Jianwei and Zhang, Hao and Li, Feng and Li, Linjie and Wang, Jianfeng and Wang, Lijuan and Gao, Jianfeng and Lee, Yong Jae},
  journal={Advances in neural information processing systems},
  volume={36},
  pages={19769--19782},
  year={2023}
}

@inproceedings{jain2023oneformer,
  title={Oneformer: One transformer to rule universal image segmentation},
  author={Jain, Jitesh and Li, Jiachen and Chiu, Mang Tik and Hassani, Ali and Orlov, Nikita and Shi, Humphrey},
  booktitle={Proceedings of the IEEE/CVF conference on computer vision and pattern recognition},
  pages={2989--2998},
  year={2023}
}

@inproceedings{huang2024sca,
  title={Segment and caption anything},
  author={Huang, Xiaoke and Wang, Jianfeng and Tang, Yansong and Zhang, Zheng and Hu, Han and Lu, Jiwen and Wang, Lijuan and Liu, Zicheng},
  booktitle={Proceedings of the IEEE/CVF conference on computer vision and pattern recognition},
  pages={13405--13417},
  year={2024}
}

@inproceedings{he2017mask,
  title={Mask r-cnn},
  author={He, Kaiming and Gkioxari, Georgia and Doll{\'a}r, Piotr and Girshick, Ross},
  booktitle={Proceedings of the IEEE international conference on computer vision},
  pages={2961--2969},
  year={2017}
}

@inproceedings{clip,
  title={Learning transferable visual models from natural language supervision},
  author={Radford, Alec and Kim, Jong Wook and Hallacy, Chris and Ramesh, Aditya and Goh, Gabriel and Agarwal, Sandhini and Sastry, Girish and Askell, Amanda and Mishkin, Pamela and Clark, Jack and others},
  booktitle={International conference on machine learning},
  pages={8748--8763},
  year={2021},
  organization={PmLR}
}

@article{wang2025ishapeinst,
  title={iShapeInst: A novel and effective instance segmentation for irregularly shaped objects},
  author={Wang, Yang and Zhou, Wanlin and Zhou, Jiakai},
  journal={Neurocomputing},
  volume={638},
  pages={130183},
  year={2025},
  publisher={Elsevier}
}
}

% appendix
\newpage
\appendix
\onecolumn

\clearpage
\setcounter{page}{1}
\appendix

\section{Additional Dataset Details}
\label{app:datasets}

\subsection{Retinal Blood Vessel and Road Satellite Images}
\label{app:datasets:synthetictree}

The retinal blood vessel and road satellite image object masks which we use to generate the synthetic images for the experiments in Sec. \ref{sec:treelikeness_exp_synth}, are from the Retina Blood Vessel \cite{noauthor_retina_nodate} and the Road Extraction Challenge data of DeepGlobe18 \cite{DeepGlobe18}. For our retinal vessel mask set, we use the vessel masks from the training set of 80 image/mask pairs, and for our road mask set, we randomly sample the same number of masks from the split-merged full dataset, in order to ensure that these two object types appear equally in the synthetic dataset of Sec. \ref{sec:treelikeness_exp_synth}.

\subsection{Synthetic Tree-like Object Images}
\label{app:treelikesynth_creation}

Our algorithm for creating synthetic images of tree-like objects (used in Sec. \ref{sec:treelikeness_exp_synth}) is shown as follows.
\begin{enumerate}
    \item Sample object mask $m$ from either retinal blood vessel or satellite road image datasets (details in App. \ref{app:datasets:synthetictree}).
    \item Randomly select one of the contiguous components of $m$, denoted $m_c$, and enclose it with a tight bounding box.
    \item Resize $m_c$ such that its bounding box is $512\times 512$, and randomly place it in a blank $1024\times 1024$ image. %(this step ensures that the object takes up a reasonable amount of space in the image). 
    \item Apply two different textures randomly sampled from Colored Brodatz \cite{abdelmounaime2013newMBT} to the foreground ($m_c=1$) and background ($m_c=0$), resulting in the final image.
\end{enumerate}
In Fig. \ref{fig:synth_treelike_combinations} we show example images generated with different foreground and background texture combinations for various objects.

\subsection{Neural Style Transfer (NST) Generated Images}
\label{app:nst_additionalimages}

\begin{figure*}[htbp]
     \centering
     \includegraphics[width=.45\linewidth]{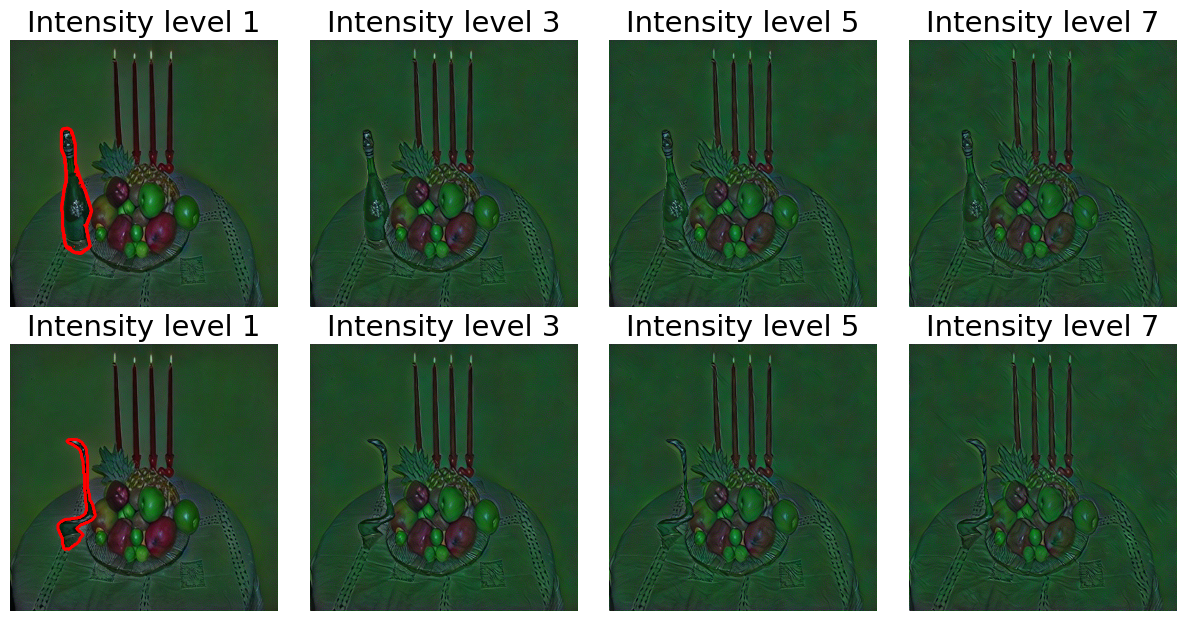}
     \includegraphics[width=.45\linewidth]{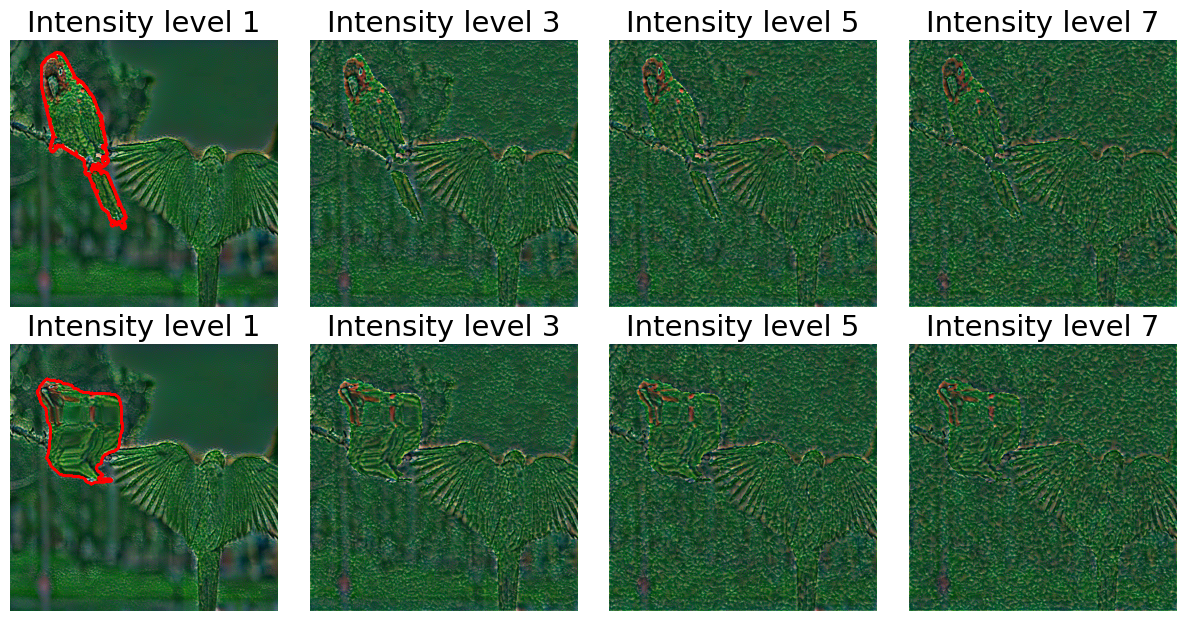}
     \includegraphics[width=.45\linewidth]{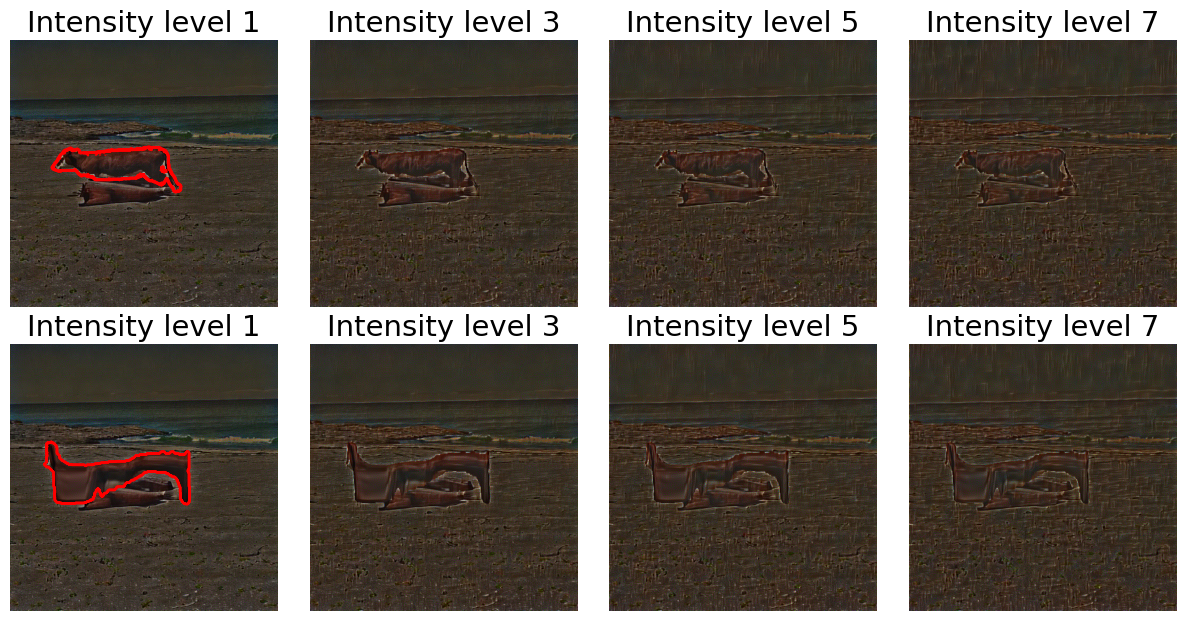}
     \includegraphics[width=.45\linewidth]{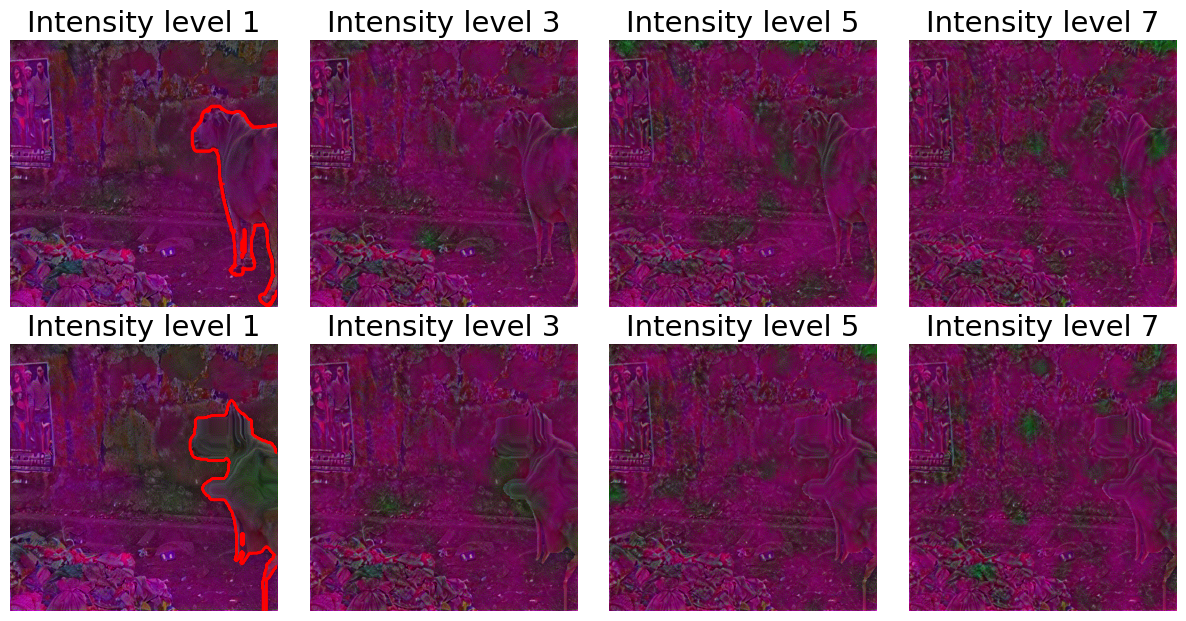}
     \includegraphics[width=.45\linewidth]{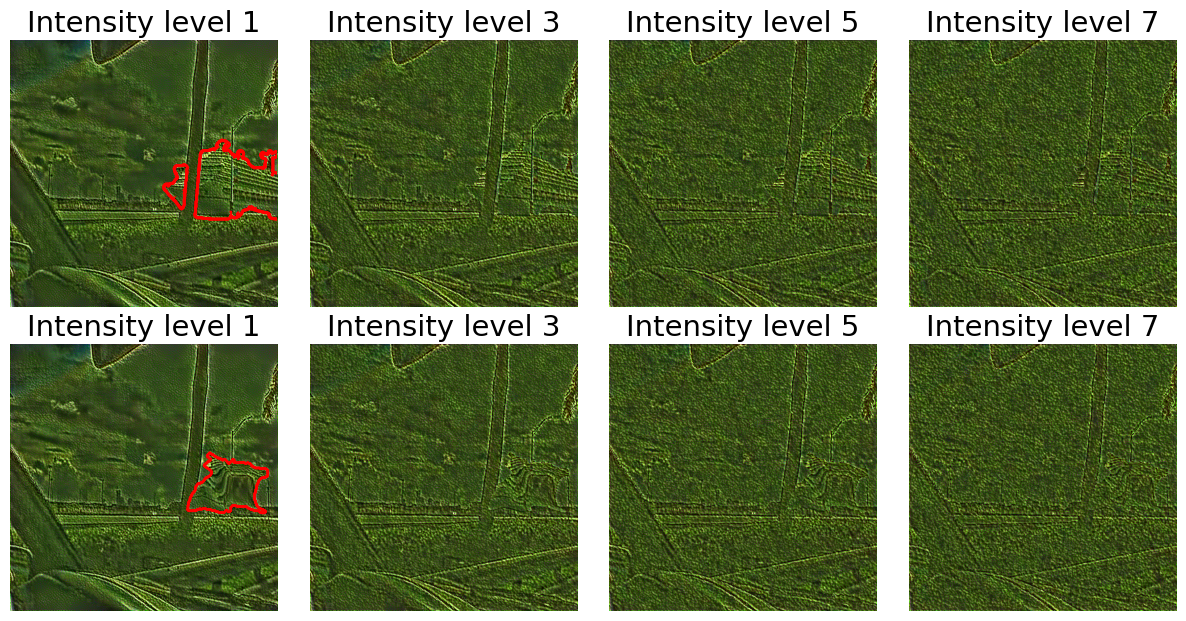}
     \includegraphics[width=.45\linewidth]{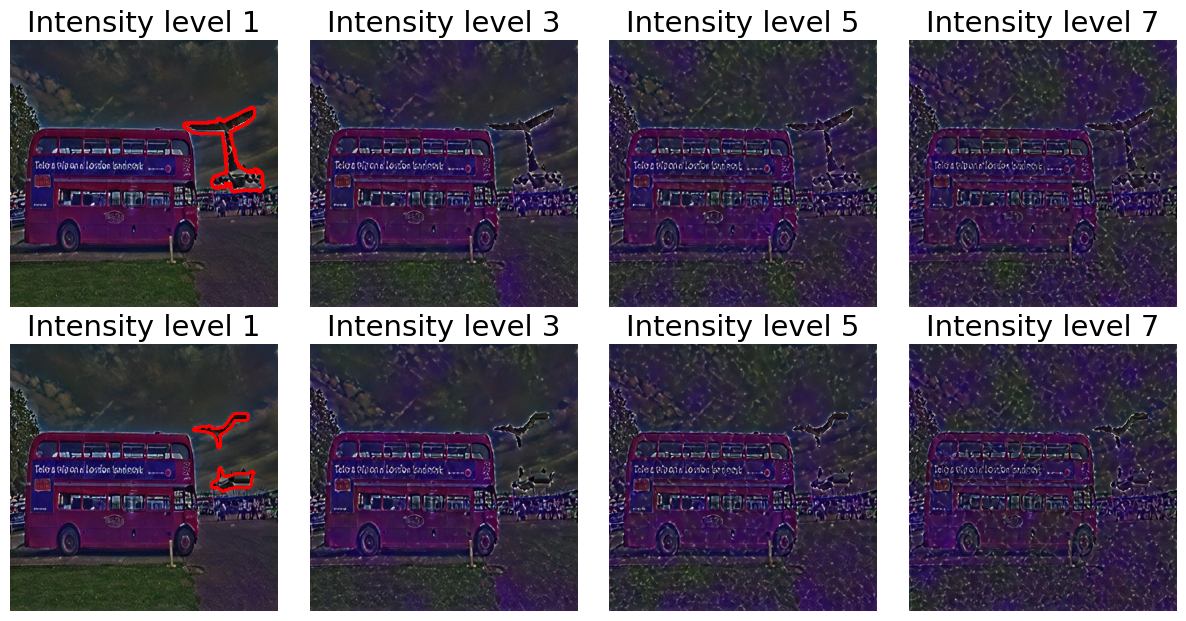}
     \includegraphics[width=.45\linewidth]{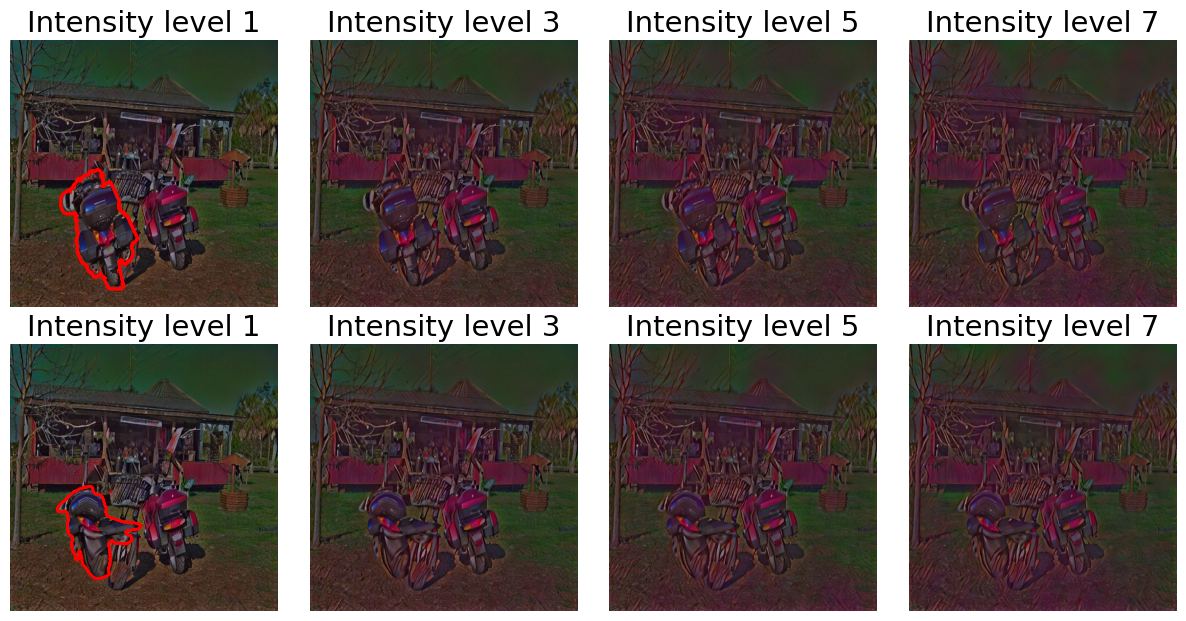}
     \includegraphics[width=.45\linewidth]{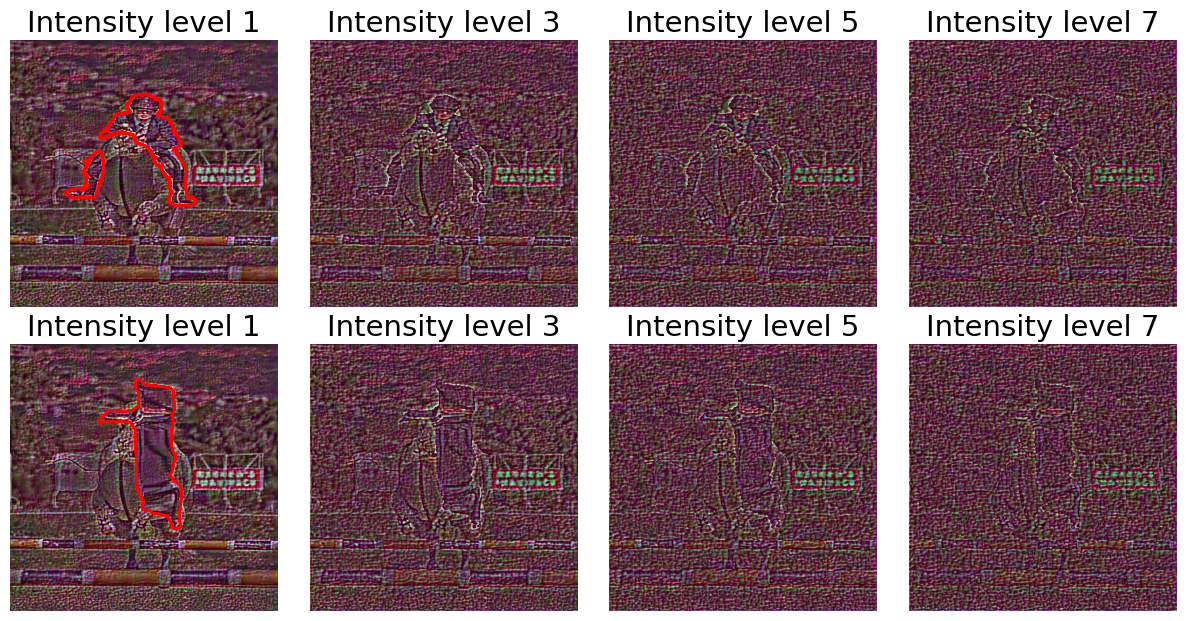}
     \caption{Additional examples of inpainting+neural style transfer-generated images to supplement Fig. \ref{fig:NST_example}.}
    \label{fig:NST_intensity_suppl}
\end{figure*}

\begin{figure}[htbp]
     \centering
     \includegraphics[width=.99\linewidth]{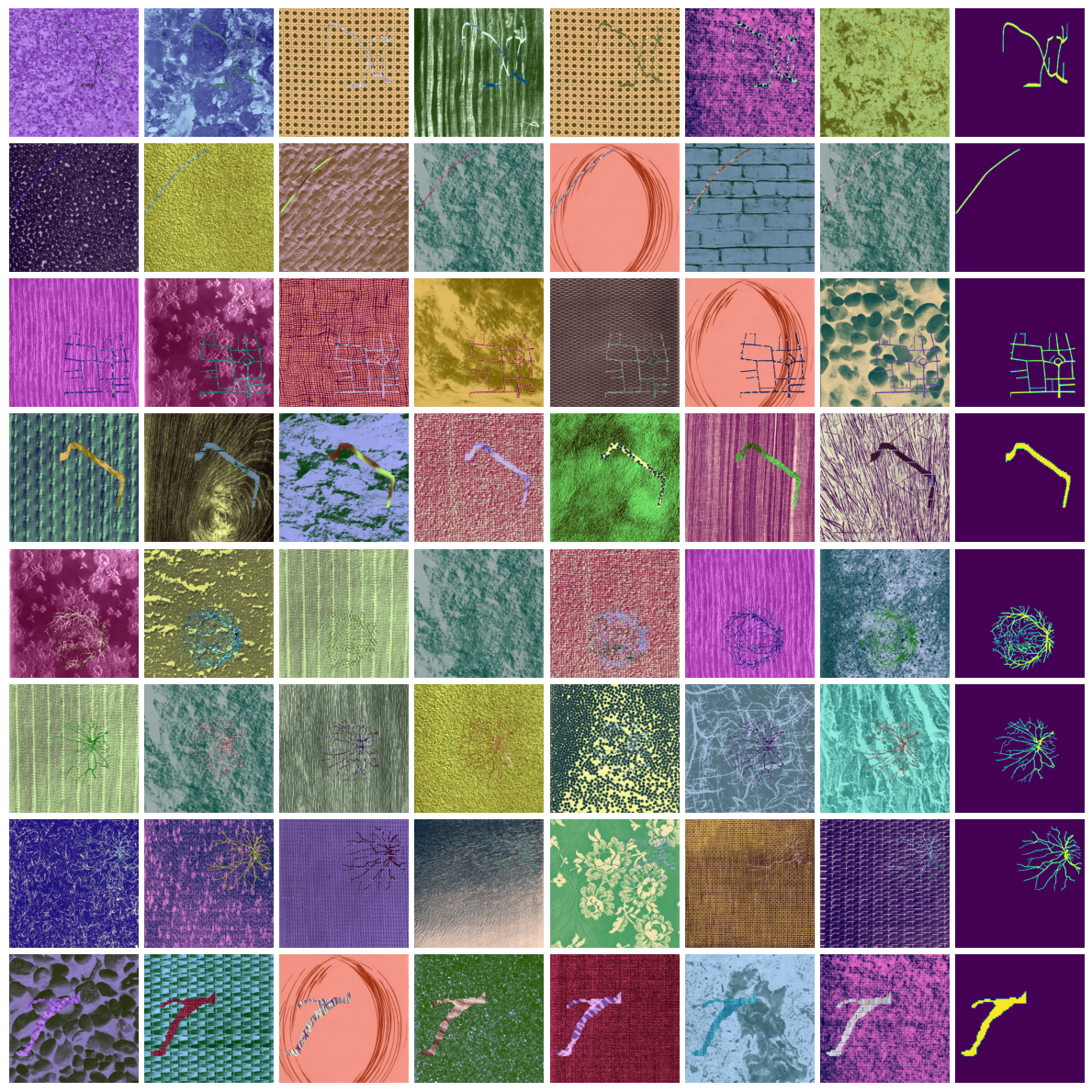}
     \caption{Examples of synthetic tree-like object images. Object masks shown in the right-most column, with images created using randomly chosen different foreground and background texture combinations in the left columns (before the majority-voting procedure described in Sec. \ref{sec:treelikeness_exp_synth}).}
    \label{fig:synth_treelike_combinations}
\end{figure}

In Fig. \ref{fig:NST_intensity_suppl} we provide additional example images generated via inpainting and neural style transfer for the textural separability experiments of Sec. \ref{sec:exp_nst}.

\subsection{Example Images from Plittersdorf and iShape}
\label{app:egplittersishape}

\begin{figure}[htbp!]
    \centering
    \includegraphics[width=0.99\linewidth]{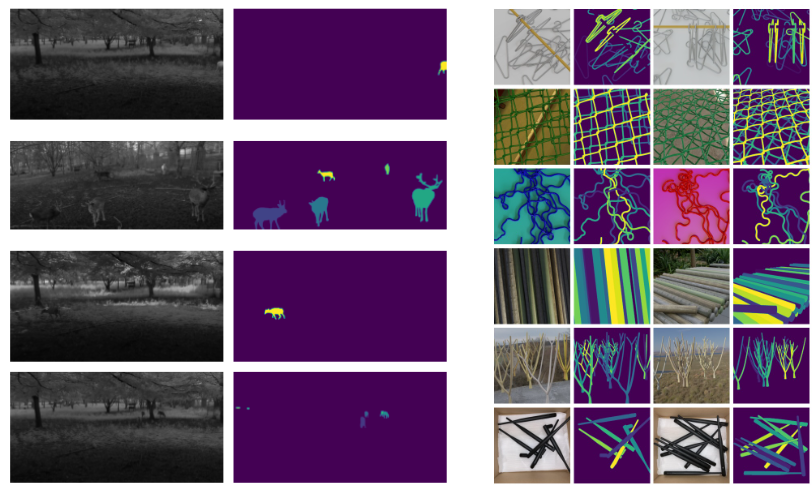}
    \caption{Example images and object masks from Plittersdorf (left group) and iShape (right group).}
    \label{fig:eg_plitters}
\end{figure}

We show example images from Plittersdorf and iShape in Fig. \ref{fig:eg_plitters}.

\section{Hyperparameter/Ablation Studies}

\subsection{Object Tree-likeness Experiments}
\label{app:treelikeness_exp}

\subsubsection{Synthetic Data Experiments}

Tables \ref{figtab:treelikeness_synthetic_cpr_suppl} and \ref{figtab:treelikeness_synthetic_dogd_suppl} show results on the synthetic tree-like dataset using a wide range of CPR and DoGD hyperparameters ($R$ and $a,b$, respectively).

\begin{table}[htbp]
\setlength{\tabcolsep}{4pt}
\centering
% \small
% \scriptsize
\begin{tabular}{c||cc|cc}
\multicolumn{1}{c}{} & \multicolumn{2}{c|}{\textbf{ViT-H}} &  \multicolumn{2}{c}{\textbf{ViT-B}}  \\
\toprule
$R$ & $\tau$ & $\rho$ & $\tau$ & $\rho$ \\
\midrule
$1$ & $-0.82$ & $-0.96$ & $-0.79$ & $-0.94$ \\
$3$ & $-0.80$ & $-0.95$ & $-0.76$ & $-0.94$ \\
$5$ & $-0.77$ & $-0.93$ & $-0.75$ & $-0.93$ \\
$7$ & $-0.74$ & $-0.92$ & $-0.73$ & $-0.91$ \\
$9$ & $-0.72$ & $-0.91$ & $-0.70$ & $-0.90$ \\
$11$ & $-0.71$ & $-0.90$ & $-0.70$ & $-0.89$ \\
\bottomrule
\end{tabular}
\caption{Rank correlation (Kendall $\tau$ and Spearman $\rho$) between SAM prediction IoU and object tree-likeness as measured by \textbf{CPR} with different hyperparameter values $R$ on the \textbf{synthetic dataset}, to supplement Fig. \ref{figtab:treelikeness_synthetic}.}
\label{figtab:treelikeness_synthetic_cpr_suppl}
\end{table}

\begin{table}[htbp]
\setlength{\tabcolsep}{4pt}
\centering
% \small
% \scriptsize
\begin{tabular}{cc||cc|cc}
\multicolumn{2}{c}{} & \multicolumn{2}{c|}{\textbf{ViT-H}} &  \multicolumn{2}{c}{\textbf{ViT-B}}  \\
\toprule
$a$ & $b$ & $\tau$ & $\rho$ & $\tau$ & $\rho$ \\
\midrule
$63$  & $3$  & $0.42$ & $0.60$ & $0.48$ & $0.66$ \\
$63$  & $7$  & $0.45$ & $0.63$ & $0.49$ & $0.68$ \\
$63$  & $15$ & $0.44$ & $0.64$ & $0.48$ & $0.68$ \\
$63$  & $31$ & $0.45$ & $0.66$ & $0.49$ & $0.71$ \\
$127$ & $3$  & $0.61$ & $0.80$ & $0.66$ & $0.84$ \\
$127$ & $7$  & $0.63$ & $0.81$ & $0.67$ & $0.84$ \\
$127$ & $15$ & $0.62$ & $0.81$ & $0.66$ & $0.84$ \\
$127$ & $31$ & $0.60$ & $0.80$ & $0.64$ & $0.83$ \\
$255$ & $3$  & $0.66$ & $0.84$ & $0.68$ & $0.85$ \\
$255$ & $7$  & $0.66$ & $0.84$ & $0.68$ & $0.86$ \\
$255$ & $15$ & $0.65$ & $0.84$ & $0.67$ & $0.85$ \\
$255$ & $31$ & $0.59$ & $0.79$ & $0.60$ & $0.80$ \\
\bottomrule
\end{tabular}
\caption{Rank correlation (Kendall $\tau$ and Spearman $\rho$) between SAM prediction IoU and object tree-likeness as measured by \textbf{DoGD} with different hyperparameter values $a,b$ on the \textbf{synthetic dataset}, to supplement Fig. \ref{figtab:treelikeness_synthetic}.}
\label{figtab:treelikeness_synthetic_dogd_suppl}
\end{table}

\subsubsection{Real Data Experiments}

Tables \ref{tab:treelikeness_realDIS_cpr_suppl}, \ref{tab:treelikeness_realiShape_cpr_suppl}, \ref{tab:treelikeness_realDIS_dogd_suppl}, and \ref{tab:treelikeness_realiShape_dogd_suppl} show results on DIS and iShape using a wide range of CPR and DoGD hyperparameters ($R$ and $a,b$, respectively).

\begin{table}[htbp]
\setlength{\tabcolsep}{4pt}
\centering
% \small
% \scriptsize
\begin{tabular}{c||cc|cc}
\multicolumn{1}{c}{} & \multicolumn{2}{c|}{\textbf{ViT-H}} &  \multicolumn{2}{c}{\textbf{ViT-B}}  \\
\toprule
$R$ & $\tau$ & $\rho$ & $\tau$ & $\rho$ \\
\midrule
$1$  & $-0.56$ & $-0.73$ & $-0.61$ & $-0.79$ \\
$3$  & $-0.55$ & $-0.72$ & $-0.61$ & $-0.79$ \\
$5$  & $-0.59$ & $-0.76$ & $-0.63$ & $-0.81$ \\
$7$  & $-0.53$ & $-0.7$  & $-0.59$ & $-0.77$ \\
$9$  & $-0.53$ & $-0.71$ & $-0.57$ & $-0.75$ \\
$11$ & $-0.51$ & $-0.69$ & $-0.56$ & $-0.74$ \\
\bottomrule
\end{tabular}
\caption{Rank correlation (Kendall $\tau$ and Spearman $\rho$) between SAM prediction IoU and object tree-likeness as measured by \textbf{CPR} with different hyperparameter values $R$ on \textbf{DIS}, to supplement Fig. \ref{figtab:treelikeness_results_real}.}
\label{tab:treelikeness_realDIS_cpr_suppl}
\end{table}

\begin{table*}[htbp]
\setlength{\tabcolsep}{4pt}
\centering
\small
%\scriptsize
\begin{tabular}{c|l||cc|cc|cc|cc|cc|cc}
\multicolumn{2}{c}{} & \multicolumn{2}{c|}{antenna} & \multicolumn{2}{c|}{branch} & \multicolumn{2}{c|}{fence} & \multicolumn{2}{c|}{hanger} & \multicolumn{2}{c|}{log} & \multicolumn{2}{c}{wire}\\
\toprule

$R$ & \textbf{SAM Enc.} & $\tau$ & $\rho$ & $\tau$ & $\rho$ & $\tau$ & $\rho$ & $\tau$ & $\rho$ & $\tau$ & $\rho$ & $\tau$ & $\rho$  \\
\midrule
\multirow{2}{*}{1} & ViT-H              & $-0.18$ & $-0.25$ & $-0.63$ & $-0.83$ & $-0.68$ & $-0.88$ & $-0.30$ & $-0.43$ & $-0.60$ & $-0.78$ & $-0.08$ & $-0.12$ \\
 & ViT-B & $-0.19$ & $-0.29$ & $-0.64$ & $-0.83$ & $-0.68$ & $-0.88$ & $-0.47$ & $-0.66$ & $-0.61$ & $-0.79$ & $-0.12$ & $-0.18$ \\ \midrule
\multirow{2}{*}{3} & ViT-H              & $-0.20$ & $-0.30$ & $-0.63$ & $-0.83$ & $-0.65$ & $-0.85$ & $-0.28$ & $-0.40$ & $-0.61$ & $-0.79$ & $-0.07$ & $-0.10$ \\
 & ViT-B & $-0.17$ & $-0.26$ & $-0.64$ & $-0.83$ & $-0.65$ & $-0.85$ & $-0.48$ & $-0.66$ & $-0.61$ & $-0.79$ & $-0.13$ & $-0.19$ \\ \midrule
\multirow{2}{*}{5} & ViT-H              & $-0.13$ & $-0.21$ & $-0.61$ & $-0.81$ & $-0.61$ & $-0.81$ & $-0.25$ & $-0.37$ & $-0.60$ & $-0.79$ & $-0.06$ & $-0.10$ \\
 & ViT-B & $-0.18$ & $-0.27$ & $-0.63$ & $-0.82$ & $-0.63$ & $-0.83$ & $-0.44$ & $-0.62$ & $-0.62$ & $-0.80$ & $-0.12$ & $-0.17$ \\ \midrule
\multirow{2}{*}{7} & ViT-H              & $-0.16$ & $-0.23$ & $-0.57$ & $-0.77$ & $-0.54$ & $-0.74$ & $-0.20$ & $-0.29$ & $-0.60$ & $-0.79$ & $-0.06$ & $-0.10$ \\
 & ViT-B & $-0.15$ & $-0.24$ & $-0.59$ & $-0.77$ & $-0.55$ & $-0.75$ & $-0.39$ & $-0.55$ & $-0.62$ & $-0.81$ & $-0.10$ & $-0.15$ \\ \midrule
\multirow{2}{*}{9} & ViT-H              & $-0.16$ & $-0.23$ & $-0.52$ & $-0.72$ & $-0.47$ & $-0.65$ & $-0.13$ & $-0.19$ & $-0.59$ & $-0.78$ & $-0.05$ & $-0.08$ \\
 & ViT-B & $-0.16$ & $-0.22$ & $-0.57$ & $-0.76$ & $-0.49$ & $-0.67$ & $-0.29$ & $-0.42$ & $-0.62$ & $-0.81$ & $-0.10$ & $-0.15$ \\ \midrule
\multirow{2}{*}{11} & ViT-H               & $-0.08$ & $-0.13$ & $-0.49$ & $-0.67$ & $-0.51$ & $-0.70$ & $-0.02$ & $-0.03$ & $-0.60$ & $-0.78$ & $-0.03$ & $-0.04$ \\
 & ViT-B & $-0.12$ & $-0.16$ & $-0.54$ & $-0.73$ & $-0.55$ & $-0.75$ & $-0.21$ & $-0.28$ & $-0.64$ & $-0.82$ & $-0.07$ & $-0.11$ \\
\bottomrule
\end{tabular}

\caption{Rank correlation (Kendall $\tau$ and Spearman $\rho$) between SAM prediction IoU and object tree-likeness as measured by \textbf{CPR} with different hyperparameter values $R$ on \textbf{iShape}, to supplement Fig. \ref{figtab:treelikeness_results_real}.}
\label{tab:treelikeness_realiShape_cpr_suppl}
\end{table*}

\begin{table}[htbp]
\setlength{\tabcolsep}{4pt}
\centering
% \small
% \scriptsize
\begin{tabular}{cc||cc|cc}
\multicolumn{2}{c}{} & \multicolumn{2}{c|}{\textbf{ViT-H}} &  \multicolumn{2}{c}{\textbf{ViT-B}}  \\
\toprule
$a$ & $b$ & $\tau$ & $\rho$ & $\tau$ & $\rho$ \\
\midrule
$63$ & $3$  & $0.63$ & $0.83$ & $0.57$ & $0.78$ \\
$63$ & $7$  & $0.59$ & $0.79$ & $0.57$ & $0.77$ \\
$63$ & $15$ & $0.54$ & $0.72$ & $0.50$ & $0.67$ \\
$63$ & $31$ & $0.45$ & $0.62$ & $0.39$ & $0.56$ \\
$127$ & $3$ & $0.58$ & $0.77$ & $0.45$ & $0.64$ \\
$127$ & $7$ & $0.58$ & $0.77$ & $0.49$ & $0.66$ \\
$127$ & $15$ & $0.51$ & $0.67$ & $0.42$ & $0.57$ \\
$127$ & $31$ & $0.36$ & $0.50$ & $0.26$ & $0.39$ \\
$255$ & $3$ & $0.53$ & $0.71$ & $0.42$ & $0.59$ \\ 
$255$ & $7$ & $0.53$ & $0.71$ & $0.42$ & $0.59$ \\ 
$255$ & $15$ & $0.47$ & $0.66$ & $0.36$ & $0.52$ \\
$255$ & $31$ & $0.29$ & $0.43$ & $0.19$ & $0.29$ \\
\bottomrule
\end{tabular}
\caption{Rank correlation (Kendall $\tau$ and Spearman $\rho$) between SAM prediction IoU and object tree-likeness as measured by \textbf{DoGD} with different hyperparameter values $a,b$ on \textbf{DIS}, to supplement Fig. \ref{figtab:treelikeness_results_real}.}
\label{tab:treelikeness_realDIS_dogd_suppl}
\end{table}

\begin{table*}[htbp]
\setlength{\tabcolsep}{4pt}
\centering
\small
%\scriptsize
\begin{tabular}{cc|l||cc|cc|cc|cc|cc|cc}
\multicolumn{3}{c}{} & \multicolumn{2}{c|}{antenna} & \multicolumn{2}{c|}{branch} & \multicolumn{2}{c|}{fence} & \multicolumn{2}{c|}{hanger} & \multicolumn{2}{c|}{log} & \multicolumn{2}{c}{wire}\\
\toprule

$a$ & $b$ & \textbf{SAM Enc.} & $\tau$ & $\rho$ & $\tau$ & $\rho$ & $\tau$ & $\rho$ & $\tau$ & $\rho$ & $\tau$ & $\rho$ & $\tau$ & $\rho$  \\
\midrule
\multirow{2}{*}{63}  & \multirow{2}{*}{3}  & ViT-H & $0.39$  & $0.55$  & $0.62$ & $0.82$ & $0.65$  & $0.85$  & $-0.15$ & $-0.23$ & $0.23$ & $0.35$ & $0.35$  & $0.50$  \\
 & & ViT-B & $0.25$  & $0.35$  & $0.62$ & $0.81$ & $0.60$  & $0.81$  & $-0.07$ & $-0.11$ & $0.17$ & $0.25$ & $0.36$  & $0.51$  \\\midrule
\multirow{2}{*}{63}  & \multirow{2}{*}{7}  & ViT-H & $0.28$  & $0.38$  & $0.61$ & $0.81$ & $0.62$  & $0.83$  & $-0.13$ & $-0.21$ & $0.22$ & $0.33$ & $0.34$  & $0.49$  \\
 & & ViT-B & $0.13$  & $0.18$  & $0.61$ & $0.81$ & $0.60$  & $0.80$  & $-0.07$ & $-0.11$ & $0.14$ & $0.22$ & $0.33$  & $0.48$  \\\midrule
\multirow{2}{*}{63}  & \multirow{2}{*}{15} & ViT-H & $0.10$  & $0.12$  & $0.59$ & $0.78$ & $0.59$  & $0.80$  & $-0.05$ & $-0.07$ & $0.19$ & $0.30$ & $0.29$  & $0.43$  \\
 & & ViT-B & $0.04$  & $0.03$  & $0.59$ & $0.77$ & $0.56$  & $0.77$  & $-0.01$ & $-0.01$ & $0.09$ & $0.16$ & $0.25$  & $0.36$  \\\midrule
\multirow{2}{*}{63}  & \multirow{2}{*}{31} & ViT-H & $0.06$  & $0.04$  & $0.55$ & $0.74$ & $0.46$  & $0.65$  & $0.07$  & $0.10$  & $0.14$ & $0.26$ & $0.09$  & $0.13$  \\
 & & ViT-B & $-0.11$ & $-0.17$ & $0.56$ & $0.74$ & $0.45$  & $0.64$  & $0.07$  & $0.11$  & $0.02$ & $0.08$ & $0.02$  & $0.03$  \\\midrule
\multirow{2}{*}{127} & \multirow{2}{*}{3}  & ViT-H & $0.23$  & $0.30$  & $0.52$ & $0.70$ & $0.46$  & $0.65$  & $0.52$  & $0.72$  & $0.30$ & $0.46$ & $0.08$  & $0.11$  \\
 & & ViT-B & $0.12$  & $0.14$  & $0.51$ & $0.69$ & $0.45$  & $0.64$  & $0.50$  & $0.70$  & $0.13$ & $0.22$ & $0.06$  & $0.08$  \\\midrule
\multirow{2}{*}{127} & \multirow{2}{*}{7}  & ViT-H & $0.23$  & $0.26$  & $0.48$ & $0.65$ & $0.41$  & $0.59$  & $0.51$  & $0.70$  & $0.29$ & $0.46$ & $-0.01$ & $-0.01$ \\
 & & ViT-B & $0.10$  & $0.09$  & $0.51$ & $0.68$ & $0.41$  & $0.59$  & $0.49$  & $0.68$  & $0.13$ & $0.22$ & $-0.03$ & $-0.05$ \\\midrule

\multirow{2}{*}{$127$} & \multirow{2}{*}{$15$} & ViT-H & $0.14$  & $0.17$  & $0.43$ & $0.59$ & $0.38$  & $0.55$  & $0.50$  & $0.69$  & $0.28$ & $0.43$ & $-0.09$ & $-0.12$ \\
 & & ViT-B & $0.07$  & $0.03$  & $0.45$ & $0.62$ & $0.36$  & $0.52$  & $0.48$  & $0.67$  & $0.12$ & $0.20$ & $-0.14$ & $-0.20$ \\\midrule
\multirow{2}{*}{$127$} & \multirow{2}{*}{$31$} & ViT-H & $0.05$  & $0.05$  & $0.36$ & $0.50$ & $0.08$  & $0.12$  & $0.48$  & $0.67$  & $0.30$ & $0.45$ & $-0.15$ & $-0.22$ \\
 & & ViT-B & $0.04$  & $0.00$  & $0.38$ & $0.52$ & $0.12$  & $0.18$  & $0.46$  & $0.65$  & $0.12$ & $0.20$ & $-0.21$ & $-0.31$ \\\midrule
\multirow{2}{*}{$255$} & \multirow{2}{*}{$3$} & ViT-H & $0.04$  & $0.06$  & $0.33$ & $0.46$ & $-0.20$ & $-0.30$ & $0.44$  & $0.61$  & $0.52$ & $0.72$ & $-0.17$ & $-0.25$ \\
 & & ViT-B & $0.21$  & $0.31$  & $0.30$ & $0.42$ & $-0.13$ & $-0.20$ & $0.42$  & $0.59$  & $0.42$ & $0.60$ & $-0.23$ & $-0.34$ \\\midrule
\multirow{2}{*}{$255$} & \multirow{2}{*}{$7$} & ViT-H & $0.02$  & $0.02$  & $0.29$ & $0.41$ & $-0.25$ & $-0.35$ & $0.43$  & $0.60$  & $0.51$ & $0.71$ & $-0.20$ & $-0.29$ \\
 & & ViT-B & $0.19$  & $0.27$  & $0.26$ & $0.36$ & $-0.18$ & $-0.26$ & $0.42$  & $0.58$  & $0.41$ & $0.59$ & $-0.25$ & $-0.36$ \\\midrule
\multirow{2}{*}{$255$} & \multirow{2}{*}{$15$} & ViT-H & $-0.04$ & $-0.06$ & $0.24$ & $0.34$ & $-0.31$ & $-0.45$ & $0.43$  & $0.61$  & $0.50$ & $0.70$ & $-0.20$ & $-0.30$ \\
 & & ViT-B & $0.21$  & $0.29$  & $0.20$ & $0.28$ & $-0.26$ & $-0.39$ & $0.40$  & $0.56$  & $0.40$ & $0.58$ & $-0.28$ & $-0.40$ \\\midrule
\multirow{2}{*}{$255$} & \multirow{2}{*}{$31$} & ViT-H & $-0.11$ & $-0.14$ & $0.11$ & $0.16$ & $-0.47$ & $-0.65$ & $0.43$  & $0.59$  & $0.46$ & $0.66$ & $-0.23$ & $-0.33$ \\
 & & ViT-B & $0.13$  & $0.20$  & $0.07$ & $0.09$ & $-0.42$ & $-0.60$ & $0.39$  & $0.55$  & $0.38$ & $0.55$ & $-0.30$ & $-0.44$ \\

\bottomrule
\end{tabular}

\caption{Rank correlation (Kendall $\tau$ and Spearman $\rho$) between SAM prediction IoU and object tree-likeness as measured by \textbf{DoGD} with different hyperparameter values $R$ on \textbf{iShape}, to supplement Fig. \ref{figtab:treelikeness_results_real}.}
\label{tab:treelikeness_realiShape_dogd_suppl}
\end{table*}

\subsection{Object Textural Separability Experiments}
\label{app:sep_exp}

Table \ref{tab:sep_real_supp} show the textural separability experiment results on real data (iShape and Plittersdorf) using various hyperparameters and models for the classifier $g$.

\begin{table*}[htbp]
\setlength{\tabcolsep}{4pt}
\centering
% \small
\scriptsize
\begin{tabular}{l|l||cc|cc|cc|cc|cc|cc||cc}
\multicolumn{2}{c}{} & \multicolumn{12}{c}{\textbf{iShape}} & \multicolumn{2}{c}{\textbf{}} \\
\multicolumn{2}{c}{} & \multicolumn{2}{c|}{antenna} & \multicolumn{2}{c|}{branch} & \multicolumn{2}{c|}{fence} & \multicolumn{2}{c|}{hanger} & \multicolumn{2}{c|}{log} & \multicolumn{2}{c||}{wire} & \multicolumn{2}{c}{\textbf{Plittersdorf}} \\
\toprule

\textbf{\tworow{Weak Classifier}{Model}} & \textbf{SAM Enc.} & $\tau$ & $\rho$ & $\tau$ & $\rho$ & $\tau$ & $\rho$ & $\tau$ & $\rho$ & $\tau$ & $\rho$ & $\tau$ & $\rho$ & $\tau$ & $\rho$ \\
\midrule
\multirow{2}{*}{\tworow{Logistic,}{$C=2$}} & ViT-H & $0.44$ & $0.59$ & $0.50$ & $0.68$ & $0.65$ & $0.85$ & $0.63$ & $0.83$ & $0.33$ & $0.48$ & $0.49$ & $0.67$ & $0.26$ & $0.38$ \\
& ViT-B & $0.46$ & $0.61$ & $0.56$ & $0.75$ & $0.67$ & $0.86$ & $0.65$ & $0.84$ & $0.36$ & $0.52$ & $0.55$ & $0.73$ & $0.36$ & $0.50$ \\\midrule
\multirow{2}{*}{\tworow{Logistic,}{$C=1$}} & ViT-H & $0.42$ & $0.58$ & $0.49$ & $0.67$ & $0.64$ & $0.84$ & $0.62$ & $0.81$ & $0.31$ & $0.46$ & $0.49$ & $0.67$ & $0.34$ & $0.49$ \\
& ViT-B & $0.44$ & $0.60$ & $0.55$ & $0.75$ & $0.66$ & $0.85$ & $0.63$ & $0.82$ & $0.35$ & $0.50$ & $0.55$ & $0.72$ & $0.39$ & $0.55$ \\\midrule
\multirow{2}{*}{\tworow{Logistic,}{$C=0.5$}} & ViT-H & $0.39$ & $0.53$ & $0.49$ & $0.67$ & $0.65$ & $0.85$ & $0.60$ & $0.80$ & $0.30$ & $0.43$ & $0.48$ & $0.65$ & $0.33$ & $0.47$ \\
& ViT-B & $0.48$ & $0.65$ & $0.57$ & $0.76$ & $0.66$ & $0.86$ & $0.62$ & $0.81$ & $0.33$ & $0.48$ & $0.52$ & $0.69$ & $0.41$ & $0.56$ \\\midrule
\multirow{2}{*}{\tworow{Random Forest,}{$n_{\mathrm{estimators}}=2$}} & ViT-H & $0.48$ & $0.66$ & $0.45$ & $0.63$ & $0.46$ & $0.64$ & $0.45$ & $0.64$ & $0.19$ & $0.27$ & $0.25$ & $0.36$ & $0.35$ & $0.50$ \\
& ViT-B & $0.50$ & $0.67$ & $0.49$ & $0.68$ & $0.47$ & $0.65$ & $0.50$ & $0.69$ & $0.24$ & $0.34$ & $0.31$ & $0.45$ & $0.41$ & $0.56$ \\\midrule
\multirow{2}{*}{\tworow{Random Forest,}{$n_{\mathrm{estimators}}=4$}} & ViT-H & $0.51$ & $0.69$ & $0.42$ & $0.60$ & $0.54$ & $0.73$ & $0.48$ & $0.67$ & $0.21$ & $0.30$ & $0.30$ & $0.43$ & $0.22$ & $0.31$ \\
& ViT-B & $0.51$ & $0.70$ & $0.48$ & $0.66$ & $0.53$ & $0.72$ & $0.51$ & $0.70$ & $0.22$ & $0.32$ & $0.37$ & $0.52$ & $0.32$ & $0.45$ \\\midrule
\multirow{2}{*}{\tworow{Random Forest,}{$n_{\mathrm{estimators}}=8$}} & ViT-H & $0.48$ & $0.65$ & $0.46$ & $0.63$ & $0.56$ & $0.75$ & $0.50$ & $0.69$ & $0.19$ & $0.28$ & $0.35$ & $0.50$ & $0.22$ & $0.31$ \\
& ViT-B & $0.49$ & $0.67$ & $0.49$ & $0.68$ & $0.55$ & $0.74$ & $0.54$ & $0.73$ & $0.21$ & $0.31$ & $0.40$ & $0.57$ & $0.25$ & $0.35$ \\\midrule
\multirow{2}{*}{\tworow{Random Forest,}{$n_{\mathrm{estimators}}=16$}} & ViT-H & $0.45$ & $0.62$ & $0.48$ & $0.66$ & $0.57$ & $0.77$ & $0.52$ & $0.72$ & $0.20$ & $0.29$ & $0.38$ & $0.53$ & $0.08$ & $0.12$ \\
& ViT-B & $0.58$ & $0.77$ & $0.52$ & $0.70$ & $0.54$ & $0.74$ & $0.54$ & $0.74$ & $0.23$ & $0.33$ & $0.41$ & $0.57$ & $0.15$ & $0.23$ \\

\bottomrule
\end{tabular}
\caption{Rank correlation (Kendall $\tau$ and Spearman $\rho$) between SAM prediction IoU and object textural separability (Algorithm \ref{alg:textural_sep}), on real datasets (iShape and Plittersdorf), using different weak classifier models $g$ and hyperparameters to measure textural separability (to supplement Fig. \ref{figtab:sep_real}).}
\label{tab:sep_real_supp}
\end{table*}

\section{Additional Experimental Details}

\subsection{SFM Prompting Strategies}
\label{app:prompting}

In addition to the SFM prompting details presented in Sec. \ref{sec:prompting}, all real and style-transferred (NST) images are prompted with (1) a tight bounding-box and (2) several positive and/or negative prompts randomly sampled from either the foreground or background of the object mask, respectively, with respective counts $n_{pos}$ and $n_{neg}$. For iShape, we use $n_{pos}=n_{neg}=5$; for DIS and MOSE, we use $n_{pos}=5$ and $n_{neg}=10$ due to the complexity of the objects. For Plittersdorf and NST images, we simply use $n_{pos}=0$ and $n_{neg}=2$, due to the objects typically possessing simple shapes.

\subsection{Neural Style Transfer Experiments}
\label{app:NSTmodels}

\paragraph{Models.} For the NST experiments (Sec. \ref{sec:exp_nst}), the inpainting model used is Runway's stable diffusion inpainting pipeline, \texttt{fp16} variant. The NST model itself is an implementation of \cite{gatys2016image} based on \url{https://github.com/pytorch/tutorials/blob/main/advanced_source/neural_style_tutorial.py}, using typical settings of a standard VGG19 CNN with content layer of \texttt{conv\_4}, style layers of \texttt{conv\_i} for \texttt{i = 1, 2, 3, 4, 5}, and ImageNet normalization.

\paragraph{Style transfer intensity.} As mentioned in Sec. \ref{sec:exp_nst}, we perform style transfer experiments for a monotonically-increasing range of eight degrees of style transfer intensity. This is created by defining content and style weights $\lambda_c$ and $\lambda_s$, respectively for the NST algorithm according to 
\begin{equation}
    \lambda_c = \frac{1}{1+\alpha} \quad \text{and} \quad \lambda_s = \frac{\alpha}{1+\alpha},
\end{equation}
where $\alpha$ ranges linearly from $\alpha=1$ to $\alpha=4,000$ with eight equally-spaced values, representing the degree of style transfer intensity.

\paragraph{Object/image altering procedure.} The procedure that creates the ``altered'' version of objects via non-affine transformations is detailed as follows.
\begin{enumerate}
    \item Apply a non-affine transformation created using the \texttt{albumentations} Python library \cite{info11020125} to both the image and the object/mask, defined shortly.
    \item Clean up unexpected obsolete regions.
    \item Properly align the location of the distorted object.
    \item Remove small isolated regions of the object.
\end{enumerate}

In Python code, this is implemented as follows, given an input image NumPy array \texttt{raw\_img\_arr} and corresponding binary object mask \texttt{raw\_msk\_arr}:

\begin{lstlisting}[language=Python, basicstyle=\ttfamily\tiny, tabsize=1]
import torch
import numpy as np
from PIL import Image
import albumentations as A
import torchvision.transforms as transforms
from skimage.morphology import dilation,label, \
disk, remove_small_holes, remove_small_objects, \
opening, closing

shape_transformation = A.Compose([
    A.ElasticTransform(
        alpha=2500, sigma=24, alpha_affine=0, 
        interpolation=1, border_mode=0, value=0, 
        mask_value=0, always_apply=True,
        approximate=False, same_dxdy=False, p=1.0),
    A.GridDistortion(
        num_steps=20, distort_limit=.99, 
        interpolation=1, border_mode=0, value=0, 
        mask_value=0, normalized=False,
        always_apply=False, p=1.0)]
)

imsize = 512 
loader = transforms.Compose([
    transforms.Resize(imsize),  # scale imported image
    transforms.CenterCrop(imsize),
    transforms.ToTensor()])

unloader = transforms.ToPILImage() 

def shift_array(array, shift_spec):
    assert len(shift_spec) == 2 and \
        isinstance(array,np.ndarray)
    
    shift_corrected_arr = array.copy()
    if shift_spec[0] != 0:
        shift = shift_spec[0]
        shift_corrected_arr = np.concatenate(
            [shift_corrected_arr[shift:],
            shift_corrected_arr[:shift]],
            axis=0)
    
    if shift_spec[1] != 0:
        shift = shift_spec[1]
        shift_corrected_arr = np.concatenate(
            [shift_corrected_arr[:,shift:],
            shift_corrected_arr[:,:shift]],
            axis=1)
    
    return shift_corrected_arr

def center_correction(raw_msk, distorted_msk):
    assert raw_msk.ndim == 4 and \
        distorted_msk.ndim == 3
    pos_idx = torch.argwhere(raw_msk[0,0])
    _raw_msk_center = (pos_idx.min(dim=0).values + \
        pos_idx.max(dim=0).values)//2
    pos_idx = torch.argwhere(distorted_msk[0])
    _distorted_msk_center = (pos_idx.min(dim=0).values + \
        pos_idx.max(dim=0).values)//2
    
    position_correction = _distorted_msk_center -_raw_msk_center  
    corrected_distorted_msk = distorted_msk.squeeze()
    print(corrected_distorted_msk.shape)
    assert position_correction.ndim == 1 and len(position_correction) == 2
    
    if position_correction[0] !=0:
        shift = position_correction[0]
        
        corrected_distorted_msk = torch.cat(
            [corrected_distorted_msk[shift:],
            corrected_distorted_msk[:shift]],
            dim=0)
    if position_correction[1] !=0:
        shift = position_correction[1]
        corrected_distorted_msk = torch.cat(
            [corrected_distorted_msk[:,shift:],
            corrected_distorted_msk[:,:shift]],
            dim=1)
    return corrected_distorted_msk
    
def get_distorted_ImageAndMask(trans, img, *msks):

    # format check
    assert all([msk.ndim == 2 for msk in msks])
    
    # apply transformation to both image and msk 
    transformed = trans(image=img, masks = msks)
    distorted_img_shifted = transformed['image']
    distorted_msks_shifted = transformed['masks']

    #clean up the unexpected obsolete regions
    ## get area excluding border
    positive_region_map = label(distorted_msks_shifted[1]) 
    label_ids, label_cnts = np.unique(positive_region_map,return_counts=True)
    max_id = label_ids[1:][np.argmax(label_cnts[1:])]
    eligible_region_map = np.where(positive_region_map==max_id, 1, 0) 
    
    distorted_msk_shifted = distorted_msks_shifted[0]*eligible_region_map
    distorted_msk_shifted = remove_small_objects(remove_small_holes(
        closing(opening(distorted_msk_shifted>0, disk(2)),disk(2))
        ))
    
    #find center shift as a results of transformation
    pos_idx = np.argwhere(msks[0]).squeeze()
    _raw_msk_center = (pos_idx.min(0)+pos_idx.max(0))//2
    pos_idx = np.argwhere(distorted_msk_shifted).squeeze()
    _distorted_msk_center = (pos_idx.min(0)+pos_idx.max(0))//2
    shift_spec = _distorted_msk_center -_raw_msk_center  

    #shift the arrays so that the obj location aligns
    distorted_img = shift_array(distorted_img_shifted,shift_spec)
    distorted_msk = shift_array(distorted_msk_shifted,shift_spec)


    # remove relatively small regions
    distorted_msk_labeled = label(distorted_msk)
    unique_ids, id_cnts = np.unique(
        distorted_msk_labeled, return_counts = True)
    area_distribution = id_cnts[1:].astype('float')/id_cnts[1:].sum()
    
    for area_id, area_rate in zip(unique_ids[1:], area_distribution):
        if area_rate < 0.125:
            distorted_msk_labeled[distorted_msk_labeled==area_id] = 0
    
    distorted_msk_arr = distorted_msk_labeled>0

    return distorted_img, distorted_msk_arr

distorted_img_arr, distorted_msk_arr = get_distorted_ImageAndMask(
    shape_transformation,raw_img_arr,raw_msk_arr,raw_msk_arr.copy())
\end{lstlisting}

\subsection{Fine-Tuning Experiment Details}
\label{app:finetune}

Here we provide all details for the SAM fine-tuning experiments of Section \ref{sec:finetune}. The loss function that we use is the sum of the Dice and cross-entropy losses (as implemented by MONAI \cite{cardoso2022monai}), optimized via AdamW \cite{loshchilov2018decoupled} with a learning rate of $10^{-5}$ and a weight decay strength of $0.1$. We use a batch size of $16$ and fine-tune for $40$ epochs on the $k$ (image, mask) examples which were sampled from DIS-train using the given selection strategy (see Section \ref{sec:finetune}). All parameters of SAM's mask decoder are fine-tuned, while the image encoder is frozen.

\section{Additional Results}

\subsection{SAM Self-Attention Maps}
\label{app:attentionmaps}

\begin{figure}[htbp!]
    \centering
    \includegraphics[width=0.51\linewidth]{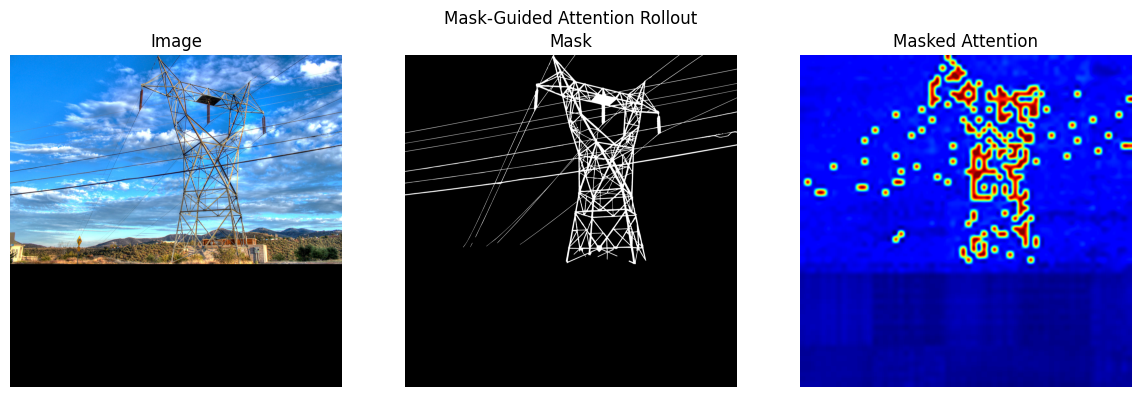}
    \includegraphics[width=0.51\linewidth]{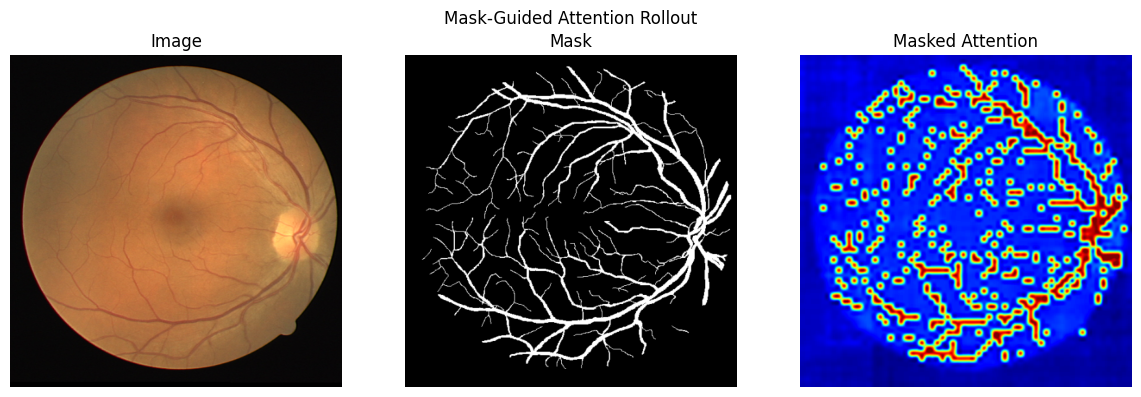}
    \includegraphics[width=0.51\linewidth]{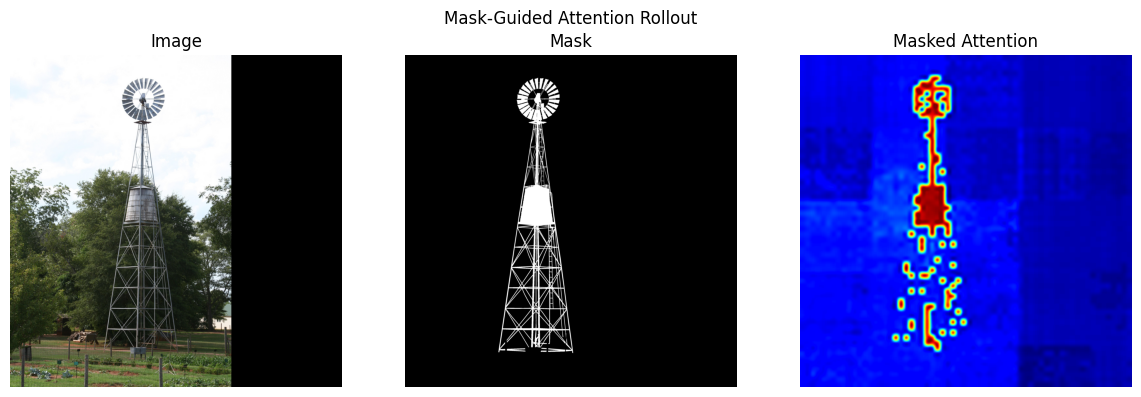}
    \caption{Attention patterns (right column) for example images (left column) with objects of high tree-likeness (shown via ground truth masks in middle column).}
    \label{fig:attnmaps}
\end{figure}

In Fig. \ref{fig:attnmaps} we show examples of SAM's attention patterns for images with objects of high tree-likeness. These were computed via ground truth mask-guided attention rollout \cite{abnar-zuidema-2020-quantifying}.

\subsection{Mechanistic Relationships between Metrics and Model Architectural Choices}

\subsubsection{The Relationship of Vision Transformer Effective Patch Size with CPR}
\label{app:patchsizeinterp}

As was described in Sec. \ref{sec:motivation_treelike}, in theory, when the patch size of an SFM's image encoder is small relative to the typical branch width of a tree-like object, patches can still contain internal pixels, but as patch size grows, CPR implies that boundary pixels dominate every patch, leaving no truly ``interior'' representation. Thus, CPR should directly interact with patch size: larger patches exacerbate the imbalance between boundary and interior content, \ie, the typical CPR of objects will be higher. In this section, we will verify this behavior experimentally.

As the patch sizes of the SFMs (SAM, HQ-SAM and SAM 2) are fixed \textit{a priori}, we propose to instead modify the \textit{effective patch size} by altering the size of the objects themselves. We do this by zooming a given image in (cropping) our out, centered on the object mask, such that the length of the longer edge of the object's tight bounding box is of size $p$ pixels (and similarly modifying the object mask accordingly). In the case of zooming out, we zero-pad the image to set it to the proper size for SFM input, $1024\times 1024$. For example, if $p$ is doubled (the object is made twice as large), the object will be covered by roughly four times as many patches, making the \textit{effective} patch size halved, because more patches are required to cover the same object. In summary, higher $p$ corresponds to smaller effective patch size.

\begin{figure*}[htbp]
     \centering
     \includegraphics[width=.33\linewidth]{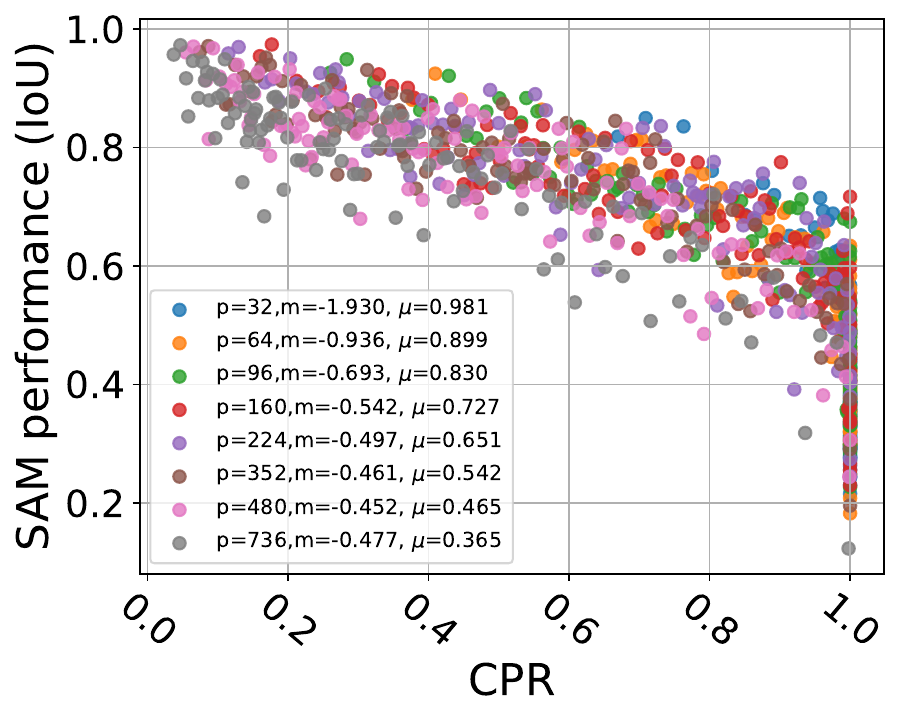}
     \includegraphics[width=.33\linewidth]{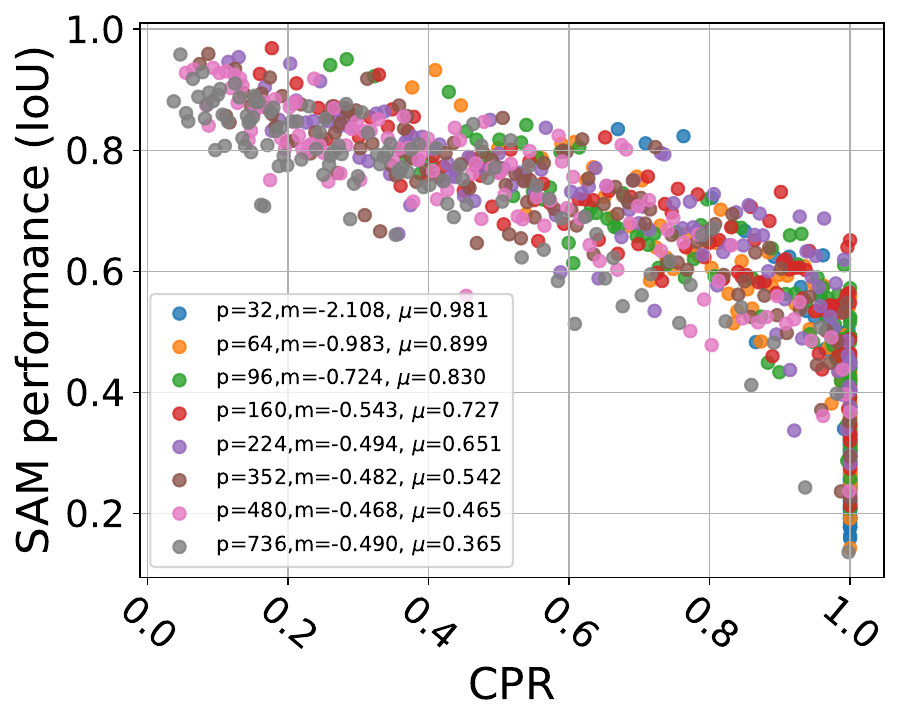}
     \caption{The effect on the relationship between segmentation performance (IoU) vs. CPR (object tree-likeness) of changing the model's effective patch size, increased with decreasing $p$, computed on DIS for the SAM ViT-H (left) and ViT-B (right) variants. Also shown are each plot's legend are average CPR values $\mu$ and estimated linear slope of IoU vs. CPR, for each $p$.}
    \label{fig:patchsizes}
\end{figure*}

In Fig. \ref{fig:patchsizes}, we show how the performance (IoU) vs. object CPR curve on DIS is affected by different choices of $p\in[32,64,96,160,224,352,480,736]$, for both variants of SAM. We note that we discard any datapoint where the zooming-in operation made the object grow outside of the image boundaries. We see that as was predicted, larger patch sizes (smaller $p$) results in objects typically having noticeably higher CPR (higher average CPR $\mu$ for a given $p$, as shown in the plot legends). Moreover, the higher typical CPR results in the model performance on a given object being typically worse.

A potential mechanistic solution to this problem could be to use vision transformer models with smaller patch sizes to perform better on tree-like objects, yet this carries a high computational burden, because the size of the self-attention matrices grows quadratically with the patch size becoming smaller.

\subsubsection{The Relationship of Vision Transformer Attention Map Fragmentation with CPR}
\label{app:attentioninterp}

As we mentioned in Section \ref{sec:motivation_treelike} and showed via a few examples in App. \ref{app:attentionmaps}, attention maps on tree-like objects typically appear to be fragmented and locally-focused, due to responding to local edges independently instead of forming clear global object representations. Here, we will test this empirically by determining if the correlation of the fragmentation of an input image's attention map with the tree-likeness of the object that the image contains. Here, the attention map for a given input image is defined via the attention rollout technique used in App. \ref{app:attentionmaps}.

We measure the fragmentation of a self-attention map $A\in[0,1]^{K\times K}$ via Moran's spatial autocorrelation index $I$ \citep{moran1950notes}. Moran's $I$ is lower for more fragmented/less clustered arrays, so we would expect theoretically that the CPR/tree-likeness of an object's mask is negatively correlated with the Moran's $I$ of the model's computed attention map for the image containing the object.

\begin{figure*}[htbp]
     \centering
     \includegraphics[width=.4\linewidth]{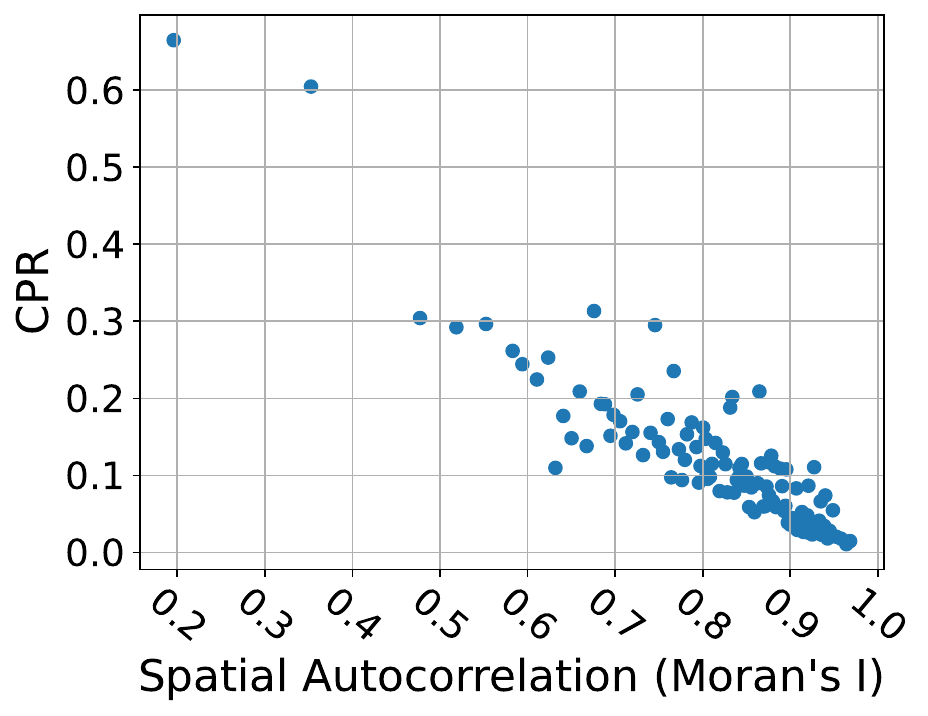}
     \includegraphics[width=.4\linewidth]{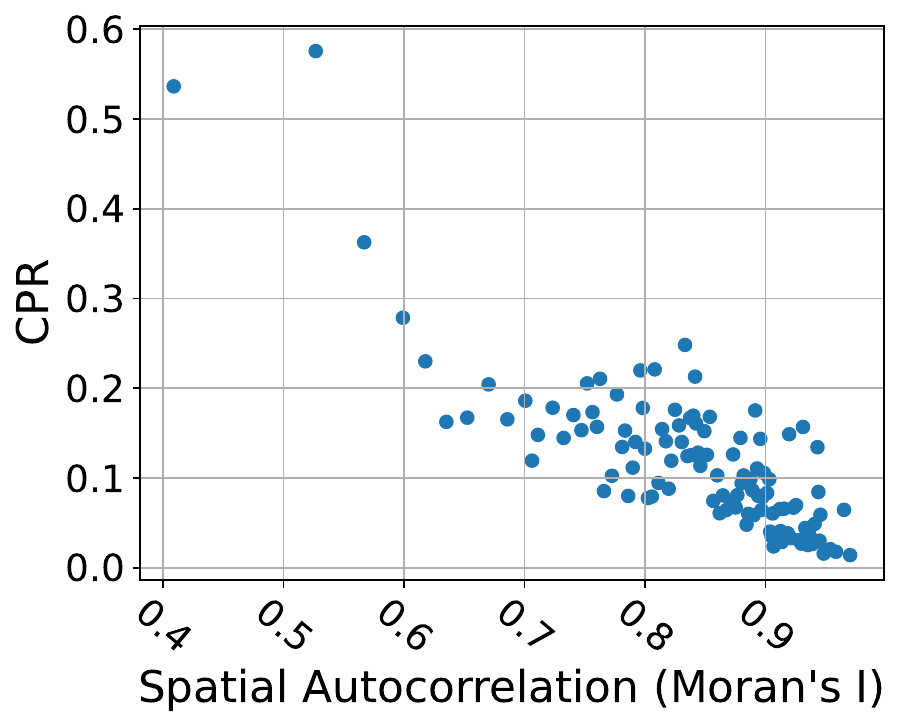}
     \includegraphics[width=.4\linewidth]{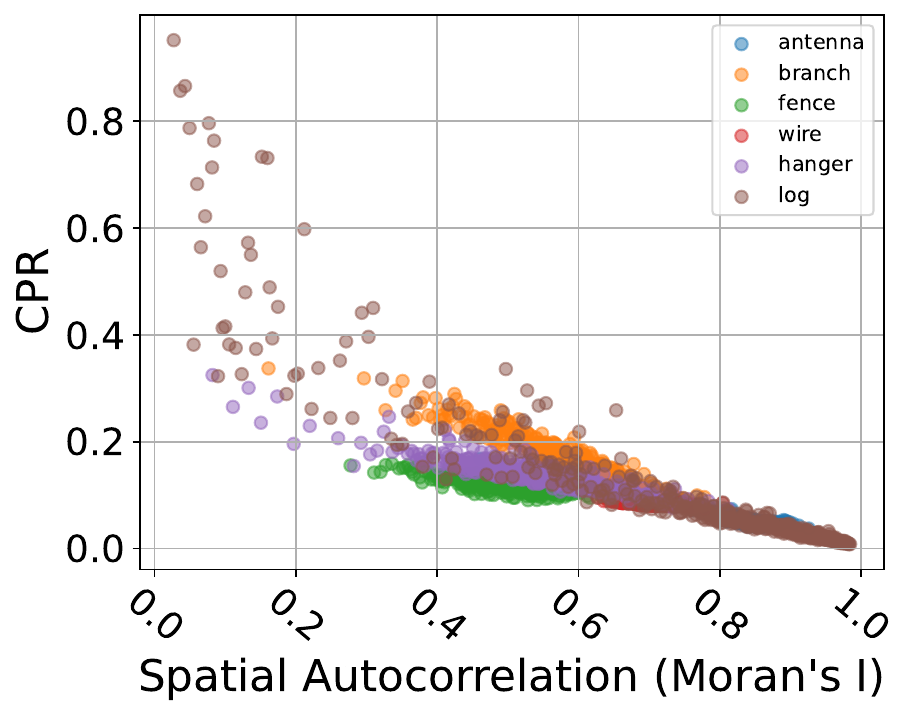}
     \includegraphics[width=.4\linewidth]{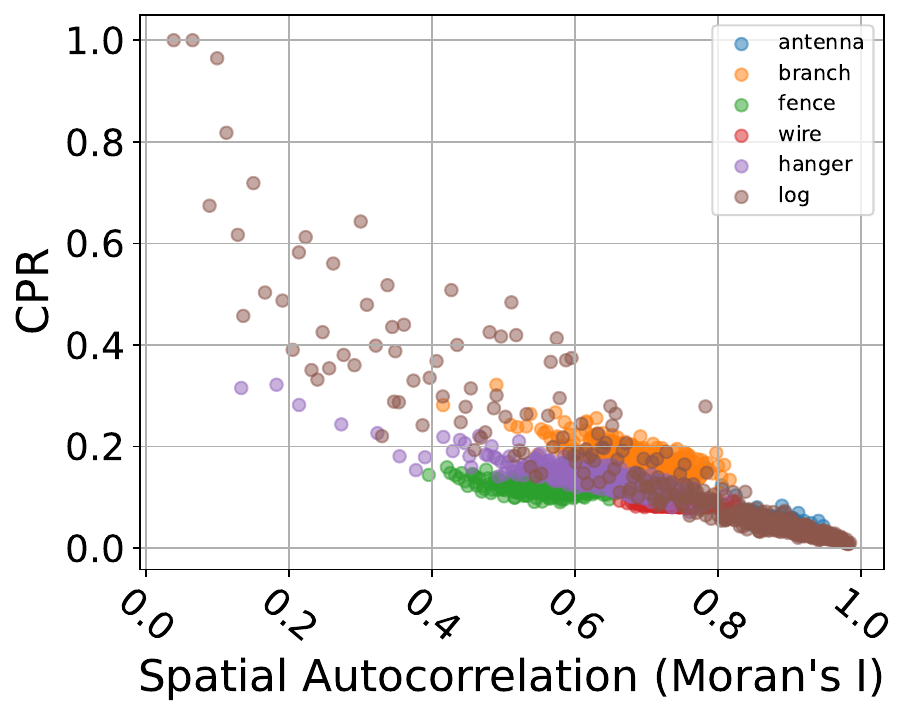}
     \caption{Object tree-likeness (CPR) vs. self-attention map fragmentation (Moran's autocorrelation $I$), on DIS (top row) and iShape (bottom row) for both ViT-H (left column) and ViT-B (right column) variants of SAM.}
    \label{fig:moran_corr}
\end{figure*}

To empirically measure the relationship between an SFM attention map's fragmentation and the tree-likeness of an object to be segmented, we measure the correlation of the Moran's $I$ of an input image's self attention map with the CPR of the input image's associated mask label, computed for both SAM-H and SAM-B on DIS and all classes of iShape, using the same aggregation procedure as in the main experiments (Sec. \ref{sec:treelikeness_exp_real}). The results are shown in Table \ref{tab:moran_corr}, with accompanying plots in Fig. \ref{fig:moran_corr}. Overall, we indeed see a strong negative correlation between object CPR and self-attention map Moran's $I$: objects of higher tree-likeness correspond to more fragmented self-attention maps, as was indicated by our qualitative results in App. \ref{app:attentionmaps}.

\begin{table*}[htbp!]
\setlength{\tabcolsep}{4pt}
\centering
% \small
% \fontsize{8pt}{8pt}\selectfont
% \scriptsize
\begin{tabular}{l||cc|cc}
\multirow{2}{*}{\textbf{SFM Model}} & \multicolumn{2}{c|}{\textbf{DIS}} & \multicolumn{2}{c}{\textbf{iShape (all classes)}} \\
 & $\tau$ & $\rho$ & $\tau$ & $\rho$ \\
\toprule
\textbf{SAM-H} & -0.72 & -0.89 & -0.76 & -0.91 \\
\textbf{SAM-B} & -0.59 & -0.78 & -0.72 & -0.88 \\
\bottomrule
\end{tabular}
\caption{Nonlinear (Spearman $\rho$, Kendall $\tau$) correlations of self-attention map fragmentation (Moran's autocorrelation $I$) with object tree-likeness (CPR).}
\label{tab:moran_corr}
\end{table*}

\subsection{Correlation Between Metrics}
\label{app:treelikeness_corr}

In Fig. \ref{app:treelikeness_corr}, we show the relationship between our proposed tree-likeness metrics CPR and DoGD on all three datasets (with default hyperparameters and ViT-H SAM), quantified by correlations shown in Table \ref{tab:treelike_corr}.

\begin{figure*}[htbp]
     \centering
     \includegraphics[width=.33\linewidth]{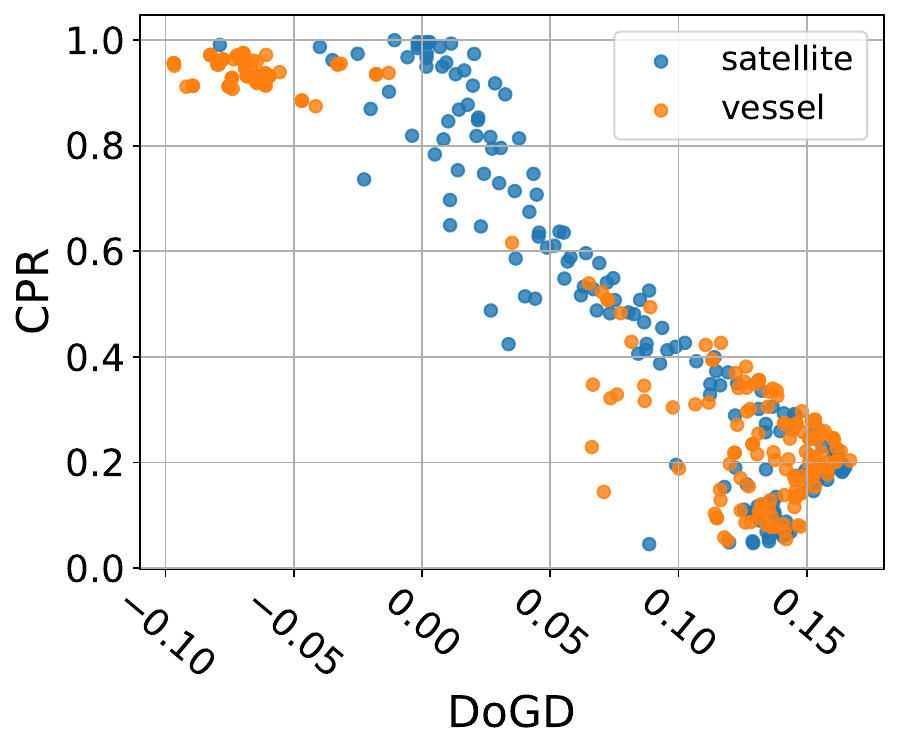}
     \includegraphics[width=.33\linewidth]{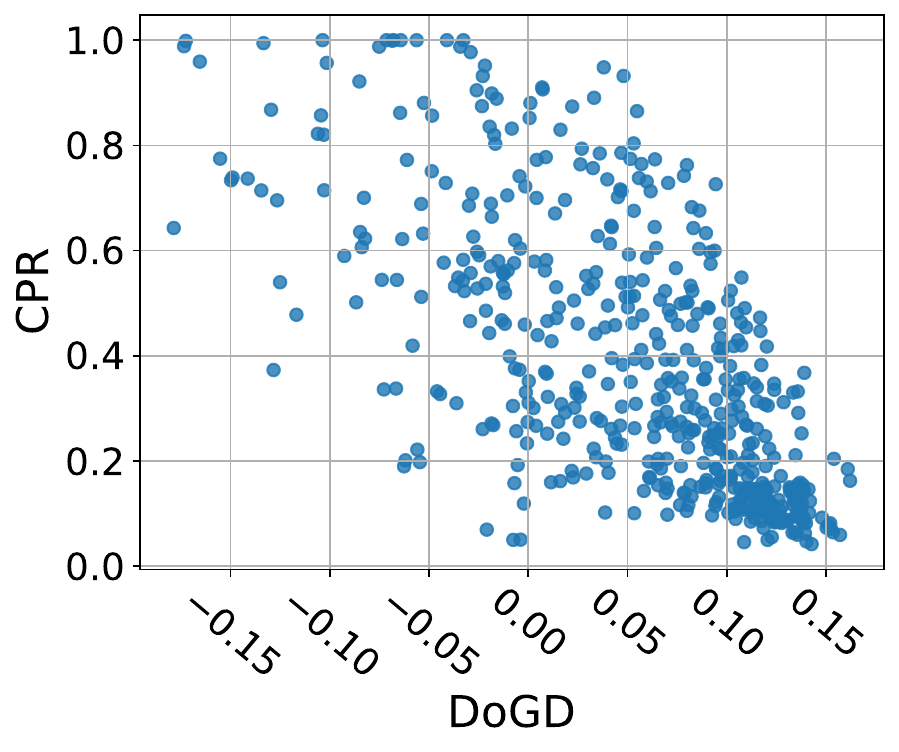}
     \includegraphics[width=.33\linewidth]{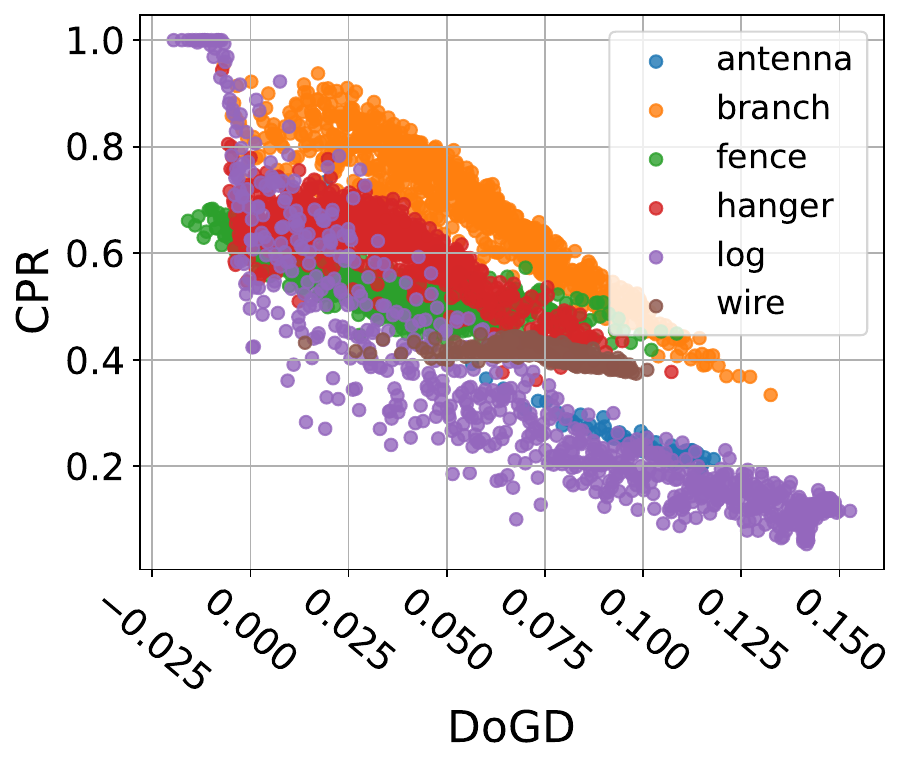}
     \caption{Relationship between tree-likeness metrics CPR and DoGD. From left to right, on synthetic data (Fig. \ref{figtab:treelikeness_synthetic}), DIS, and iShape (Fig. \ref{figtab:treelikeness_results_real}).}
    \label{fig:treelikeness_corr}
\end{figure*}

\begin{table*}[htbp!]
\setlength{\tabcolsep}{4pt}
\centering
% \small
% \fontsize{8pt}{8pt}\selectfont
% \scriptsize
\begin{tabular}{l|c||c||c|c|c|c|c|c|c||c}
\multicolumn{3}{c}{} & \multicolumn{8}{c}{\textbf{iShape}} \\
\multicolumn{1}{c}{} & \textbf{Synth.} & \textbf{DIS} & \multicolumn{1}{c|}{antenna} & \multicolumn{1}{c|}{branch} & \multicolumn{1}{c|}{fence} & \multicolumn{1}{c|}{hanger} & \multicolumn{1}{c|}{log} & \multicolumn{1}{c|}{wire} & \multicolumn{1}{c||}{\textit{all}} & \multicolumn{1}{c||}{\textbf{Avg.}}\\
\toprule
Pearson $r$ & $-0.94$ & $-0.67$ & $-0.92$ & $-0.93$ & $-0.82$ & $-0.81$ & $-0.90$ & $-0.66$ & $-0.79$ & $-0.83$ \\
Spearman $\rho$ & $-0.82$ & $-0.70$ & $-0.91$ & $-0.93$ & $-0.86$ & $-0.76$ & $-0.94$ & $-0.79$ & $-0.76$ & $-0.83$ \\
Kendall $\tau$ & $-0.62$ & $-0.51$ & $-0.75$ & $-0.78$ & $-0.68$ & $-0.58$ & $-0.79$ & $-0.61$ & $-0.57$ & $-0.65$ \\
\bottomrule
\end{tabular}
\caption{Linear (Pearson $r$) and nonlinear (Spearman $\rho$, Kendall $\tau$) correlations between CPR and DoGD on all evaluated datasets.}
\label{tab:treelike_corr}
\end{table*}

\subsection{Effect of Object Line Thickness on Performance}
\label{app:thickness}

In Section \ref{sec:motivation_treelike}, we hypothesized that the noticeable SAM segmentation performance difference between retinal vessel images and satellite road images, two types of objects with branching structures, is due to the former objects typically having more dense and irregularly-spaced branching structures than the latter. However, it is important to consider the possibility that this is simply due to retinal vessels having thinner lines than road objects, which would help evaluate if the proposed tree-likeness metrics (CPR and DoGD) are actually characterizing the true underlying features that drive performance. 

To test this, we performed the same synthetic data experiments as in Section \ref{sec:treelikeness_exp_synth}, but applied transformations which thickened object lines, to see how this affected the overall relationship between measured object tree-likeness and segmentation performance. To do so, between steps 2 and 3 of the procedure used to generate these images (App. \ref{app:treelikesynth_creation}), we:
\begin{enumerate}
    \item Converted the object mask $m_c$ to its skeleton, effectively resulting in a mask consisting of lines with 1-pixel widths.
    \item Apply a dilation transformation with a radius of 4 pixels to this skeleton, resulting in each path/line structure in the mask being strictly 9 pixels wide.
\end{enumerate}

\begin{figure}[htbp!]
    \centering
    \includegraphics[width=0.49\linewidth]{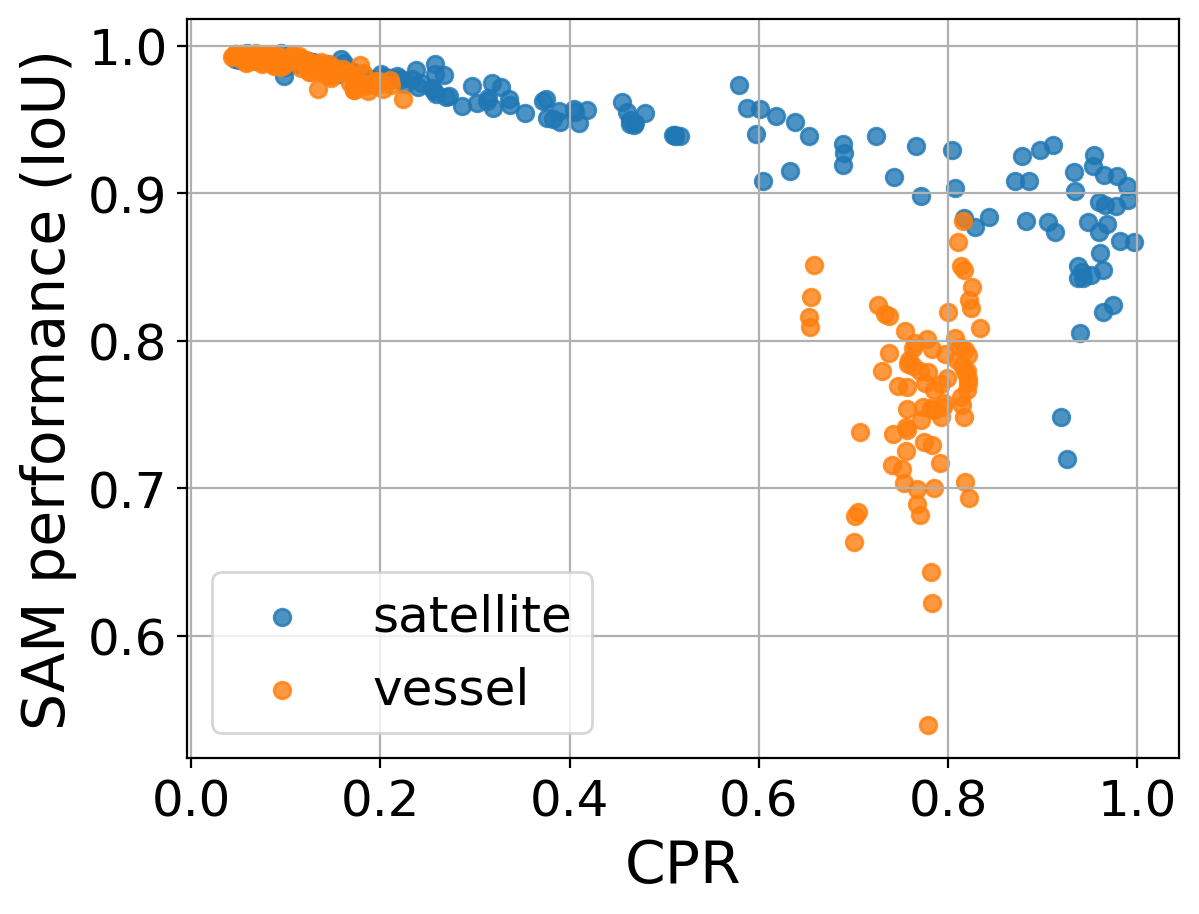}
    \includegraphics[width=0.49\linewidth]{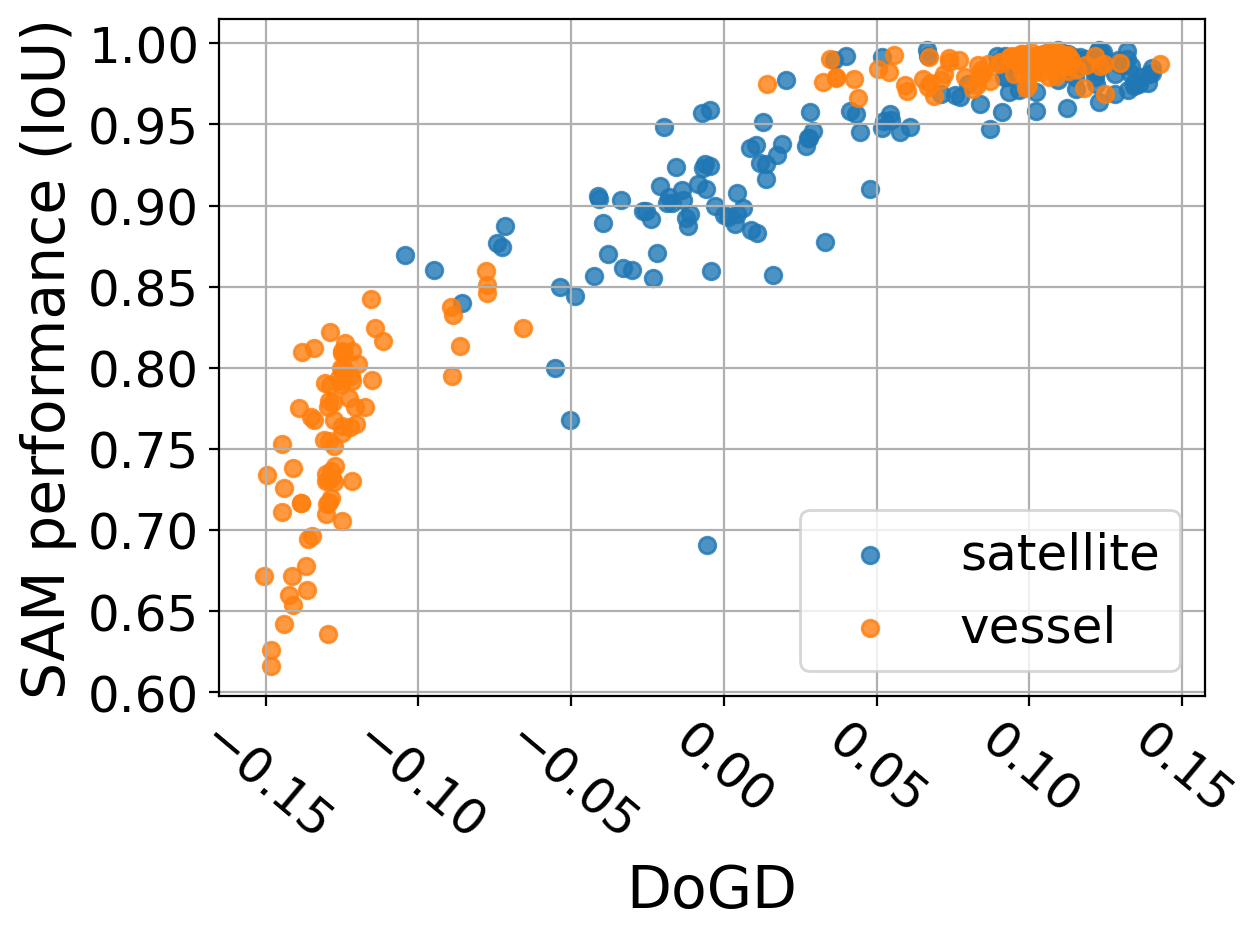}
    \caption{SAM segmentation IoU vs. object tree-likeness (CPR, left; DoGD, right) on the synthetic dataset with dilated object lines.}
    \label{fig:treelikeness_synthetic_dilated}
\end{figure}

We show the results of this procedure (how it affects the relationship between segmentation IoU and tree-likeness) in Fig. \ref{fig:treelikeness_synthetic_dilated}. We see that the result of IoU and object tree-likeness being correlated is mostly invariant to the object line thickening transformation. However, while the DoGD results are generally maintained because DoGD quantifies the deviation of filling-rate at different scales, we see that CPR experiences some degradation in its correlation with IoU for certain retinal vessel components. This is possibly due to the highly dense tree structures, originally thin with varying widths, forced to be similarly thickened and then possibly merged with each other, thereby reducing the overall CPR of the object. This may be a limitaiton of CPR in quantifying the tree or net-like geometric structure. Fortunately, such special cases where all object pixels fall around a fixed range of object skeletons are rare in realistic scenarios. 
% Also, despite the slight degradation in correlation statistics, as the CPR for those vessel structures decreases, they can also be segmented by SAM with higher IoU.

\subsection{Performance vs. NST intensity for all SFMs}
\label{app:NSTperformance}

\begin{figure}[htbp!]
    \centering
    \includegraphics[width=0.3\linewidth]{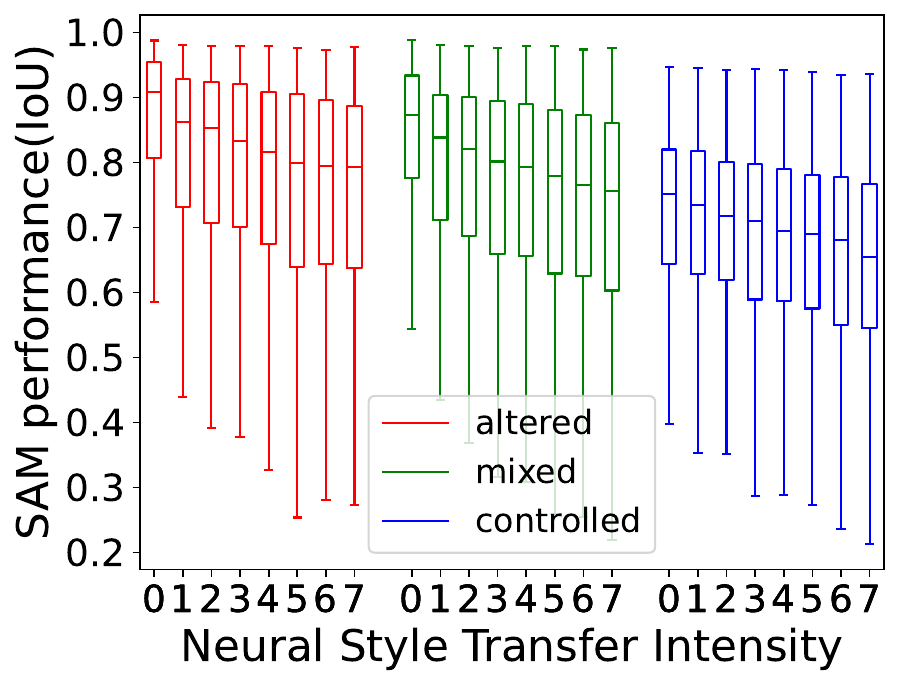}
    \includegraphics[width=0.3\linewidth]{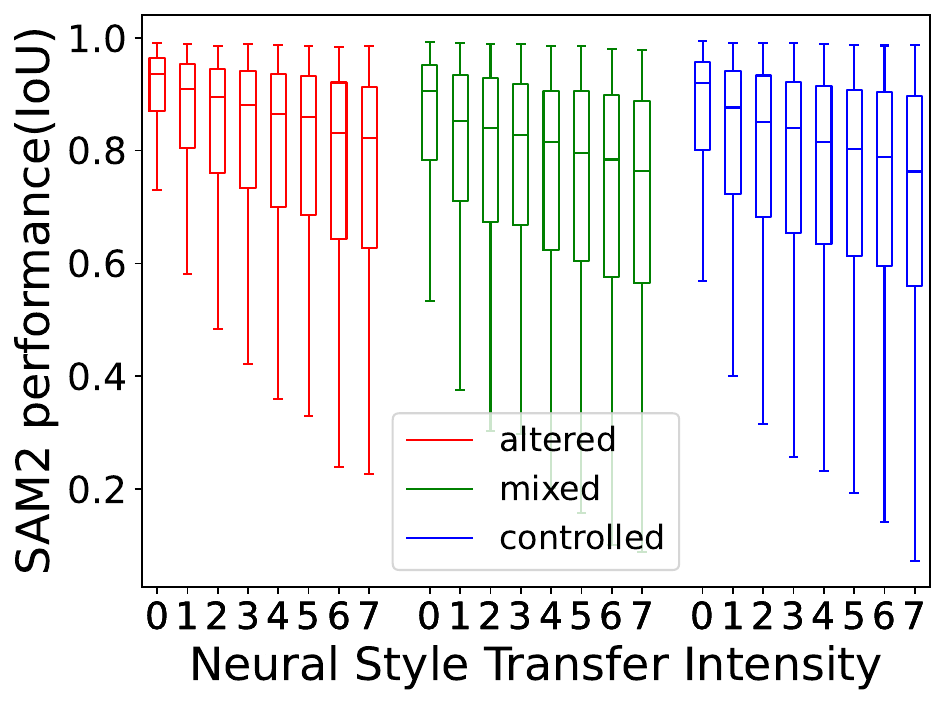}
    \includegraphics[width=0.3\linewidth]{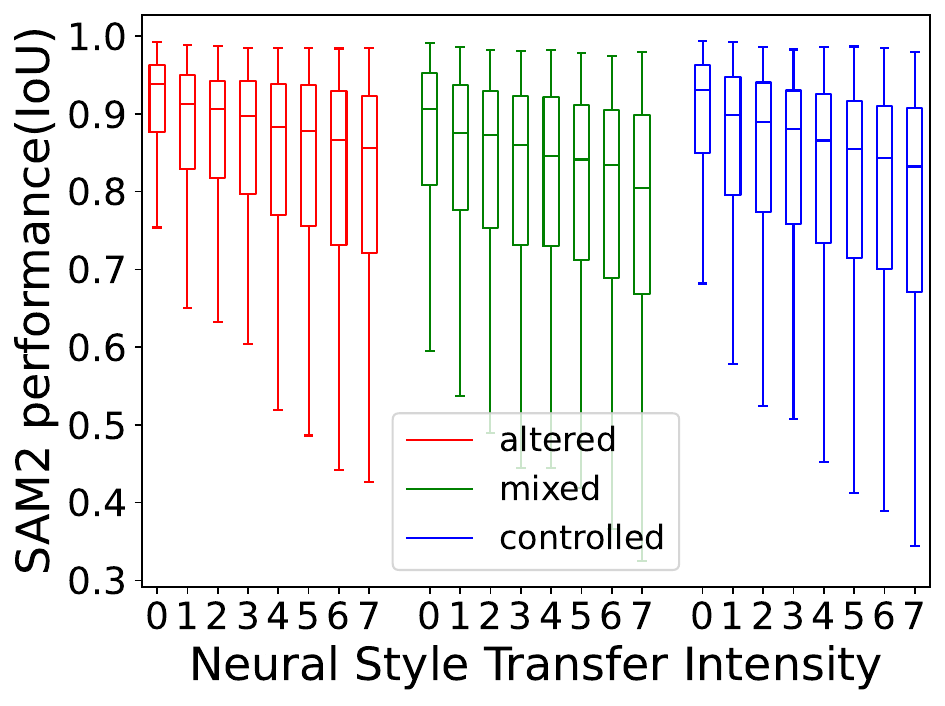}
    \includegraphics[width=0.3\linewidth]{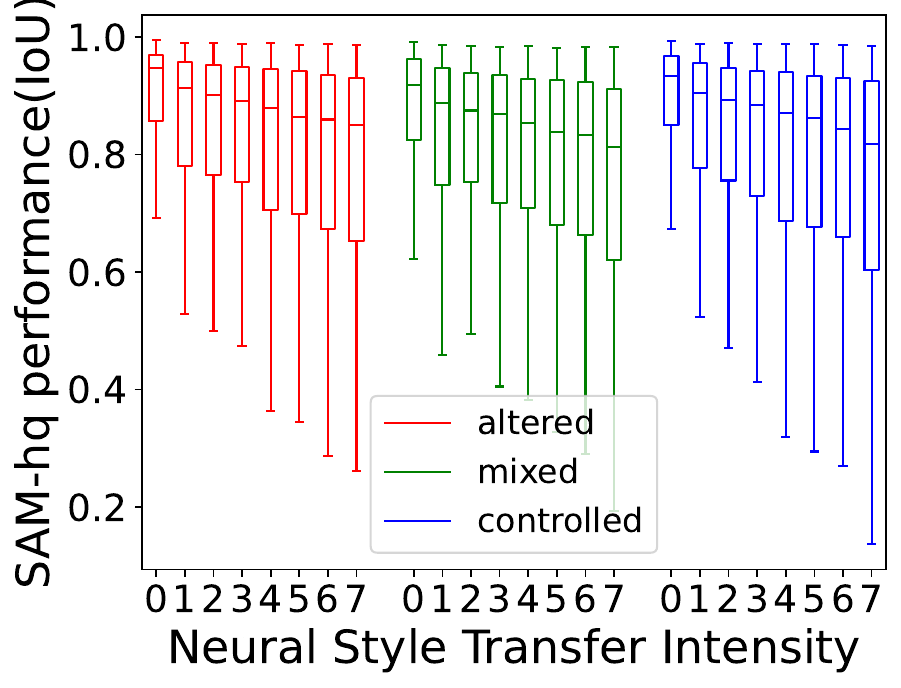}
    \includegraphics[width=0.3\linewidth]{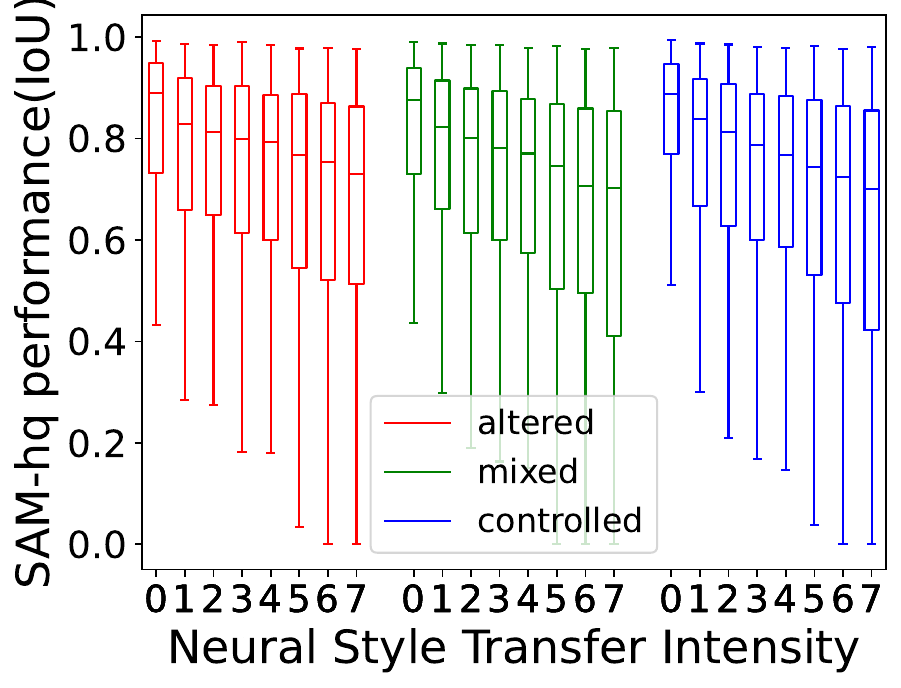}
    \caption{From left-to-right and top-to-bottom: segmentation IoU vs. NST intensity for all groups (controlled, altered, and mixed) for SAM ViT-B, SAM 2 ViT-B+, SAM 2 ViT-L, HQ-SAM ViT-H, and HQ-SAM ViT-B.}
    \label{fig:NST_intensity_effect_sep_allSFMs}
\end{figure}

To supplement Fig. \ref{fig:NST_intensity_effect_sep} right, we show the same results of segmentation IoU vs. NST intensity for all groups (controlled, altered, and mixed) for the additional SFMs (SAM ViT-B, SAM2 ViT-B+ and ViT-L, HQ-SAM ViT-H and ViT-B) in Fig. \ref{fig:NST_intensity_effect_sep_allSFMs}.

\subsection{Comparison to Segmentation Models Trained From Scratch}
\label{app:trainedfromscratch}

We additionally evaluate training models from scratch on each dataset to see if the discovered relationships of performance with tree-likeness (CPR and DoGD) and textural separability metrics is indeed a weakness unique to segmentation foundation models (SFMs), or if such cases are simply universally challenging from a general standpoint, and cannot be solved by training a model from scratch for the specific dataset.

For DIS, we use the pretrained IS-Net model available directly from the original paper \citep{DIS}, obtained from \url{https://github.com/xuebinqin/DIS}. For iShape, we use the iShapeInst models from \citet{wang2025ishapeinst}, obtained from \url{https://github.com/youngnuaa/iShapeInst}. We defer to those works for all architectural and training details. For Plittersdorf, no existing pretrained models could be found, so we trained a Mask R-CNN \citep{he2017mask} on the Plittersdorf training set, fine-tuning from COCO weights (we could not train \textit{completely} from scratch due to the relatively small size of this dataset), and tested on the test set. The Plittersdorf model utilized a ResNet-50-FPN backbone, with the \texttt{COCO-InstanceSegmentation/mask\_rcnn\_R\_50\_FPN\_3x.yaml} configuration of the Detectron2 \citep{wu2019detectron2} model zoo. For training, we used a batch size of 2, a learning rate of $2.5\times10^{-4}$, $30,000$ total training iterations, $300$ warmup iterations, and a learning rate schedule that reduced the learning rate by factor of 0.1 at 60\% and 80\% of total training time.

% \suggest{ add tables for ishape, and in-text results for DIS and plitters}

\begin{figure}[htbp!]
    \centering
    \includegraphics[width=0.25\linewidth]{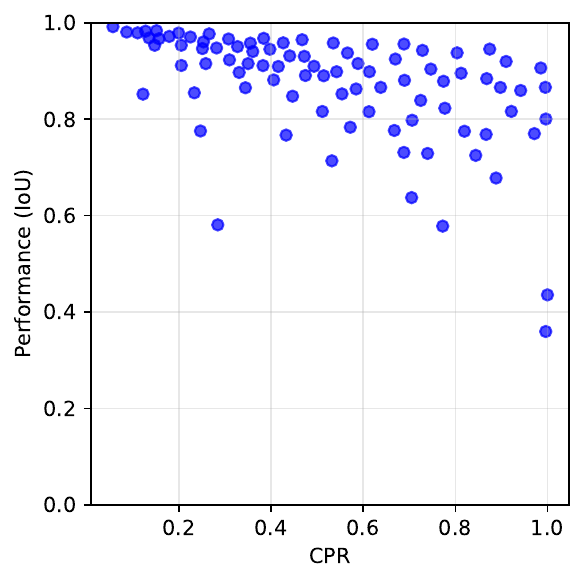}
    \includegraphics[width=0.25\linewidth]{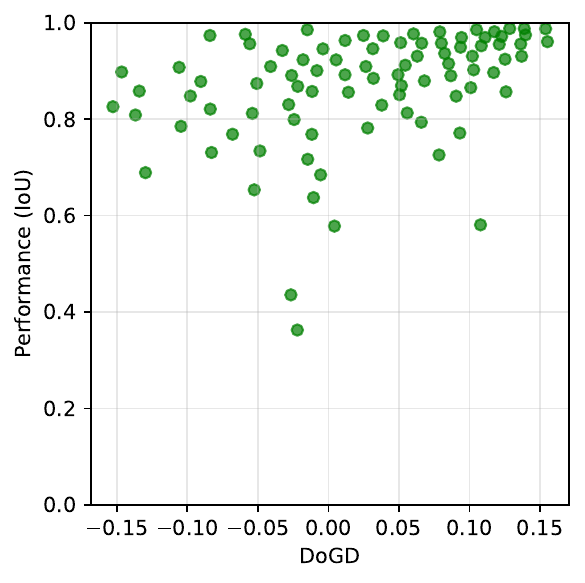}
    \caption{For IS-Net model trained from scratch on DIS: relationship of performance (IoU on the DIS validation set) with tree-likeness metrics: CPR, left and DoGD, right. For CPR, Spearman correlation $\rho=-0.61$ and Kendall $\tau=-0.45$. For DoGD, $\rho=0.46$ and $\tau=0.32$.}
    \label{fig:DIS_fromscratch}
\end{figure}

\begin{figure}[htbp!]
    \centering
    \includegraphics[width=0.25\linewidth]{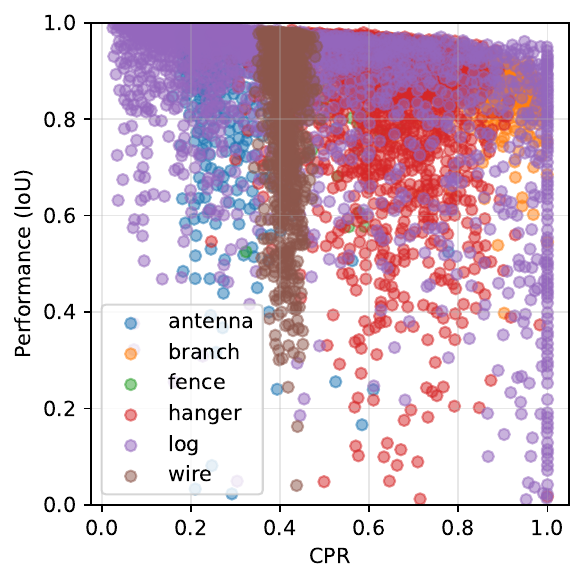}
    \includegraphics[width=0.25\linewidth]{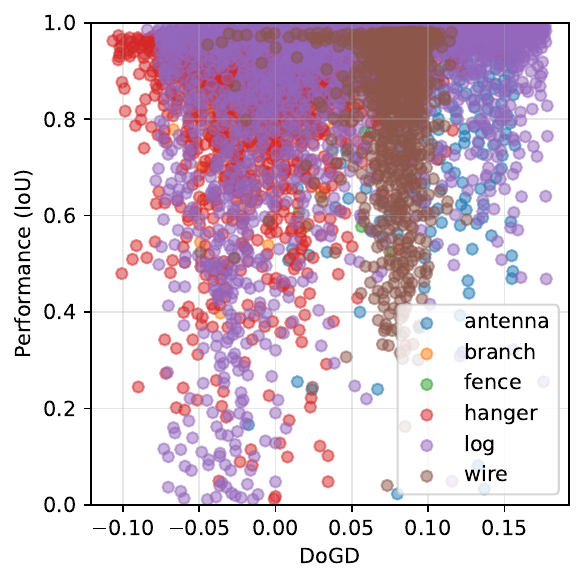}
    \includegraphics[width=0.25\linewidth]{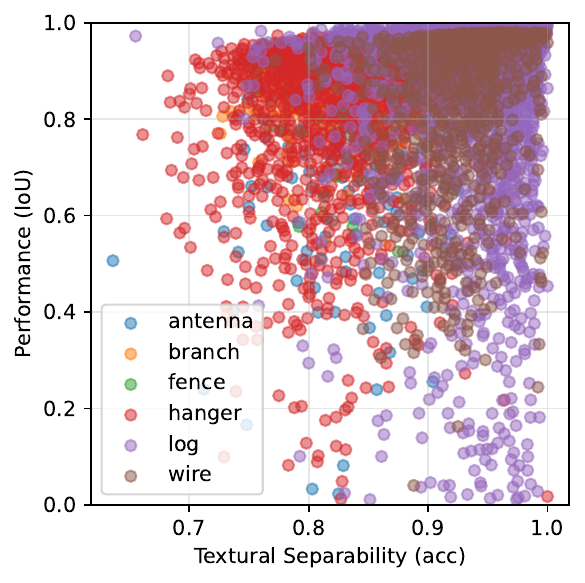}
    \caption{For iShapeInst models trained from scratch on iShape: relationship of performance (IoU) with tree-likeness metrics of CPR (left) and DoGD (center), and with textural separability (right). See Table \ref{tab:ishape_fromscratch} for numerical correlation results.}
    \label{fig:ishape_fromscratch}
\end{figure}

\begin{figure}[htbp!]
    \centering
    \includegraphics[width=0.25\linewidth]{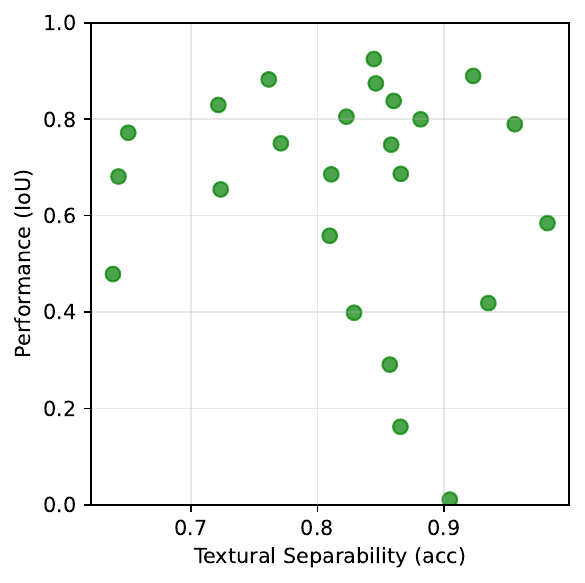}
    \caption{For Mask R-CNN model trained on Plittersdorf: relationship of performance (IoU) with textural separability (Spearman correlation $\rho=-0.07$ and Kendall $\tau=-0.05$.)}
    \label{fig:plitters_fromscratch}
\end{figure}

We show all results in Fig. \ref{fig:DIS_fromscratch} for DIS, Fig. \ref{fig:ishape_fromscratch} and Table \ref{tab:ishape_fromscratch} for iShape, and Fig. \ref{fig:plitters_fromscratch} for Plittersdorf. There are several important takeaways when we compare these results of models trained from scratch to the behavior of the segmentation foundation models in the main text. First, in all cases, we see that the models perform noticeably better (in average IoU) than the SFMs, which makes sense given that they were trained specifically and only on these datasets, which is unsurprising. In the case of object tree-likeness (on DIS and iShape), we see that the sharpness (approximate slope) of the effect of CPR and DoGD on performance is far lower than compared to the SFM results in the main text (Sec. \ref{sec:treelikeness_exp_real}). In other words, these models typically perform much better on objects of high tree-likeness (\ie, low CPR/high DoGD) than the SFMs, indicating that they do not share the same ``failure mode'' behavior. The correlations between IoU and tree-likeness are lower as well for these models trained from scratch (comparing Fig. \ref{figtab:treelikeness_results_real} for the SFMs to Figs. \ref{fig:DIS_fromscratch} and \ref{fig:ishape_fromscratch} and Table \ref{tab:ishape_fromscratch} here). We see similar, if not stronger results for textural separability: there is, on average, little to no correlation of performance with separability on both iShape (Fig. \ref{fig:ishape_fromscratch} and Table \ref{tab:ishape_fromscratch}) and Plittersdorf (Fig. \ref{fig:plitters_fromscratch}), compared to the noticeably stronger effect of an object's textural separability on SFM performance (Fig. \ref{figtab:sep_real}).  

We hypothesize that any remaining correlation of performance with these object characteristics (tree-likeness and textural separability) is simply because as shown in the figures, objects with higher tree-likeness or lower textural separability are simply less common in the dataset, so models trained from scratch will perform slightly worse on such cases compared to on cases which have more examples to learn from. Overall, these results support that SFMs are uniquely susceptible to objects of high tree-likeness or low textural separability compared to models trained from scratch on datasets that have these types of uncommon objects.

\begin{table*}[htbp!]
\setlength{\tabcolsep}{4pt}
\centering
% \small
% \fontsize{8pt}{8pt}\selectfont
% \scriptsize
\begin{tabular}{l||c|c|c|c|c|c|c||c}
\multicolumn{2}{c}{} & \multicolumn{1}{c|}{antenna} & \multicolumn{1}{c|}{branch} & \multicolumn{1}{c|}{fence} & \multicolumn{1}{c|}{hanger} & \multicolumn{1}{c|}{log} & \multicolumn{1}{c|}{wire} & \multicolumn{1}{c||}{\textbf{Avg.}}\\
\toprule
\multirow{2}{*}{\textbf{CPR}} & Spearman $\rho$ & -0.31 & -0.78 & -0.61 & -0.36 & -0.53 & -0.08 & \textbf{-0.45} \\
& Kendall $\tau$ & -0.21 & -0.59 & -0.45 & -0.25 & -0.38 & -0.05 & \textbf{-0.32} \\\midrule
\multirow{2}{*}{\textbf{DoGD}} & Spearman $\rho$ & 0.08 & 0.55 & 0.39 & 0.16 & 0.36 & -0.01 & \textbf{0.26} \\
& Kendall $\tau$ & 0.06 & 0.39 & 0.26 & 0.11 & 0.25 & -0.01 & \textbf{0.18} \\\midrule
\multirow{2}{*}{\tworow{\textbf{Textural}}{\textbf{Separability}}} & Spearman $\rho$ & 0.34 & 0.40 & 0.51 & 0.43 & 0.10 & 0.22 & \textbf{0.33} \\
& Kendall $\tau$ & 0.23 & 0.27 & 0.35 & 0.30 & 0.07 & 0.15 & \textbf{0.23} \\
\bottomrule
\end{tabular}
\caption{For iShape, nonlinear (Spearman $\rho$, Kendall $\tau$) correlations between performance of models trained from scratch (IoU) and both tree-likeness metrics (CPR and DoGD) as well as textural separability.}
\label{tab:ishape_fromscratch}
\end{table*}

\subsection{All results on MOSE}
\label{app:MOSEallresults}

\begin{figure}[htbp!]
    \centering
    \includegraphics[width=0.33\linewidth]{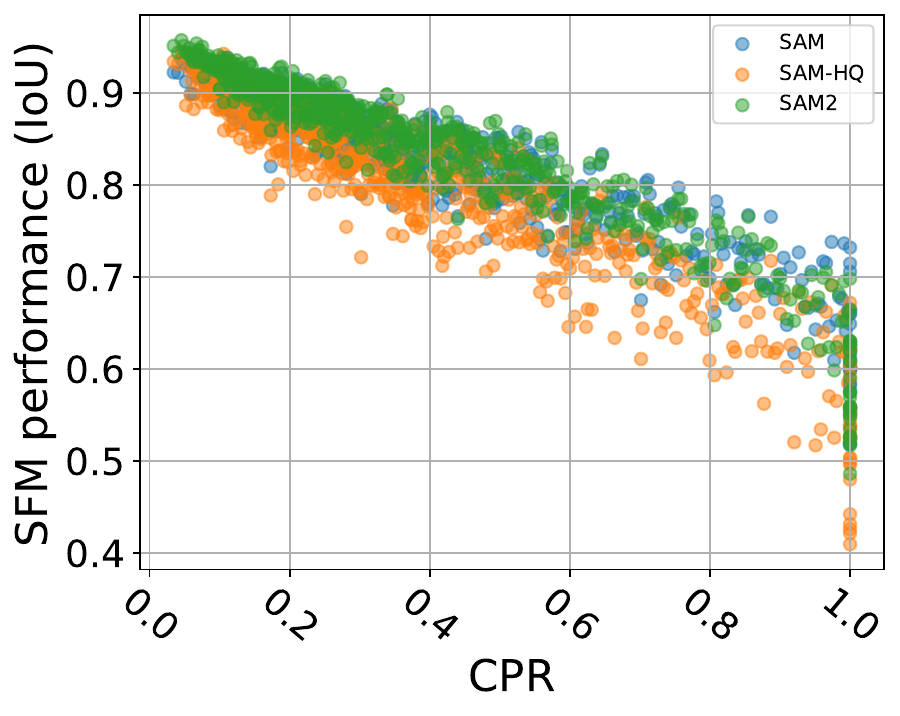}
    \includegraphics[width=0.33\linewidth]{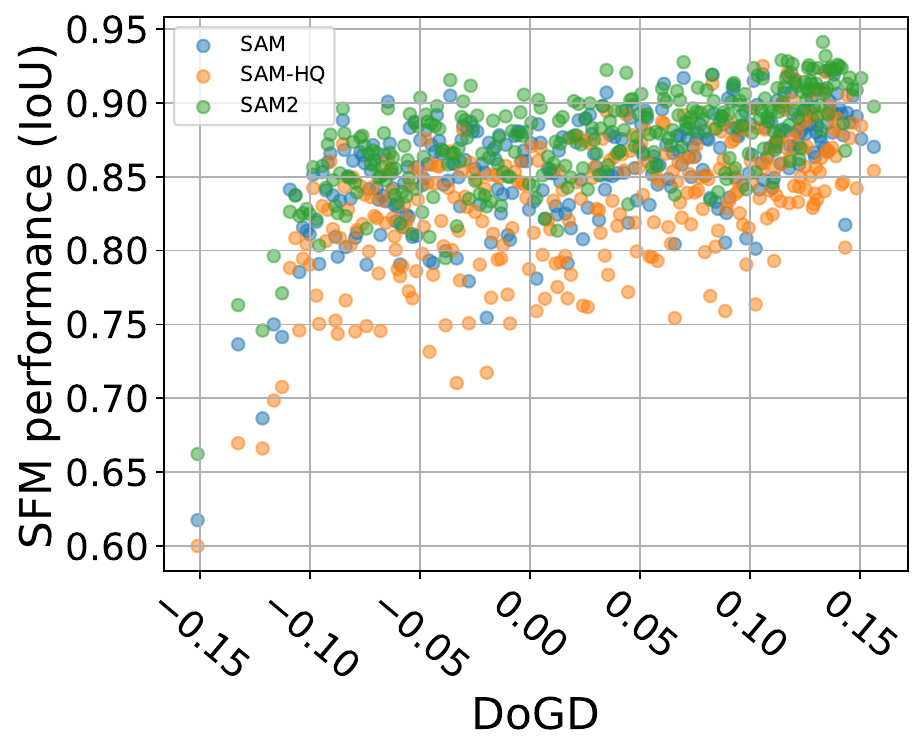}
    \includegraphics[width=0.33\linewidth]{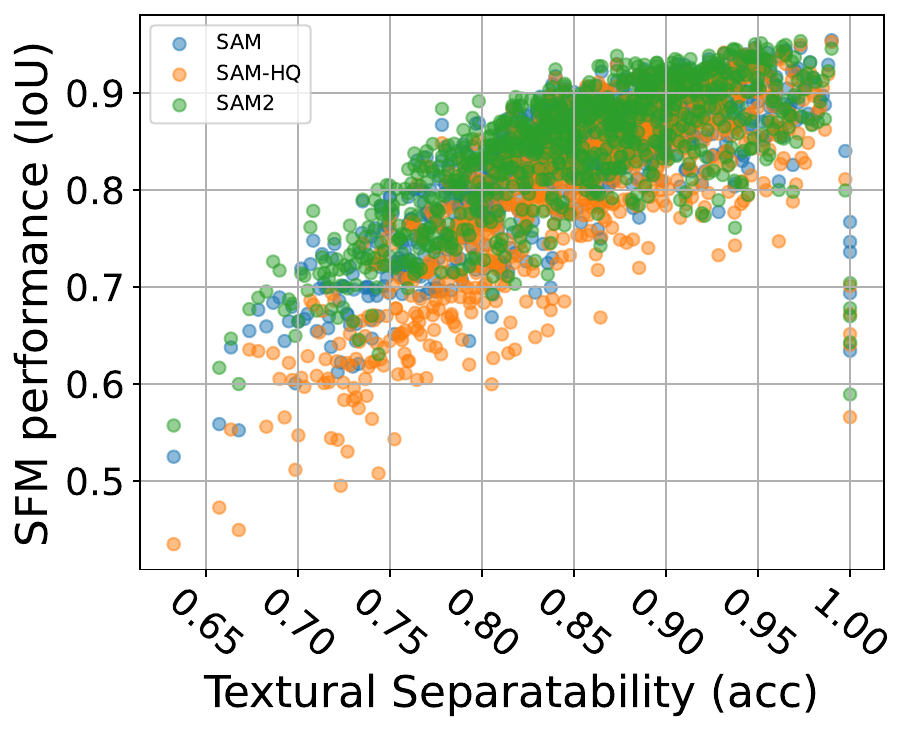}
    \caption{Segmentation IoU vs. object tree-likeness (CPR, left, and DoGD, center) and textural separability (right) for the MOSE dataset, for all three evaluated SFMs.}
    \label{fig:MOSEplots}
\end{figure}

In Fig. \ref{fig:MOSEplots} we show the corresponding plots for the SFM performance (IoU) vs. tree-likeness (table in Fig. \ref{figtab:treelikeness_results_real}) and textural separability (table in Fig. \ref{figtab:sep_real} relationship results on the MOSE dataset. These are for the same SFM architectures as shown in the plots of Figs. \ref{figtab:treelikeness_results_real}) and \ref{figtab:sep_real}).

\end{document}